\documentclass[lettersize,journal]{IEEEtran}

\IEEEoverridecommandlockouts                              
\newtheorem{theorem}{Theorem}[section]
\newtheorem{cor}[theorem]{Corollary}

\newtheorem{prop}[theorem]{Proposition}
\newtheorem{assumption}[theorem]{Assumption}
\newtheorem{problem}{Problem}
\newtheorem{definition}[theorem]{Definition}
\newtheorem{rem}[theorem]{Remark}
\newtheorem{ex}[theorem]{Example}

\usepackage{graphicx} 
\usepackage{amsmath}
\usepackage{amssymb}
\usepackage{epstopdf}
\usepackage{cite}
\usepackage[noend,ruled,linesnumbered]{algorithm2e}
\SetKwComment{Comment}{$\triangleright$\ } {}
\usepackage{multirow}
\usepackage{rotating}
\usepackage{subfigure} 
\usepackage{color} 
\usepackage{mysymbol}
\usepackage[dvipsnames]{xcolor}
\usepackage[hyphens]{url}

\usepackage[breaklinks=true, colorlinks, bookmarks=true, citecolor=Black, urlcolor=Violet,linkcolor=Black]{hyperref}

\newcommand{\set}[1]{\left\{#1\right\}}

\usepackage{comment}
\usepackage[colorinlistoftodos,prependcaption,textwidth=1.5cm,textsize=tiny]{todonotes}
\begin{document}
\title{
Minimum-Violation Temporal Logic Planning for Heterogeneous Robots under Robot Skill Failures
}
\author{Samarth Kalluraya, Beichen Zhou, Yiannis Kantaros
\thanks{S. Kalluraya, B. Zhou, and Y. Kantaros are with the Department of Electrical and Systems Engineering, Washington University in St. Louis, St. Louis, MO, 63130, USA. This work was supported by the ARL grant DCIST CRA W911NF-17-2-0181 and the NSF award CCF \#2403758.
        {\tt\small \{k.samarth,beichen,ioannisk@wustl.edu\}}}%
}

\maketitle 
\begin{abstract}
In this paper, we consider teams of robots with heterogeneous skills (e.g., sensing and manipulation) tasked with collaborative missions described by Linear Temporal Logic (LTL) formulas. These LTL-encoded tasks require robots to apply their skills to specific regions and objects in a temporal and logical order. While existing temporal logic planning algorithms can synthesize correct-by-construction plans, they typically lack reactivity to unexpected failures of robot skills, which can compromise mission performance.
This paper addresses this challenge by proposing a reactive LTL planning algorithm that adapts to unexpected failures during deployment. 
Specifically, the proposed algorithm reassigns sub-tasks to robots based on their functioning skills and locally revises team plans to accommodate these new assignments and ensure mission completion. The main novelty of the proposed algorithm is its ability to  handle cases where mission completion becomes impossible due to limited functioning robots.
Instead of reporting mission failure, the algorithm strategically prioritizes the most crucial sub-tasks and locally revises the team's plans, as per user-specified priorities, to minimize mission violations. We provide theoretical conditions under which the proposed framework computes the minimum-violation task reassignments and team plans. We provide \textcolor{black}{numerical and hardware} experiments to demonstrate the efficiency of the proposed method.
\end{abstract}
\IEEEpeerreviewmaketitle
\vspace{-0.1cm}

\vspace{-0.5cm}
\section{Introduction} \label{sec:Intro}
\vspace{-0.1cm}
\IEEEPARstart{M}{otion} planning is a fundamental problem in robotics that has received significant research attention over the years. This problem traditionally involves generating robot trajectories that reach a goal configuration from an initial one, while avoiding obstacles \cite{lavalle2006planning}. To address this challenge, several planning algorithms have been proposed, including potential fields and navigation functions \cite{koditschek1990robot,loizou2017navigation,loizou2022mobile,paternain2017navigation}, search-based methods \cite{koenig2006new,sun2008generalized}, and sampling-based approaches \cite{karaman2011sampling,kavraki1996probabilistic,lavalle2001randomized}. A comprehensive literature review can be found in \cite{garrett2021integrated,antonyshyn2023multiple,matni2024towards}. 

More recently, new planning approaches have been proposed that can handle a richer class of tasks, than the classical reach-avoid tasks, and can capture temporal and logical requirements. Such tasks include surveillance \cite{leahy2016persistent}, coverage \cite{fainekos2005temporal}, data-gathering \cite{guo2017distributed}, and intermittent connectivity \cite{kantaros2016distributedInterm} and can be captured using formal languages, such as Linear
Temporal Logic (LTL) \cite{baier2008principles}.
Task and motion planning algorithms for LTL-encoded requirements are presented in \cite{kloetzer2008fully, fainekos2005hybrid, smith2011optimal, chen2012formal, vasile2013sampling, luo2021abstraction, tumova2016multi, shoukry2017linear, kantaros2020stylus, gujarathi2022mt,chen2024fast,Cardona2024Planning,kurtz2023temporal,liu2024nngtl,luo2024decomposition} assuming robot teams with known dynamics operating known environments. These works have been extended recently to handle unknown static environments \cite{guo2013revising,guo2015multi,maly2013iterative,livingston2012backtracking,livingston2013patching,kantaros2020reactive,Kantaros2022perception}, unknown dynamic environments \cite{Kalluraya2023multi,li2022online,purohit2021dt,maity2015motion,li2021safe} and uncertain robot dynamics \cite{hasanbeig2019reinforcement,sun2022neurosymbolic,kantaros2022accelerated,guo2023hierarchical,guo2018probabilistic}.
However, these works lack reactivity to unexpected failures of robot capabilities that may occur unexpectedly during mission execution  due to e.g., inclement weather, human interventions, system component malfunctions, or adversarial attacks \cite{schlotfeldt2021resilient,notomista2021resilient,ramachandran2023resilient}.

This paper aims to address this challenge by proposing a reactive-to-failures planning algorithm for multi-robot systems with collaborative LTL tasks. Specifically, we consider robots that are heterogeneous with respect to their skills which may include e.g., mobility, sensing, or manipulation. The robots are responsible for accomplishing a high-level collaborative mission, expressed as an LTL formula, requiring them to apply their capabilities at certain areas/objects. We consider cases where failures that include permanent loss of capabilities (e.g., grasping) or complete removal of robots (e.g., due to battery draining) may occur unexpectedly at any time during deployment. 
Our objective is to design reactive multi-robot plans, defined as sequences of robot states and actions, that can adapt to these failures. 
To tackle this task planning problem, we propose a joint
task re-allocation and re-planning framework.
First, when failures occur, the task re-allocation algorithm assigns robots to new sub-tasks based on their remaining functional capabilities. A key challenge is that mission completion may no longer be possible due to the limited number of functioning robots. 
Instead of reporting mission failure, the proposed method strategically prioritizes re-allocating the most crucial sub-tasks to robots with the required functioning skills, as per user-specified priorities.  
\textcolor{black}{Second, we introduce a planner that revises the current team \textcolor{black}{plans} to accommodate these new task assignments while minimally violating the LTL-encoded mission specification.
The proposed algorithm aims to minimally disrupt the team behavior by performing the fewest possible task re-assignments and locally  revising the team \textcolor{black}{plans} to accommodate the robot failures and the task re-allocations, all while ensuring minimal mission violations.} 
We demonstrate the efficiency of the proposed method through extensive comparative simulations and hardware experiments.

\textbf{Related works:} 
(i) \textit{Task assignment} methods for LTL-encoded missions have been proposed in \cite{schillinger2016decomposition,banks2020multi,Luo2022temporal, Li2023Fast,chen2024distributed,liu2022time,fang2024continuous}. These works perform task assignment offline and do not consider robot failures. In case of failures, these approaches could be employed online to globally re-allocate tasks to the robots; \textcolor{black}{however, this is impractical and, possibly, unnecessary especially for small number of failures relative to the team size.} Furthermore, they cannot handle cases where there is not enough number of surviving
robots to take over the subtasks and, therefore, they cannot generate minimum violation solutions. \textcolor{black}{
(ii) Related are also the works on \textit{minimum-violation temporal logic planning} \cite{tumova2016Least, lahijanian2016iterative, lahijanian2016specification, Vasile2017Minimum, cai2023learning}. Similar to this paper, the key idea in these works is to allow for partial fulfillment of LTL specifications when it is impossible to satisfy all encoded requirements. However, there are two fundamental differences with our work that prevent them from directly addressing the problem considered in this paper. First, they do not consider robot failures. Instead, violations of the LTL formula may occur due to factors such as logical conflicts, timing constraints, environmental obstacles (e.g., unknown obstacles blocking access to mission-critical regions), exogenous disturbances that make certain mission components hard or impossible to satisfy. Second, they address single-robot planning problems. While these approaches could theoretically be applied to multi-robot systems in a centralized manner (treating the team as a high-dimensional robot), their reactivity is limited to re-planning in response to changes, without task reallocation mechanisms. This is inefficient in multi-robot settings, where dynamically reassigning tasks—rather than simply abandoning them—is often preferable. For instance, in a multi-robot scenario, if a robot cannot reach a desired region due to a narrow corridor, it is preferable to re-assign the sub-task to another robot capable of navigating narrow corridors, rather than abandoning it (as, e.g., a centralized implementation of \cite{lahijanian2016iterative} would do).}
%
(iii) \textcolor{black}{The closest works to this paper are presented in \cite{Feifei2022failure,Faruq2018Simultaneous,Zhou2022Reactive,kalluraya2023resilient} that propose \textit{multi-robot planners that adapt to unexpected robot failures}. Specifically, \cite{Feifei2022failure,Faruq2018Simultaneous} consider homogeneous robots and build a product automaton modeling the multi-robot state space, the specification space, as well as possible robot failures. Using this product system, reactive control strategies are constructed. It is worth noting that, unlike our work, \cite{Feifei2022failure,Faruq2018Simultaneous} consider only failure cases where a robot is entirely out-of-service and, therefore, removed from the field. 
\textcolor{black}{While \cite{Zhou2022Reactive} addresses heterogeneous tasks and proposes a reactive framework for adapting to environment and robot state changes, in case of a robot skill failure they necessitate reassignment of all the remaining tasks in the mission to the available robots.
This limitation is addressed in our prior work \cite{kalluraya2023resilient} by proposing a local task re-allocation and re-planning framework that minimizes the number of task reassignments.} A common assumption in these works is that the mission remains feasible despite robot failures, which may not always hold in practice due to limited number of functioning robots. If the mission becomes infeasible, then these planners, unlike the proposed one, return a mission failure message. Building upon our prior work \cite{kalluraya2023resilient}, this paper aims to address this limitation by proposing a planner that computes minimum mission violation plans.} 
\textcolor{black}{Specifically, we extend \cite{kalluraya2023resilient} in the following ways. First, we augment the task re-allocation method from \cite{kalluraya2023resilient} by incorporating a mission violation cost function to handle cases where the mission becomes infeasible due to failures. Second, we develop a novel re-planning algorithm capable of `locally' revising the \textcolor{black}{plans}, ensuring that the new plans minimally violate the mission specification. 
The re-planner in \cite{kalluraya2023resilient} assumes the LTL task remains feasible, and therefore, cannot straightforwardly handle infeasible missions. 
Third, we provide minimum-violation guarantees regarding the resulting task allocation and team plan that do not exist in \cite{kalluraya2023resilient}.} 

\textbf{Contributions:}
\textit{First}, we formulate a new reactive temporal logic planning problem for heterogeneous robot teams in the presence of \textcolor{black}{robot skill failures} that may render an assigned LTL-encoded collaborative mission infeasible. 
\textit{Second}, we propose a reactive planning algorithm that can adapt to unexpected robot failures while generating least-violating plans when mission completion is not possible. 
\textit{Third}, our algorithm prioritizes minimal disruption of the original plan in the event of failures, eliminating the need for global re-assignment or re-planning. \textit{Fourth}, we provide theoretical conditions under which the proposed framework computes the minimum-violation task allocation and team plan.
\textit{Fifth}, we provide \textcolor{black}{numerical and hardware experiments} validating the efficiency of our algorithm\footnote{\textcolor{black}{The code is available at \href{https://github.com/kantaroslab/MinVio-MRP}{https://github.com/kantaroslab/MinVio-MRP}}.
}.


\vspace{-0.2cm}
\section{Problem Definition} \label{sec:PF}
\vspace{-0.1cm}
\subsection{Environment and Modeling of Robots}\label{sec:PFmodelRobot}
\vspace{-0.1cm}
We consider a known environment $\Omega$ with obstacle-free space denoted by $\Omega_{\text{free}}\subseteq\Omega$. 
The space $\Omega_{\text{free}}$ contains $M>0$ regions/objects of interests, denoted by $\ell_e$, with known locations $\bbx_e$, $e\in\{1,\dots,M\}$. A team of $N>0$ mobile robots operate in $\Omega_{\text{free}}$ with dynamics: $\bbp_{j}(t+1)=\bbf_j(\bbp_{j}(t),\bbu_{j}(t))$, for all $j\in \ccalR=\{1,\dots,N\}$,  where $\bbp_{j}(t)\in\mathbb{R}^n$ stands for the state (e.g., position and orientation) of robot $j$ at discrete time $t$, and $\bbu_{j}(t)\in\mathbb{R}^b$ stands for control input. The dynamics of all robots are concisely represented as: $$\bbp(t+1)=\bbf(\bbp(t),\bbu(t)),$$
where $\bbp(t)\in \mathbb{R}^{nN}$ and $\bbu(t)\in \mathbb{R}^{bN}$ for $t\geq0$. \textcolor{black}{We assume that $\bbp(t)$ is known at all time steps and that $\bbf$ models holonomic dynamics allowing robots to follow any desired \textcolor{black}{plan}. We also assume that all regions/objects $\ell_e$ are accessible to all robots.} 

\vspace{-0.4cm}
\subsection{Heterogeneous Robot 
Abilities and Robot Failures}\label{sec:abilities_def}
\vspace{-0.1cm}
The robots are heterogeneous with respect to their skills and together possess a total of $C>0$ distinct abilities. Each ability is denoted by $c\in\{1,\dots,C\}$, representing skills like mobility, communication, manipulation, sensing capabilities, etc. We define the set $\ccalC=\{0,1,\dots,c,\dots,C\}$, which encompasses all the robot capabilities. \textcolor{black}{We explicitly include $0$ in this set to denote idling, that is the robot does nothing.}
%
The vector $\bbZ_j(t)=[\zeta_1^j(t),\dots,\zeta_c^j(t),\dots,\zeta_C^j(t)]$ represents skills of robot $j$, where $\zeta_c^j(t)$ is $1$ if robot $j$ has the ability $c$ at time $t$ and is $0$ otherwise.
If $\zeta_c^j(t-1)=1$ and $\zeta_c^j(t)=0$, we say capability $c$ of robot $j$ failed at time $t$. Complete robot failure is represented by setting $\bbZ_j(t)$ to a zero vector.
%
We assume robots have a health monitoring system and all-to-all communication. Thus, even if failures occur at unknown time instants $t\geq0$, vectors $\bbZ_j(t)$ for all robots $j$ are known to all. 
%
%
We further define a robot team $\ccalT_c(t)$ at time $t$ as a set collecting the robots with $\zeta_c^j(t) = 1$,
%
i.e., $\ccalT_c(t)=\{j\in \ccalR~|~\zeta_c^j(t) = 1 \}$; a robot may belong to more than one team. 

\vspace{-0.4cm}
\subsection{Mission Specification and Penalties}\label{sec:PFLTLMission}
\vspace{-0.1cm}
The objective for the robots is to complete a complex, long-term collaborative mission encoded by a global Linear Temporal Logic (LTL) specification $\phi$. 
This mission requires them to apply their skills at specific regions $\ell_e$ in a temporal and logical order using LTL grammar which can be found in \cite{baier2008principles}. 
Particularly, the LTL formula comprises a set of atomic propositions (AP), i.e., Boolean variables, denoted by $\mathcal{AP}$, Boolean operators, (i.e., conjunction $\wedge$, and negation $\neg$), and two temporal operators, next $\bigcirc$ and until $\mathcal{U}$. 
%
We consider LTL tasks constructed using the following team-based atomic predicates: 
\vspace{-0.3cm}
\begin{equation}\label{eq:pip}
\pi_{\ccalT_c}(j, \ell_e)=
 \begin{cases}
  \text{true}, & \text{if $j\in\ccalT_c \text{ applies } c\text{ at }\ell_e$}\\
  \text{false}, & \text{otherwise.}
 \end{cases}       
\end{equation}
\vspace{-0.1cm}
A predicate in \eqref{eq:pip} is true when \textit{any} robot in the team $\ccalT_c$ applies the skill $c$ (e.g., `grasp') at the region/object $\ell_e$. The robot that has been assigned with this sub-task/predicate is denoted by $j$. \textcolor{black}{We also define the  predicate $\Bar{\pi}_{\ccalT_{\hat{c}}}(j, c, \ell_e)$ as:} 
\vspace{-0.15cm}
\begin{equation}\label{eq:negpip}
\Bar{\pi}_{\ccalT_{\hat{c}}}(j, c, \ell_e)=
 \begin{cases}
  \text{true}, & \text{if $j\in\ccalT_{\hat{c}} ~\text{does not apply } c\text{ at }\ell_e$}\\
  \text{false}, & \text{otherwise.}
 \end{cases}       
\end{equation}
The predicate in \eqref{eq:negpip} is satisfied if robot $j$ in $\ccalT_{\hat{c}}$ \textit{does not} apply the skill $c$ at $\ell_e$. Observe in \eqref{eq:negpip} that $\hat{c}$ is not necessarily the same as $c$. \textcolor{black}{If \eqref{eq:negpip} should hold for all robots $j$ in the team $\ccalT_{\hat{c}}$ then, with slight abuse of notation, we represent it as $\Bar{\pi}_{\ccalT_{\hat{c}}}(\varnothing, c, \ell_e)$. Formally, we have that $\Bar{\pi}_{\ccalT_{\hat{c}}}(\varnothing, c, \ell_e)=\bigwedge_{j\in\ccalT_{\hat{c}}} \Bar{\pi}_{\ccalT_{\hat{c}}}(j, c, \ell_e)$.
} 
%
An example of a simple LTL mission defined over predicates of the form \eqref{eq:pip}-\eqref{eq:negpip} is given in Ex. \ref{ex:LTL}.  

Each predicate in $\phi$ is assigned a penalty value indicating the priority of the sub-task captured by the predicate. The larger the penalty, the more important the corresponding sub-task is. This is modeled by the following function.
\begin{definition} [Penalty Function]\label{def:penalty}
\textcolor{black}{The function $F:\mathcal{AP} \rightarrow \mathbb{R}_{+} \cup {\infty}$ assigns a positive finite penalty if a false predicate ($\pi \in \mathcal{AP}$) of form (1) is treated as true, and $\infty$ if a false predicate of form (2) is treated as true.}
\end{definition}

Throughout the paper, we make the following assumptions regarding the atomic predicates \eqref{eq:pip}-\eqref{eq:negpip} and the robot skills; see also Remark \ref{rem:assumptionsPred}.

\begin{assumption}[Robot Skills]\label{as2}
    \textcolor{black}{We assume that the robots cannot apply more than one skill at a time.}
\end{assumption}

\begin{assumption}[Initial Assignment]\label{as1}  We assume an initial assignment of robots $j$ to all predicates of the form \eqref{eq:pip} is given and that the resulting mission $\phi$ is feasible, meaning there are no (i) logical conflicts (e.g., requiring a robot to both reach and avoid the same region), (ii) `physical' conflicts (e.g., requiring the same robot to apply more than one skill simultaneously).
\end{assumption}

\begin{assumption}[Independent Subtasks]\label{as3}
\textcolor{black}{All predicates in $\mathcal{AP}$ are independent, i.e., there does not exist any pair of predicates $\pi_{\mathcal{T}_c}(j, \ell_e)$ (see \eqref{eq:pip}) in $\mathcal{AP}$ that are required to be satisfied by the same robot $j$.}
\end{assumption}

\begin{assumption}[Hard Safety Constraints]\label{as4}
\textcolor{black}{We assume that the predicates of the form \eqref{eq:negpip} are used to model hard safety constraints, i.e., to capture skills that certain robots should always or temporally avoid applying. Relaxing/sacrificing these requirements is not allowed. This has the following two implications. First, the predicates in \eqref{eq:negpip} can only appear without a negation $\neg$ in front of them and as parts of Boolean formulas $\psi$ that appear in an LTL formula in the following two ways: (i) $\square \psi$ and (ii) $\psi \ccalU \xi$. Second, the function $F$ returns $\infty$ for all predicates of the form \eqref{eq:negpip}.} 
\end{assumption}

\vspace{-0.4cm}
\subsection{From LTL Missions to Automata}\label{sec:nba}

\vspace{-0.1cm}
Given an LTL mission $\phi$, we translate it, offline, into a Nondeterministic B$\ddot{\text{u}}$chi Automaton (NBA)  \cite{baier2008principles}. 

\begin{definition}[NBA]
A Nondeterministic B$\ddot{\text{u}}$chi Automaton (NBA) $B$ over $\Sigma=2^{\mathcal{AP}}$ is defined as a tuple $B=\left(\ccalQ_{B}, \ccalQ_{B}^0,\Sigma,\delta_B,\ccalQ_{B}^F\right)$, where $\ccalQ_{B}$ is the set of states, $\ccalQ_{B}^0\subseteq\ccalQ_{B}$ is a set of initial states, $\Sigma$ is an alphabet, $\delta_B:\ccalQ_B\times\Sigma\rightarrow2^{\ccalQ_B}$ is a non-deterministic transition relation, and $\ccalQ_{B}^F\subseteq\ccalQ_{B}$ is a set of accepting/final states. 
\end{definition}


Next, we discuss the accepting condition of the NBA that is used to find plans that satisfy $\phi$. \textcolor{black}{We define a labeling function $L:\mathbb{R}^{nN}\times\mathcal{C}^{N}\rightarrow \Sigma$ determining which atomic propositions are true} given the multi-robot state $\bbp(t)$ and the applied skills $\bbs(t)$. 
An infinite run $\rho_B= q_B(1),q_B(2) \dots$ of $B$ over an infinite word $w = \sigma(0)\sigma(1)\sigma(2)\dots\in\Sigma^{\omega}$, where $\sigma(t) \in \Sigma$, $\forall t \in \mathbb{N}$, is an infinite sequence of NBA states $q_B(t)$, $\forall t \in \mathbb{N}$, such that $q_B(t+1))\in\delta_B(q_B(t), \sigma(t))$ and $q_B(0)\in\ccalQ_B^0$. An infinite run $\rho_B$ is called 
\textit{accepting} if $\text{Inf} ( \rho_B) \bigcap \mathcal{Q}_B^F \neq \emptyset$, where $\text{Inf} (\rho_B)$ represents the set of states that appear in $\rho_B$ infinitely often.\footnote{Since in Assumption \ref{as2} we assume that robots cannot apply more than one skill at a time, in what follows, we assume that the NBA is pruned as in \cite{luo2021abstraction} by removing transitions that violate this assumption.}

\begin{figure}[t]
  \centering
    \subfigure[Offline-designed plans]{
        \label{fig:3ra}
        \includegraphics[width=0.47\linewidth]{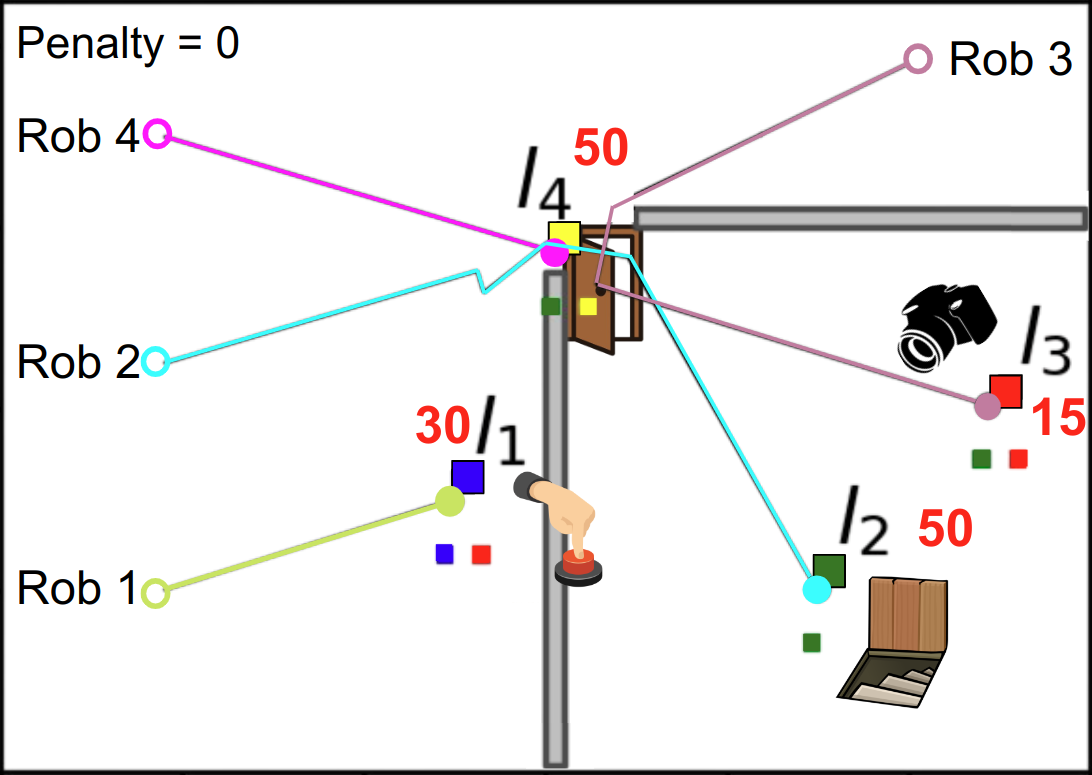}}
    \subfigure[Plans after failure]{      
        \label{fig:3rb}
        \includegraphics[width=0.47\linewidth]{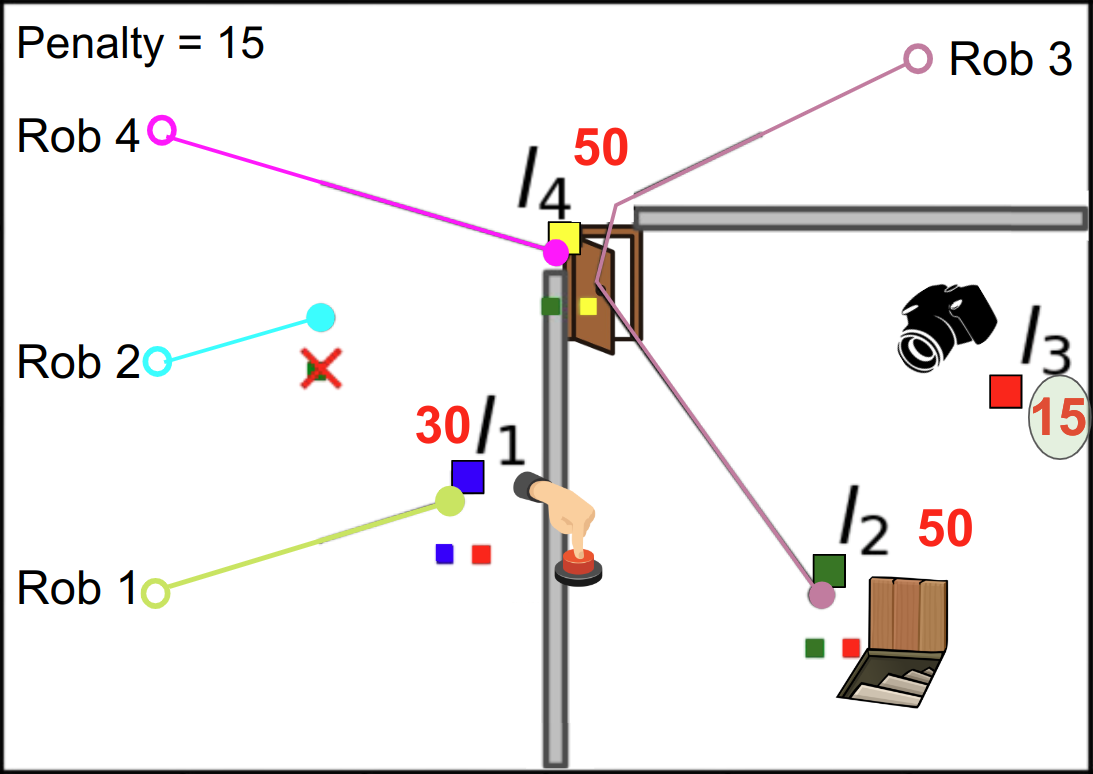}}
 \vspace{-0.2cm}
 \caption{
\textcolor{black}{The small squares below each robot indicate the abilities each robot possesses; the ability to push buttons (blue), retrieve objects (green), take photos (red), and open doors (yellow). The locations are represented by the larger squares, whose color indicates which ability needs to be used at that location as per the LTL-encoded mission $\phi$ discussed in Example \ref{ex:LTL}. The mission requires robots $1$, $2$, and $3$ to simultaneously execute their sub-tasks. The penalties for not completing each sub-task is shown in red next to each location. Fig. \ref{fig:3ra} shows the plans designed offline and \ref{fig:3rb} shows the minimum violation plans, planned after robot $2$ fails (red X on skill).}
}
\vspace{-0.6cm}
\label{fig:3robot}
\end{figure}

\vspace{-0.4cm}
\subsection{Multi-Robot Plans}
\vspace{-0.1cm}

A multi-robot plan satisfying an LTL-encoded task $\phi$ can be constructed using existing methods such as \cite{luo2021abstraction}.
The \textcolor{black}{plan} $\tau$ is defined as an infinite sequence of states i.e., $\tau=\tau(0),\dots,\tau(t)\dots$. In $\tau$, each state $\tau(t)$ is defined as $\tau(t)=[\bbp(t),\bbs(t)]$, 
where $\bbp(t)$ is the multi-robot system state and $\bbs(t)=[s_1(t),\dots,s_N(t)]$, $s_j(t)\in\ccalC$. \textcolor{black}{In other words, $s_j(t)$ determines which skill robot $j$ should apply at time $t$.} 
A plan $\tau=\tau(0),\tau(1),\dots,\tau(t),\dots$, $\tau(t)=[\bbp(t),\bbs(t)]$,
%
satisfies an LTL formula $\phi$ if the word $w=\sigma(0)\sigma(1)\dots \sigma(t)\dots$ where $\sigma(t)=L(\tau(t))$, results in at least one accepting run $\rho_B$.

Combining a feasible plan $\tau$ and its corresponding accepting NBA run $\rho_B$ yields a plan $\tau_H$, where $\tau_H(t)=[\bbp(t), \bbs(t), q_B(t)]$.  If the LTL formula is feasible, then there exists at least one feasible plan $\tau_H$ that can be written in a prefix suffix-structure, i.e., $\tau_H=\tau_H^{\text{pre}}[\tau_H^{\text{suf}}]^{\omega}$; this also implies that there exists a feasible plan $\tau$ in a prefix-suffix structure. The prefix $\tau_H^{\text{pre}}$ is executed first followed by the indefinite execution of the suffix $\tau_H^{\text{suf}}$; in $\tau_H^{\text{suf}}$, $\omega$ stands for indefinite repetition. 
\textcolor{black}{The prefix part is defined as $\tau_H^{\text{pre}}=\tau_H(0),\tau_H(1),\dots, \tau_H(T)$, for some horizon $T\geq 0$, and the suffix part is defined as $\tau_H^{\text{suf}}=\tau_H(T+1),\tau_H(T+2),\dots, \tau_H(T+K)$, for some $K\geq0$, \textcolor{black}{where $q_B(T+1)\in\ccalQ_B^F$}. The suffix part $\tau_H^{\text{suf}}$ is periodic and repeats indefinitely. We define its cycle length $K$ as the length of the shortest contiguous subsequence that, when repeated, generates the entire infinite suffix sequence.
Observe that the prefix part $\tau_H^{\text{pre}}$ allows the robot to reach an accepting NBA state while the suffix part $\tau_H^{\text{suf}}$ allows the robot to revisit that state infinitely often satisfying the NBA accepting condition.}

\vspace{-0.4cm}
\subsection{\textcolor{black}{Violation Cost Function of Multi-Robot Plans}}\label{sec:Vscore}
\vspace{-0.1cm}

Given a plan $\tau_H$, let $q_B'=q_B(t),q_B''=q_B(t+1)$, for $t\geq 0$. 
We denote by $b_{q_B',q_B''}$ the Boolean formula, defined over $\mathcal{AP}$, 
for which it holds that if $\sigma\models b_{q_B',q_B''}$ then $q_B''\in\delta_B(q_B',\sigma)$. Such Boolean formulas can be constructed automatically using existing tools such as \cite{gastin2001fast}. Since $\tau_H$ is a feasible plan, we have that
$\sigma(t)\models b_{q_B',q_B''}$, where $\sigma(t)=L(\tau(t))$, and $\tau(t)=[\bbp(t),\bbs(t)]$. Assume that robot failures occur at time $t$ and, therefore, certain actions in $\bbs(t)$ cannot be applied; this can be represented by setting $s_j(t)=0$ for the affected robots. Consider the case where after the failures, we have that $\sigma(t)\not\models b_{q_B',q_B''}$, i.e., the transition from $q_B'$ to $q_B''$ cannot be enabled anymore.
Let $\mathcal{AP}_b$ collect all predicates that appear in $b_{q_B',q_B''}$ and let $\Sigma_b=2^{\mathcal{AP}_b}$.
There exists at least one $\sigma^* \in \Sigma_b$, such that the concatenation of the symbols $\sigma(t)$ and $\sigma^*$ satisfies $b_{q_B',q_B''}$, i.e., $\sigma(t)\sigma^*\models b_{q_B',q_B''}$\footnote{\textcolor{black}{The concatenation $\sigma(t)\sigma^*$ denotes a symbol in $\Sigma_b = 2^{\mathcal{AP}_b}$ that is generated when both $\sigma(t)$ and $\sigma^*$ are produced simultaneously.}}.
Thus the predicates in $\sigma^*$, if assumed true at time $t$, allow the transition from $q_B'$ to $q_B''$. 
We allow this assumption by taking into account the total penalty for treating $\sigma^*$ as true. Formally, the 
\textit{violation} score of the symbol $\sigma(t)$ over a transition from $q_B'$ to $q_B''$
is defined as 
\begin{equation}\label{eq:edgeviolation}
\mathbb{C}_{\sigma(t)} =  \min_{\forall \sigma^* \in \Sigma_b^*} (\sum_{\pi\in\sigma^*} F(\pi)),   
\end{equation}
where $\Sigma_b^*=\{\sigma\in\Sigma_b~|~\sigma(t)\sigma\models b_{q_B',q_B''} \}$ and $\sigma(t)=L(\tau(t))$. 
Thus the violation score is the lowest possible penalty that we can take to enable this transition; see Ex. \ref{ex:leastViolating}. 

The violation score associated with the execution of a prefix-suffix plan $\tau_H$ after a time instant $t\geq 0$ is the sum of all violation scores for each transition in the plan, i.e.,  
\begin{equation}\label{eq:violation}
\mathbb{C}_{\tau_H}(t) =  \sum_{t'=t}^{T+K}\mathbb{C}_{\sigma(t')}, 
\end{equation}
where $\sigma(t')=L(\tau(t'))$ is the symbol to enable the transition from $q_B(t')$ to $q_B(t'+1)$ as required by $\tau_H$.
%
%

\begin{ex}[Least Violating Transition]\label{ex:leastViolating}
Consider a transition from $q_B'=q_B(t)$ to $q_B''=q_B(t+1)$ at $t\geq0$, where $b_{q_B',q_B''} = (\pi_1\wedge\pi_2) \lor \pi_3$, with penalties $F(\pi_1)=10$, $F(\pi_2)=20$, $F(\pi_3)=50$. \textcolor{black}{Assume that $\sigma(t)=L(\tau(t))=\pi_1$ due to a failure of a robot that was originally responsible for satisfying $\pi_2$.} Then, we have that $\Sigma_b^*=\{\pi_2,\pi_3,\pi_2\pi_3\}$ and $\mathbb{C}_{\textcolor{black}{\sigma(}t\textcolor{black}{)}}=20$ as $\pi_2\in\Sigma_b^*$ has the lowest penalty. \textcolor{black}{Thus, a plan that reaches the final state through the transition from $q_B'$ to $q_B''$ will incur penalty of $20$. Note there may exist alternative ways to reach the final state, without going through this transition, that result in lower or zero cost.}
\end{ex}

\begin{rem}[Violation Cost Function]\label{rem:cost}
   \textcolor{black}{Observe that the cost $\mathbb{C}{\tau_H}(t)$ in \eqref{eq:violation} for a given infinite plan $\tau_H$ depends on the parameters $T$ and $K$ in the prefix-suffix representation of the plan. Since $K$ is consistently designed as discussed earlier, the violation cost of a given plan $\tau_H$ is uniquely determined. Moreover, notice that the cost $\mathbb{C}{\tau_H}(t)$ of any plan $\tau_H$ that does not violate the hard safety constraints (see Assumption \ref{as4}) is bounded and finite, since $\mathbb{C}_{\tau_H}(t)$ is defined as the summation over a finite horizon $T+K$ of penalty terms that are finite by construction (see Definition \ref{def:penalty}).}
\end{rem}

\vspace{-0.4cm}
\subsection{Problem Statement: Reactive Temporal Logic Planning}
\vspace{-0.1cm}
Consider a robot team tasked with completing a mission $\phi$. As the robots execute a designed feasible plan $\tau_H$, certain robot skills may fail unexpectedly; see Ex. \ref{ex:LTL}. In this case, $\tau_H$  may no longer be feasible compromising mission performance. Our goal is to address the following problem: 
\begin{problem}\label{prob_statement}
Consider a mission $\phi$, an initial assignment of predicates to robots, and an offline generated plan $\tau_H$ satisfying $\phi$. 
\textcolor{black}{When failures of robot skills occur (possibly more than one at a time) at time $t$, design a new plan $\tau_H^*$ by re-allocating robots, based on their functioning capabilities, to the predicates $\pi_{\mathcal{T}_c}(j,\ell_e)$ of the form \eqref{eq:pip} associated with robots $j$ that can no longer apply the skill $c$ due to failures. The goal is to design $\tau_H^*$ to minimize the violation score $\mathbb{C}_{\tau_H^*}(t)$ thereby satisfying as much of the LTL mission as possible.}
%
\end{problem}

\begin{ex}\label{ex:LTL}
\textcolor{black}{Consider a team of $4$ robots with skills $c_1$, $c_2$, $c_3$, $c_4$, and $c_5$. The skills are the ability to move, press buttons, retrieve objects, take photos, and open doors, respectively; see Fig. \ref{fig:3robot}.
The skill-based teams are $\ccalT_{c_1}(0)=\{1,2,3,4\}$, $\ccalT_{c_2}(0)=\{1\}$, $\ccalT_{c_3}(0)=\{2,3,4\}$, $\ccalT_{c_4}(0)=\{1,3\}$, and $\ccalT_{c_5}(0)=\{4\}$. Consider an LTL mission: $\phi = \Diamond\pi_1 \wedge (\bar{\pi}_2\wedge\bar{\pi}_3)\bigcup\pi_1 \wedge \Diamond(\pi_4\wedge\pi_5\wedge\pi_6)\wedge\square\bar{\pi}_7$, where 
$\pi_1=\pi_{\ccalT_{c_5}}(4, \ell_4)$, 
$\bar{\pi}_2=\bar{\pi}_{\ccalT_{c_1}} (2, c_1, \ell_4)$,
$\bar{\pi}_3=\bar{\pi}_{\ccalT_{c_1}} (3, c_1, \ell_4)$, 
$\pi_4=\pi_{\ccalT_{c_2}}(1, \ell_1)$, 
$\pi_5=\pi_{\ccalT_{c_3}}(2, \ell_2)$,  
$\pi_6=\pi_{\ccalT_{c_4}}(3, \ell_3)$, and
$\bar{\pi}_7=\bar{\pi}_{\ccalT_{c_5}}(\varnothing, c_1, \ell_2)$. 
The associated penalties are $F(\pi_1)=F(\pi_5)=50, F(\pi_4)=30$, and $F(\pi_6)=15$.
The mission $\phi$ demands that robot $4$ first proceed to location $\ell_4$ to open the door to the room ($\Diamond\pi_1$). Until then, robots $2$ and $3$ cannot enter the room ($(\bar{\pi}_2\wedge\bar{\pi}_3)\bigcup\pi_1$). Then $\phi$ demands robot $1$ to press and hold a button ($\pi_4$), to keep open a trap door, while robot $2$ retrieves an object within the trap door ($\pi_5$), and robot $3$ should take a photo of this sampling action ($\pi_6$) as proof of completion. \textcolor{black}{Note that all three actions need to happen simultaneously, as releasing the button closes the trap door, and capturing the photo after retrieval offers no usable evidence of task success.} 
The dimensions of the environment restricts robot $4$ from coming near $\ell_2$, which is captured by \textcolor{black}{$\bar{\pi}_7$}.
If skill $c_3$ of robot $2$ fails at $t=2$, then a robot $i\in\ccalT_{c_3}(2)=\{3,4\}$ needs to take over $\pi_5$. Notice that even though robot $4$ is not allocated any predicate at that time, assigning it $\pi_5$ is not possible because if robot $4$ satisfies $\pi_5$, it would violate $\bar{\pi}_7$, leading to a violation of $\phi$. As a result, only robot $3$ can take over $\pi_5$. However, note that robot $3$ as well as robot $1$ need to complete tasks at the same time as $\pi_5$ and thus we are forced to abandon one task.} Thus, the problem is to determine a sequence of reassignments that minimizes the mission violation. Our proposed algorithm is designed to handle such challenging scenarios. 
This example is re-visited in Sec. \ref{sec:ResilientPlanning}-\ref{sec:Sim}.
\end{ex}

\begin{rem}[Assumptions \ref{as2}-\ref{as4}]\label{rem:assumptionsPred}
    \textcolor{black}{
   Assumptions \ref{as2}-\ref{as3} are quite common in related deterministic temporal logic planning algorithms; see e.g., \cite{kloetzer2008fully, fainekos2005hybrid, smith2011optimal, chen2012formal, vasile2013sampling, luo2021abstraction, tumova2016multi, shoukry2017linear, kantaros2020stylus, gujarathi2022mt,chen2024fast,kantaros2020reactive,schillinger2016decomposition,chen2024distributed}. 
    Assumption \ref{as2} can be relaxed by tracking the ability of robots to execute multiple skills simultaneously, and incorporating that into the task re-assignment and re-planning process.
   %
   \textcolor{black}{Assumption \ref{as1}  can be relaxed by applying task assignment methods before deployment \cite{schillinger2016decomposition,banks2020multi,Luo2022temporal, Li2023Fast,chen2024distributed,liu2022time,fang2024continuous}}.  
    %
    \textcolor{black}{Assumption \ref{as3} will be used in Section \ref{sec:ResilientPlanning} to independently reallocate sub-tasks associated with failed robots. Relaxing this assumption would require to track dependencies across the predicates during the task reassignment process which is part of our future work.}
    \textcolor{black}{Assumption \ref{as4} is reasonable as it models hard safety constraints that the overall mission cannot tolerate their violation; similar assumptions are made e.g., in \cite{lahijanian2016iterative}.} 
    }
\end{rem}







\vspace{-0.2cm}
\section{Minimum-Violation Temporal Logic Planning}\label{sec:ResilientPlanning}
\vspace{-0.1cm}
In this section, we outline the proposed \textcolor{black}{minimum-violation} algorithm to address Problem \ref{prob_statement}. Our solution comprises  three components. First, we create an offline plan $\tau$ that satisfies $\phi$ using existing LTL planners such as \cite{luo2021abstraction}. \textcolor{black}{Second, we propose a task re-allocation algorithm that reassigns sub-tasks to operational robots, as soon as failures occur. In our approach,} we use the concepts from \cite{kalluraya2023resilient} to set up all the constraints needed for the task reallocation process; see Section \ref{sec:taskRealloc0}. 
Then, in Section \ref{sec:taskRealloc3}, we reason about robot failures and reassign sub-tasks to the remaining functioning robots, while minimizing the total number of sub-task re-allocations. Unlike \cite{kalluraya2023resilient}, the proposed re-allocation algorithm can address cases where feasible task re-assignments cannot be made due to a limited number of available robots. This is achieved by strategically prioritizing assignment of tasks/predicates with high penalty scores (see Definition \ref{def:penalty}).
\textcolor{black}{Third,} given the revised LTL formula according to the reassignments,  we propose a new online re-planning method in Section \ref{sec:onlineReplan} that locally revises the current team plans to accommodate the new task assignments and ensure minimum mission violations as per \eqref{eq:violation}. We note that the re-planner proposed in our earlier work \cite{kalluraya2023resilient} cannot be applied in the considered settings, as it assumes the existence of a feasible plan; if this assumption does not hold, it returns a message that the task is infeasible.

\vspace{-0.3cm}
\subsection{Offline Temporal Logic Planning}\label{sec:samplingAlg}
\vspace{-0.1cm}
Given an LTL task $\phi$, we generate offline a plan $\tau_H$ that optimizes the violation cost function $\mathbb{C}_{\tau_H}(0)$ defined in \eqref{eq:violation}. 
To design $\tau_H$, we use TL-RRT$^*$, a sampling-based planner proposed in \cite{luo2021abstraction}. We select this planner due to its scalability benefits and its abstraction-free and optimality properties. \textcolor{black}{We emphasize that alternative optimal temporal logic planners can be used such as \cite{gujarathi2022mt,vasile2013sampling}}.

\textcolor{black}{In what follows, we provide a brief overview of TL-RRT$^*$. The key idea in \cite{luo2021abstraction} is to build trees incrementally that explore both the multi-robot motion space and the NBA state-space. The nodes of the tree are defined as $\bbq(t)=[\bbp(t), \bbs(t), q_B(t)]$. The root $\bbq(0)$ of the tree is defined based on the initial robot states $\bbp(0)$, a null vector $\bbs(0)$, and an initial NBA state $q_B(0)\in\ccalQ_B^0$. At every iteration of the algorithm, a new state $\bbq(t)$ is sampled and added to the tree if is feasible (i.e., it does not result in violation of $\phi$). 
This sampling-based approach is capable of generating plans, i.e., sequences of states $\bbq(t)$ in a prefix-suffix form $\tau_H=\tau_H^{\text{pre}}[\tau_H^{\text{suf}}]^{\omega}$ as defined in Section \ref{sec:Vscore}. This planner is asymptotically optimal, i.e., as the tree grows, the probability of finding the optimal plan 
goes to one. 
Due to Assumption \ref{as1}, the violation cost of the optimal plan $\tau_H$ with respect to \eqref{eq:violation} must be $\mathbb{C}_{\tau_H}(0)=0$. Among all optimal plans with zero violation cost, we select one with minimum traveled distance required for the execution of the plan; any other optimality metric can be used. }

\vspace{-0.3cm}
\subsection{Setting Up the Online Task Reallocation Process}\label{sec:taskRealloc0}
\vspace{-0.1cm}
\textcolor{black}{Assume that at time $t$, as the robots execute $\tau_H$, a sub-set of robot capabilities fail, resulting in the new vector $\bbZ_j(t)$ for some robots $j\in\ccalR$. Let also $q_B^\text{cur}=q_B(t)$ represent the current NBA state of the robots at time $t$, when executing plan $\tau_H$.  
Our objective is to reassign the affected atomic predicates, each of the form $\pi_{\ccalT_c}(j,\ell_e)$ in \eqref{eq:pip}, to other robots that still possess the required capability $c$, \textcolor{black}{so as to minimize the violation cost of the executed mission} (as defined in \eqref{eq:violation}).
We refer to these predicates as `failed' atomic predicates, a process we informally call `fixing/repairing' of the failed predicates. Our first goal is to re-assign failed predicates to robots with the required function skills; see Alg. \ref{alg:RP}.
A challenge in re-assigning a predicate $\pi_{\ccalT_c}(j,\ell_e)$ is that  all robots $i\in\ccalT_c$ may be occupied with other sub-tasks; see Ex. \ref{ex:LTL}. To repair the failed predicate, our algorithm will trigger a sequence of task reassignments, as fixing the failed predicate requires a robot $i\in\ccalT_c$ to take over, potentially requiring its current sub-task to be reassigned, and so on until all sub-tasks are reassigned or the sub-task with the lowest priority, as per Definition \ref{def:penalty}, is sacrificed by not assigning it to any robot.}

Let $\mathcal{AP}_F\subseteq\mathcal{AP}$ denote the set of failed predicates, i.e., predicates that are no longer satisfiable given the updated capability vectors $\bbZ_j(t)$. The following process is individually and in parallel applied to all failed atomic predicates; as enabled by Assumption \ref{as3} (line \ref{rp:for1}, Alg. \ref{alg:RP}). Given a failed predicate $\pi=\pi_{\ccalT_c}(j,\ell_e)\in\mathcal{AP}_F$, we calculate all NBA states that can be reached from $q_B^\text{cur}$ using a multi-hop plan. We collect these states, including $q_B^\text{cur}$, in a set called $\hat{\ccalQ}_B^{\text{cur}}\subseteq\ccalQ_B$. Let $e=(q_B',q_B'')$ denote an NBA transition from $q_B'$ to $q_B''$, if $\pi$ appears in the corresponding Boolean formula $b_{q_B',q_B''}$ , where $q_B',q_B''\in\hat{\ccalQ}_B^{\text{cur}}$.  \textcolor{black}{Let $\ccalE_{\pi}$} be a set collecting all edges $e$ (line \ref{rp:collect}, Alg. \ref{alg:RP}).
Our aim is to re-assign $\pi$ to a different robot $i\in\ccalT_c$. The key idea  is to inspect all edges $e\in\ccalE_{\pi}$ and re-allocate $\pi$ to a robot. Note that we do not require the robot assigned to undertake $\pi$ in each edge to be the same since we assume independent sub-tasks; see Assumption \ref{as3}. 

Consider a failed predicate $\pi$ and an edge $e=(q_B',q_B'')\in\ccalE_{\pi}$. A necessary condition to preserve the feasibility of the LTL formula after task allocation (see Assumption \ref{as1}) is that all predicates in $b_{q_B',q_B''}$ are assigned to robots so that requirements (i)-(ii) in Assumption \ref{as1} hold locally for $b_{q_B',q_B''}$.
However, there may be cases where this is not possible as certain predicates in $b_{q_B',q_B''}$  may have to remain unassigned due to limited number of available functioning robots. Instead of reporting `assignment failure' (as e.g., in \cite{kalluraya2023resilient}), our goal is to compute a task reassignment that minimizes a violation task re-allocation objective, defined later using the penalty function $F$ introduced in Section \ref{sec:PFLTLMission}. We refer to this as minimum-violation task re-allocation that is described in Section \ref{sec:taskRealloc3}.

\textcolor{black}{To formally define this objective}, we need to introduce the following definitions \textcolor{black}{which are adopted from \cite{kalluraya2023resilient}}. First, we re-write the Boolean $b_{q_B',q_B''}$ in a disjunctive normal form (DNF), i.e., $b_{q_B',q_B''}=\bigvee_{d=1}^Db_{q_B',q_B''}^d$, for some $D>0$. Also, for each Boolean formula $b_{q_B',q_B''}^d$, we define the set $\ccalR_{q_B',q_B''}^d\subseteq\ccalR$ that collects robot indices that appear in $b_{q_B',q_B''}^d$. We further define the set $\mathcal{AP}_i$ that collects all predicates that appear in $b_{q_B',q_B''}^d$ assuming that the ones that are associated with skills $c$ for which $\zeta_c^i(t)=1$ are all assigned to robot $i$. Using $\mathcal{AP}_i$, we construct the alphabet $\Sigma_i=2^{\mathcal{AP}_i}$. For instance, if $b_{q_B',q_B''}^d=\pi_{\ccalT_c}(j,\ell_e)\wedge\pi_{\ccalT_{\hat{c}}}(i,\ell_f)$, then $\mathcal{AP}_i=\{\pi_{\ccalT_c}(i,\ell_e), \pi_{\ccalT_{\hat{c}}}(i,\ell_f) \}$ if $i\in\ccalT_{\hat{c}}\cap\ccalT_c$ 
and $\Sigma_i=\{\pi_{\ccalT_c}(i,\ell_e),\pi_{\ccalT_{\hat{c}}}(i,\ell_f), \pi_{\ccalT_c}(i,\ell_e)\pi_{\ccalT_{\hat{c}}}(i,\ell_f),\epsilon\}$, where $\epsilon$ stands for the empty symbol. 
\textcolor{black}{Using these definitions, we can define the following functions that capture (i) the tasks/predicates that if a robot $i$ undertakes, then $b_{q_B',q_B''}^d$ will become infeasible and (ii) which robots are currently busy with other sub-tasks; see Ex. \ref{ex:LTL} and Fig. \ref{fig:bfs}.}

\begin{definition} [Function $V_{q_B',q_B''}^d$]\label{def:constrainedSet}
\textcolor{black}{The set-valued function $V_{q_B',q_B''}^d:\ccalR\rightarrow \Sigma_i$, given as input a robot index $i\in\ccalR$, returns a set collecting all symbols $\sigma_i\in\Sigma_i$ that if robot $i\in\ccalR$ generates, then $b_{q_B',q_B''}^d$ will be `false' regardless of the values of the other predicates}. 
We define $V_{q_B',q_B''}^d(i)=\emptyset$ for all robots $i\in\ccalR\setminus\ccalR_{q_B',q_B''}^d$. 
\end{definition}


\begin{definition} [Function $g_{q_B',q_B''}^d$]
The function $g_{q_B',q_B''}^d:\ccalR\rightarrow \mathcal{AP}$, given as an input a robot index $i\in\ccalR$, returns a set collecting the atomic predicates that are assigned to robot $i$ in $b_{q_B',q_B''}^d$ excluding the negated ones and the failed predicate.\footnote{There is at most one predicate assigned to a robot $i\in\ccalR_{q_B',q_B''}^d$ as all NBA transitions requiring a robot to satisfy more than one predicate at a time are pruned; see footnote in Section \ref{sec:nba}. 
} We define $g(i)=\varnothing$, for all robots $i\not\in\ccalR_{q_B',q_B''}^d$ and for all robots $i\in\ccalR_{q_B',q_B''}^d$ appearing in negated predicates or in the failed predicate. 
\end{definition}

\begin{ex}[Function $g_{q_B',q_B''}^d$, $V_{q_B',q_B''}^d$ and sets $\ccalR_{q_B',q_B''}^d$]\label{ex:Rg}
Consider the LTL formula given in Ex. \ref{ex:LTL}. We focus on an NBA transition with: $b_{q_B',q_B''}^{d}$$=$$\pi_4\wedge\pi_5\wedge\pi_6\wedge\bar{\pi}_7$, where
$\pi_4$$=$$\pi_{\ccalT_{c_2}}(1, \ell_1)$, $\pi_5$$=$$\pi_{\ccalT_{c_3}}(2, \ell_2)$, $\pi_6$$=$$\pi_{\ccalT_{c_4}}(3, \ell_3)$, and $\bar{\pi}_7$$=$$\bar{\pi}_{\ccalT_{c_5}}(\varnothing,c_1, \ell_2)$.
Then, $\ccalR_{q_B',q_B''}^d$$=$$\{1,2,3\}\cup\ccalT_{c_5}$. Also, $g_{q_B',q_B''}^d(i)$$=$$\varnothing$, for all robots $i\in\ccalR_{q_B',q_B''}^d\setminus\{1,2,3\}$ and  $g(1)$$=$$\pi_4$, $g(2)$$=$$\pi_5$, $g(3)$$=$$\pi_6$. We also have $V_{q_B',q_B''}^d(i)=\emptyset$ for all $i\notin\ccalT_{c_5}$ and $V_{q_B',q_B''}^d(i)=\{\pi_{\ccalT_{c_5}}(i,\ell_2), \pi_{\ccalT_{c_3}}(i,\ell_2), \pi_{\ccalT_{c_5}}(i,\ell_2)\pi_{\ccalT_{c_3}}(i ,\ell_2)\}$ for all $i\in\ccalT_{c_5}\cap\ccalT_{c_3}$. Notice that $\pi_{\ccalT_{c_5}}(i,\ell_2)$ and $\pi_{\ccalT_{c_5}}(i,\ell_2)\pi_{\ccalT_{c_3}}(i,\ell_2)$ are included because of $\bar{\pi}_{\ccalT_{c_5}}(\varnothing,c_1, \ell_2)$ in $b_{q_B',q_B''}^{d}$. Also, $\pi_{\ccalT_{c_3}}(i,\ell_2)$ is included because if robot $4$ satisfies it, then it will be close to location $\ell_2$; therefore, $\bar{\pi}_{\ccalT_{c_5}}(\varnothing,c_1,\ell_2)$ will be violated resulting in violation of  $b_{q_B',q_B''}^{d}$.
\end{ex}

\begin{algorithm}[t]
\caption{Minimum-Violation Task Re-allocation}
\LinesNumbered
\label{alg:RP}
\KwIn{ (i) NBA $\ccalB$, (ii) Current NBA state $q_B^{\text{cur}}$; (iii) Set of failed predicates $\mathcal{AP}_F$}
\KwOut{Revised NBA}

\For{every $\pi\in\mathcal{AP}_F$}{\label{rp:for1}
Define the ordered set of edges $\ccalE_{\pi}$\;\label{rp:collect}
\For{every $e=(q_B',q_B'')\in\mathcal{E}_{\pi}$}{\label{rp:for2}
    Rewrite: $b_{q_B',q_B''}=\bigvee_{d=1}^D b_{q_B',q_B''}^d$\;\label{rp:DNF} 
    \For{$d=1,\dots,D$}{\label{rp:for3}
    Define $\ccalG$ and functions $V^{d}_{q_B',q_B''}, g^{d}_{q_B',q_B''}$\;\label{rp:graph}
    Apply Alg. \ref{alg:bfs} to compute a sequence of re-assignments $p=p(0),\dots,p(P)$\;\label{rp:applyBFS}
    Re-assign atomic predicates as per $p$\;\label{rp:reassign}
    Revise $b_{q_B',q_B''}^d$\;\label{rp:revise}
    }
    }
    }
\end{algorithm}
\setlength{\textfloatsep}{2pt plus 0pt minus 2.0pt}

\vspace{-0.2cm}
\subsection{Minimum-Violation Local Task Reallocation}\label{sec:taskRealloc3}

In this section, we present our proposed minimum-violation task re-allocation algorithm to repair failed predicates. \textcolor{black}{The proposed algorithm augments the one from \cite{kalluraya2023resilient} by enabling it to handle cases where the number of available functioning robots is too limited to ensure that all predicates are assigned to a robot.
}
For each $e\in\ccalE_{\pi}$, we express its respective Boolean formula in a DNF form $b_{q_B',q_B''}=\bigvee_{d=1}^Db_{q_B',q_B''}^d$ (lines \ref{rp:collect}-\ref{rp:DNF}, Alg. \ref{alg:RP}).
Then for each sub-formula $b_{q_B',q_B''}^d$ (lines \ref{rp:for3}-\ref{rp:applyBFS}, Alg. \ref{alg:RP}) we search for task re-assignments over a directed graph $\ccalG$ capturing all possible reassignments in $b_{q_B',q_B''}^d$; see Alg. \ref{alg:bfs}.

This graph is defined as $\ccalG=\{\ccalV_{\ccalG},\ccalE_{\ccalG},w_{\ccalG}\}$, where $\ccalV_{\ccalG},\ccalE_{\ccalG}$, and $w_{\ccalG}$ denote the set of nodes, edges, and a cost function, respectively. The set of nodes is defined as $\ccalV_{\ccalG}=\ccalR$ and a directed edge from node $a$ to $a'\neq a$ exists if $a'\in\ccalT_{c}$, where $c$ is the skill required to satisfy the predicate $g_{q_B',q_B''}^d(a)$. The directed edge indicates that robot $a'$ can take over predicate of $a$ in $b_{q_B',q_B''}^d$.
The cost function $w_{\ccalG}$ assigns cost of $1$ to each edge. 
Note that, our algorithm only needs knowledge of the nodes $a'$ \textcolor{black}{$\in\ccalT_c$} that can be reached in one hop from any node $a$; thus, there is no need to explicitly construct $\ccalG$. 

\begin{algorithm}[t]
\caption{Breadth First Search}
\LinesNumbered
\label{alg:bfs}
\KwIn{ (i) Failed predicate $\pi_{\ccalT_c}(j, c,\ell_e$), 
(ii) $V^{d}_{q_B',q_B''}$, (iii) $g^{d}_{q_B',q_B''}$, (iv) \textcolor{black}{Teams $\ccalT_c(t), ~\forall c\in\ccalC$}}
\KwOut{Path $p$}
$a_{\text{root}}$ = $j$\;\label{bfs:root}
$\mathcal{Q}$ = [$a_{\text{root}}$]\;\label{bfs:queue}
$\texttt{Flag}_{\text{root}}$ = True\;\label{bfs:falgSetTrue}
$a^*$ = $a_{\text{root}}$\;\label{bfs:assign2ndBest}
\While{$\sim$empty($\mathcal{Q}$)}{
    $a\leftarrow$ POP($\mathcal{Q}$)\;\label{bfs:pop}
    \If{$a\in\ccalA$ \& $\sim \texttt{Flag}_{\text{root}}$}{\label{bfs:terminate}
        Using $\texttt{Parent}$ function return path $p$ from $a$\;\label{bfs:return}
    }
    $\texttt{Flag}_{\text{root}}$=False\;\label{bfs:falgSetFalse}
    \For{$a'$ adjacent to $a$ in $\ccalG$}{
        \If{$g_{q_B',q_B''}^d(a)\notin V^{d}_{q_B',q_B''}(a')$ \& $a'$ not explored}
        {\label{bfs:conditions}
            Label $a'$ as explored\;
            $\texttt{Parent}(a') = a$\;\label{bfs:addparent}
            Append $a'$ to $\mathcal{Q}$;\label{bfs:appendToQ}
            
            \If{$F(g_{q_B',q_B''}^d(a'))<F(g_{q_B',q_B''}^d(a^*))$}
            {\label{bfs:penaltyConditions}
                $a^*$ = $a'$ ;\label{bfs:reassign2ndBest}
            }
        }
        
    }
    \If{empty(Q)}{
        Using $\texttt{Parent}$ function return path $p$ from $a^*$\;\label{bfs:penalty_return}
    }
}
\end{algorithm}

\textcolor{black}{We define a set $\ccalA$  collecting all robots that are not involved in the satisfaction of a Boolean formula $b_{q_B',q_B''}^d$, i.e.,
$$\ccalA=\{a\in\ccalR~|~g_{q_B',q_B''}^d(a)=\varnothing\}.$$
Then, the goal of running Algorithm \ref{alg:bfs} (line \ref{rp:applyBFS} of Alg. \ref{alg:RP}), is to find a path in $\ccalG$ from the robot associated with the failed predicate, denoted by $a_{\text{root}}$, to any node $a'\in\ccalA$ subject to `feasibility' constraints, determined by $V_{q_B',q_B''}^d$ that will be defined later. 
%
We define such a path over $\ccalG$ as follows: $$p=p(0),p(1),\dots,p(P),$$ where $p(0)=a_{\text{root}}$, $p(P)=a'$ and $p(k)\notin\ccalA$, for all $k\in\{2,\dots,P-1\}$. 
Such a path will dictate the re-assignment of tasks required to fix the failed predicate while minimizing the penalty in case of a violating solution. 
Specifically, robot $p(k+1)$ assumes the sub-task currently assigned to robot $p(k)$ in $b_{q_B',q_B''}^d$, represented by the atomic predicate $g_{q_B',q_B''}^d(p(k))$. This means that robot $p(k+1)$ relinquishes its current sub-task, which will be taken over by robot $p(k+2)$. 
Thus, this path must adhere to the constraint that 
\begin{equation}\label{eq:constraintRealloc}
    g_{q_B',q_B''}^d(p(k))\notin V_{q_B',q_B''}^d(p(k+1)),~\forall k\in\{2,\dots,P-1\}
\end{equation}
Notice that since $a'=p(P)\in\ccalA$, this means that $g_{q_B',q_B''}^d(p(k))(a')=\varnothing$, i.e., $a'$ is not currently assigned to a task in $b_{q_B',q_B''}^d$.}

\begin{figure}[t]
  \centering
    \includegraphics[width=0.75\linewidth]{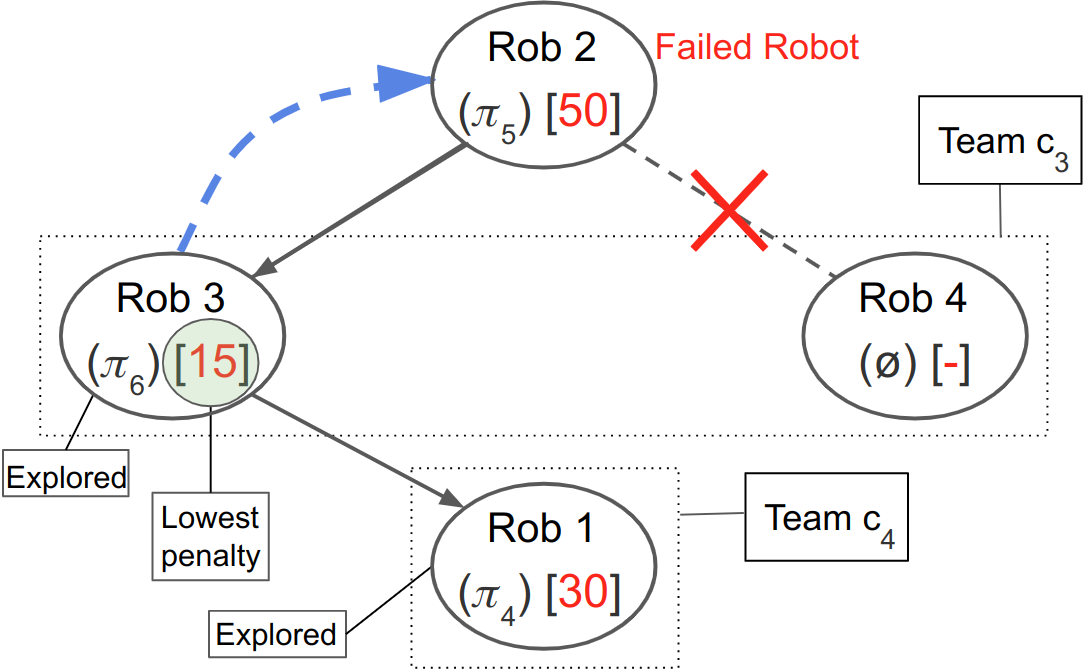}
  \caption{Consider in Example \ref{ex:LTL} the case where skill $c_3$ of robot $2$ fails, i.e., the failed predicate is $\pi_5$. We present the BFS tree (Alg. \ref{alg:bfs}) built to fix $\pi_5$ for the NBA transition enabled by  $b_{q_B',q_B''}^{d}=\pi_4\wedge\pi_5\wedge\pi_6\wedge\bar{\pi}_7$. The set $\ccalA$ is defined as $\ccalA=\{4\}$ and the root of the tree is robot $2$. Robots $3$, and $4$ are adjacent to robot $2$ in $\ccalG$. Robot $4$ is not connected to robot $2$ because it does not satisfy $g_{q_B',q_B''}^d(2)\notin V^{d}_{q_B',q_B''}(4)=\{\pi_5,\pi_7,\pi_5\pi_7\}$. 
  Robot $3$ is connected to robot $2$ and subsequently, robot $1$ is connected to robot $3$. Since we did not find a feasible path from $a_{\text{root}}$ to $\ccalA$, we pick the node $a^*=3$ which has the lowest predicate penalty, $F(g_{q_B',q_B''}^d(a^*))=15$. The blue dashed arrows show the re-assignment process along the computed path $p$, i.e., robot $3$ will take over the failed predicate and $\pi_4$ will be sacrificed resulting in an assignment cost/penalty of 15. In an alternate case if $\pi_4$ had the lowest penalty, then we would have seen robot $1$ replace robot $3$, and robot $3$ replace failed robot $2$, thus sacrificing $\pi_4$. 
  }
  \label{fig:bfs}
\end{figure}

In what follows, we adopt a  Breadth First Search (BFS) inspired approach to find the shortest path $p$ summarized in Alg. \ref{alg:bfs}; see Fig. \ref{fig:bfs}.
%
If such a path does not exist, then the proposed algorithm can compute the path with the minimum violation penalty cost (to be defined later).
We use a queue data structure $\mathcal{Q}$ similar to traditional BFS algorithms. When a node $a$ is removed from $\mathcal{Q}$, then each adjacent node $a'$ is added  to $\mathcal{Q}$ if (1) it has not been explored yet (as in standard BFS) and (2) $g_{q_B',q_B''}^d(a)\notin V^{d}_{q_B',q_B''}(a')$ (line \ref{bfs:conditions}, Alg. \ref{alg:bfs}).
The initial constraint serves to avoid situations where a single robot might have to complete two tasks at the same time, \textcolor{black}{(see Assumption \ref{as2}, and  Assumption \ref{as1}-(ii))}, 
\textcolor{black}{while the second constraint guarantees that the constraints in \eqref{eq:constraintRealloc}} are met, \textcolor{black}{thus avoiding any logical conflicts (see Assumption \ref{as1}-(i))}.
Also, the root node is not initially marked as `explored'. This deliberate choice permits the robot with the failed skill (the root) to undertake other sub-tasks utilizing its remaining (if any) operational skills. 
Simultaneously, every time we add a node to the queue, we update $a^*$ to point to the node with the \textit{lowest penalty} defined as $F(g_{q_B',q_B''}^d(a^*))$ (line \ref{bfs:penaltyConditions}-\ref{bfs:reassign2ndBest}, Alg. \ref{alg:bfs}). 

Finally, we note that the graph-search process is terminated in two ways. The first way is when the first feasible path from $a_{\text{root}}$ to any node in $\ccalA$ is found (line \ref{bfs:terminate}-\ref{bfs:return}, Alg. \ref{alg:bfs}). In this case, the assignment cost is $0$. If such paths cannot be computed (i.e., there is no available robot), \textcolor{black}{then  Alg. \ref{alg:bfs} terminates when it has emptied the $\mathcal{Q}$, meaning it has searched all possible reassignments in $\ccalG$.} 
Then Alg. \ref{alg:bfs} returns a solution
with the smallest assignment cost, i.e., the path $p$ to the
node $a^*$ that has the smallest penalty $F(g_{q_B',q_B''}^d(a^*))$ where the penalty function $F$ is defined in Definition \ref{def:penalty} (line \ref{bfs:penalty_return}, Alg. \ref{alg:bfs}).  
\textcolor{black}{The assignment cost incurred by the path $p$, connecting the root to $a^*$, is defined as
\begin{equation}\label{eq:assignmentCost}
    \mathbb{C}_{q_B',q_B'',d}^{\pi}=F(g_{q_B',q_B''}^d(a^*)).
\end{equation}
In \eqref{eq:assignmentCost}, the subscripts in $\mathbb{C}_{q_B',q_B'',d}^{\pi}$ refer to the failed predicate and the edge $e=(q_B',q_B'')$ that is currently repaired while the subscript $d$ refers to the part of the Boolean formula $b_{q_B',q_B''}$ that is under consideration. Observe that $\mathbb{C}_{q_B',q_B'',d}^{\pi}=0$, if $a^*\in\ccalA$, i.e., $g_{q_B',q_B''}^d(a^*)=\emptyset$; and $\mathbb{C}_{q_B',q_B'',d}^{\pi}>0$, otherwise.}
\footnote{\textcolor{black}{While fixing the predicate in $a_{\text{root}}$, we may sacrifice at most one predicate, by construction of the BFS algorithm.
}}
%
%
Once all failed predicates are fixed with the least amount of violation, the associated formulas $b_{q_B',q_B''}$ are accordingly revised, yielding a new NBA (lines \ref{rp:reassign}-\ref{rp:revise}, Alg. \ref{alg:RP}). 
This revised NBA is an input to an online planner that designs new \textcolor{black}{plans}.

\vspace{-0.1cm}
\subsection{Minimum Violation Online Re-planning}\label{sec:onlineReplan}

\textcolor{black}{Assume that the team state is $\tau_H(t)=[\bbp(t), \bbs(t), q_B(t)]$ when capability failures occurred and that Alg. \ref{alg:RP} has re-assigned tasks to robots. In this section, we discuss how to design a new team plan that satisfies the revised LTL formula. Hereafter, we denote the revised plan by $\tau_H^*=\tau_H^{\text{pre},*},\tau_H^{\text{suf},*}$ to differentiate it from the current plan $\tau_H$.}

\subsubsection{Global Re-planning}
A possible approach to design $\tau_H^*$ is to re-plan globally. Specifically, given the revised NBA, we can use any existing optimal temporal logic planner, such as the one discussed in Section \ref{sec:samplingAlg}, to compute plans satisfying the revised LTL formula starting from the current state $\tau_H(t)$ that minimize the cost function \eqref{eq:violation} while treating any unassigned predicates as `true'.
%
If the total cost of the new plan is zero, it means we have found the plan that completely satisfies the mission.\footnote{Note that 
it is possible that the violation cost of the revised plan is zero even though Alg. \ref{alg:bfs} fails to fix/assign all predicates.
} 
%
%
If the cost is non-zero, it means we have found the least violating plan such that the tasks which will not be completed, have cumulatively the smallest penalty \eqref{eq:violation}.

\subsubsection{Local Re-planning}
%
Nevertheless, global re-planning can be impractical for large robot teams or complex missions and often unnecessary, particularly in cases of a small number of failures. \textcolor{black}{To address this, in our prior work \cite{kalluraya2023resilient}, we presented a local re-planning approach that assumes all predicates are re-assigned after failures and that the LTL task remains feasible. A potential way to relax this assumption would be: (i) to assign a penalty to each sacrificed or un-assigned predicate using the function $F$ defined in Definition \ref{def:penalty}, and (ii) to apply the re-planner from \cite{kalluraya2023resilient} to design a plan that minimizes \eqref{eq:violation}. However, since the re-planner from \cite{kalluraya2023resilient} cannot construct optimal plans, in our setup, it would produce a plan that does not necessarily minimize \eqref{eq:violation}.
}


To address this limitation, we propose a new re-planning approach that locally revises the current plans of those robots affected by failures \textcolor{black}{while also ensuring that the resulting plan is the optimal one with respect to the violation cost function in \eqref{eq:violation}; see the theoretical analysis in Section \ref{sec:analysis}.} Our goal is to determine the local parts of the global plan $\tau_H$ that need to revised. The proposed local re-planner is summarized in Alg. \ref{alg:local}. Our approach has the following key steps and definitions. 
%
%

%
\textit{Sequence $\ccalP$:} Using any simple graph search method over the revised NBA, we can compute a sequence $\ccalP$ of NBA states that consists of two (sub)sequences: (i) a sequence  starting from the current state  $q_B(t)$ and ending in $q_B(T)$, denoted by $\ccalP_{\text{pre}}$; followed by (ii) a sequence of NBA states starting from $q_B(T)$ and ending in $q_B(T)$, denoted by $\ccalP_{\text{suf}}$. \textcolor{black}{Recall from Section \ref{sec:Vscore} that $q_B(T)$ stands for an accepting NBA state associated with the prefix-suffix plan $\tau_H$.}  Thus, we get a sequence $\ccalP$ defined as $\ccalP=\ccalP_{\text{pre}}\ccalP_{\text{suf}}$. During the construction of $\ccalP$, we omit any NBA loops, i.e., $\ccalP_{\text{pre}}(e)\neq \ccalP_{\text{pre}}(e+1)$, for all $e$; the same also holds for $\ccalP_{\text{suf}}$.\footnote{\textcolor{black}{The only exception to this is if the suffix loop only consists of a self-loop of the final accepting state $q_B(T)$. In this case $\ccalP_{\text{suf}}$ will have two elements, $\ccalP_{\text{suf}}(\textcolor{black}{m})=\ccalP_{\text{suf}}(\textcolor{black}{m}+1)=q_B(T)$.} 
} 

\textit{Sequence $\ccalD$:} \textcolor{black}{Next we write the Boolean formulas $b_{q_B',q_B''}$ of all NBA transitions appearing in $P$ in DNF, i.e., $b_{q_B',q_B''}=\bigvee_{d=1}^D b_{q_B',q_B''}^d$. Then, we construct a sequence, denoted by $\ccalD$, of indices $d$, pointing to Boolean subformulas $b_{q_B',q_B''}^d$, along $\ccalP$. Specifically, the $m$-th entry in the sequence $\ccalD$, denoted by $\ccalD(m)$ is associated with the NBA transition from $q_B'=\ccalP(m)$ to $q_B''=\ccalP(m+1)$. The sequence $\ccalD$ requires that the transition from $q_B'$ to $q_B''$ should be enabled by satisfying the sub-formula $b_{q_B',q_B''}^d$, where $d=\ccalD(m)$ (satisfaction of other sub-formulas is irrelevant). The length of a sequence $\ccalD$ is $|\ccalD|=|\ccalP|-1$. Notice that there may be multiple sequences $\ccalD$ for a given sequence $\ccalP$. In this case, we compute all possible sequences $\ccalD$ associated with $\ccalP$. We denote a pair of $\ccalP$ and $\ccalD$ by $(\ccalP,\ccalD)$.\footnote{\textcolor{black}{For instance, consider $\ccalP=q_B',q_B'',q_B'''$. Then, the length of all sequences $\ccalD$ is $|\ccalD|=|\ccalP|-1=2$. Assume that $b_{q_B',q_B''}=\bigvee_{d=1}^2 b_{q_B',q_B''}^d$ and $b_{q_B'',q_B'''}=\bigvee_{d=1}^2 b_{q_B'',q_B'''}^d$. Thus, e.g., $b_{q_B',q_B''}$ can be enabled by satisfying either $b_{q_B',q_B''}^1$ or $b_{q_B',q_B''}^2$. Then, there are four possible sequences $\ccalD$ associated with $\ccalP$: $\ccalD=\{1,1\}$, $\ccalD=\{1,2\}$, $\ccalD=\{2,1\}$, and $\ccalD=\{2,2\}$. For example, the sequence $\ccalD=\{1,2\}$ determines that $b_{q_B',q_B''}$  should be satisfied by satisfying $b_{q_B',q_B''}^1$ (satisfaction of $b_{q_B',q_B''}^2$ is irrelevant) and $b_{q_B'',q_B'''}$ should be satisfied by satisfying $b_{q_B'',q_B'''}^2$.}}} 

\begin{algorithm}[t]
\caption{Re-planning Framework }
\LinesNumbered
\label{alg:local}
\KwIn{ (i) Revised NBA, 
(ii) Current plan $\hat{\tau}_H$}
\KwOut{Revised plan $\tau^*_H$}
Compute all sequences ($\ccalP$, $\ccalD$) and their cost $\mathbb{C}_\ccalP$\;\label{local:CalcP}
Determine $\ccalP^{\tau}$ from current plan $\hat{\tau}_H$\;\label{local:CalcPtau}
\textcolor{black}{Pick $(\ccalP^{\text{min}},\ccalD^{\text{min}})$; where $\ccalP^{\text{min}}=\ccalP_{\text{pre}}^{\text{min}}\ccalP_{\text{suf}}^{\text{min}}$\;\label{local:Pmin}}
Define set $\ccalO$ that collects common (overlap) edges in $\ccalP^{\text{min}}$ and $\ccalP^{\tau}$\;\label{local:CalcO}
Determine $\ccalO^*$ which collects edges in $\ccalO$ that are true overlaps \;\label{local:true}
\If{\textcolor{black}{ $\ccalO^*=\emptyset$}}
{\label{local:globalCond}
    Use Global re-planner to find $\tau^*_H$ \;\label{local:replanGlobal}
    Return $\tau^*_H$\;\label{local:returnGlobal}
    }
$\tau^*_H=[]$ \;\label{local:initTau}
\texttt{Plan\_start} = $\hat{\tau}_H(k)$; where $k= z(\ccalP^{\text{min}}(1))$\;\label{local:initStart}
$\tau^{\text{pre,new}}_H$ = \texttt{Re-planner}(\texttt{Plan\_start}, $\ccalP_{\text{pre}}^{\text{min}}$, $\hat{\tau}_H$)\;\label{local:prefixPlanner}
\texttt{Plan\_start} = $\hat{\tau}_H(\textcolor{black}{\bar{k}})$; where $\textcolor{black}{\bar{k}}= z(\ccalP^{\text{min}}(|\ccalP_{\text{pre}}^{\text{min}}|))$\;\label{local:initSufStart}
$\tau^{\text{suf,new}}_H$=\texttt{Re-planner}(\texttt{Plan\_start}, $\ccalP_{\text{suf}}^{\text{min}}$, $\hat{\tau}_H$)\;\label{local:suffixPlanner}
Return $\tau^*_H = \tau^{\text{pre,new}}_H[\tau^{\text{suf,new}}_H]^{\omega}$\;\label{local:returnPlan}
\end{algorithm}

\textit{Optimal Sequences $\ccalP^{\text{min}}$ and $\ccalD^{\text{min}}$:} \textcolor{black}{Observe that there may exist multiple candidate sequences $\ccalP$ over the revised NBA. In this case, we compute all of them, along with their corresponding, possibly more one, sequences $\ccalD$, and we select the optimal one based on a cost function defined over them.}  We define this cost function using the assignment cost defined in \eqref{eq:assignmentCost}.  To formally define it, consider any pair $(\ccalP,\ccalD)$. Also, consider any two consecutive NBA states in $\ccalP$, i.e., $q_B'=\ccalP(m)$ to $q_B''=\ccalP(m+1)$, for some $m\in\{1,\dots,|\ccalP|-1\}$ in a pair $(\ccalP,\ccalD)$. 
%
%
\textcolor{black}{The transition from $q_B'$ to $q_B''$ will be enabled, as per $\ccalD$, only if the robots generate a symbol $\sigma$ satisfying the Boolean condition $\bigvee_{d=1}^D b_{q_B',q_B''}^d$, where $d=\ccalD(m)$. However, after the reassignment, there may exist predicates in $b_{q_B',q_B''}^d$ with no robots assigned to them.} 
Thus, we define as the cost of the transition from $q_B'=\ccalP(m)$ to $q_B''=\ccalP(m+1)$ the sum of all the assignment costs $\mathbb{C}^{\pi}_{q_B',q_B'',d}$ of any unassigned predicates $\pi$ in the Boolean formula
$b_{q_B',q_B''}^d$, where $d=\ccalD(m)$. To formally define it,
%
let $\mathcal{AP}^U_{q_B',q_B'',d}$ collect any unassigned predicates in $b_{q_B',q_B''}^d$ after the task reallocation. \textcolor{black}{Note that this set will also consist of any predicates that were sacrificed when fixing failures that occurred at past time steps $\bar{t}<t$.} Then the total assignment cost is given by 
\begin{equation}\label{eq:NBAEdgeviolation}
\mathbb{C}_{q_B',q_B'',d} =  \sum_{\pi\in\mathcal{AP}^U_{q_B',q_B'',d}}\mathbb{C}^{\pi}_{q_B',q_B'',d}
\end{equation}
Then, we define the violation score of the entire sequence $(\ccalP,\ccalD)$ as:
%
\begin{equation}\label{eq:NBAviolation}
\mathbb{C}_\ccalP =  \sum_{m=1}^{|\ccalP|-1} \mathbb{C}_{\ccalP(m),\ccalP(m+1),d},
\end{equation}
where $d=\ccalD(m)$ for all $m\in\{1,\dots,|\ccalP|-1\}$.
Among all possible sequences $(\ccalP,\ccalD)$ 
we compute the one that achieves the minimum violation score denoted by $\ccalP^{\text{min}}=\ccalP_{\text{pre}}^{\text{min}}\ccalP_{\text{suf}}^{\text{min}}$ \textcolor{black}{and $\ccalD^{\text{min}}$} (lines \ref{local:CalcP}-\ref{local:Pmin}, Alg. \ref{alg:local}).\footnote{\textcolor{black}{In case there are multiple sequences $\ccalP$ with the same minimum cost, we pick one randomly or based on any user-defined criterion. In our implementation, we pick the sequence that yields the largest set $\ccalO^*$, which will be defined later in the text.
%
}} 

\textit{Sequence $\ccalP^{\tau}$:} Our goal is to find a new multi-robot plan that can generate an NBA run passing through all NBA states in $\ccalP^{\text{min}}$ in the same order they appear while enabling the sub-formulas determined by $\ccalD^{\text{min}}$. 
As it will be shown in Section \ref{sec:analysis}, this plan is the optimal one under mild conditions on the NBA structure. In what follows, we use the sequence $\ccalP^{\text{min}}$ to determine which parts in $\tau_H$ need to be revised to construct $\tau_H^*$. To explain this, we need first to introduce the following definitions for the current plan $\tau_H$. First, we define the plan $\hat{\tau}_H=\tau_H^{\text{pre}}\tau_H^{\text{suf}}=[\bbq(1),\dots,\bbq(T)],[\bbq(T+1),\dots, \bbq(T+K)]$ that concatenates the prefix and the suffix part of the current plan $\tau_H$ (without repeating  $\tau_H^{\text{suf}}$); see also Section \ref{sec:Vscore}.
Second, we compute the state $\hat{\tau}_H(k)$ for which it holds $\hat{\tau}_H(k)=\tau_H(t)$.\footnote{It is possible that $t>T+K$ since the state $\tau_H(t)$ may belong to the suffix part and $\tau_H$ contains an infinite repetition of the suffix part. In $\hat{\tau}_H(k)$, $k$ points to the $k$-th entry in $\hat{\tau}_H$ where   $\hat{\tau}_H(k)=\tau_H(t)$. 
}
Third, we define the sequence $\Upsilon_{\text{pre}}$ collecting all states $\hat{\tau}_H(k')=[\bbp(k'), \bbs(k'), q_B(k')]$ in $\hat{\tau}_H$ that satisfy the following requirements: \textcolor{black}{i) $k' \geq k$; and ii) $q_B(k')\neq q_B(k'-1)$.} 
The states in $\Upsilon_{\text{pre}}$ are projected onto the NBA state space to get a sequence $\ccalP_{\text{pre}}^{\tau}$ of NBA states. Notice that the first state in $\Upsilon_{\text{pre}}$ is $q_B(t)$.  
We denote by $\Upsilon_{\text{pre}}(m)$ and $\ccalP_{\text{pre}}^{\tau}(m)$ the $m$-th entry in $\Upsilon_{\text{pre}}$ and $\ccalP_{\text{pre}}^{\tau}$, respectively. 
We similarly define $\Upsilon_{\text{suf}}$ and $\ccalP_{\text{suf}}^{\tau}$ where the third requirement is replaced by $T+1\leq k'\leq T+K$. 
This way we construct the sequence $\ccalP^{\tau}=\ccalP_{\text{pre}}^{\tau}\ccalP_{\text{suf}}^{\tau}$ (line \ref{local:CalcPtau}, Alg. \ref{alg:local}). \textcolor{black}{We also define a function $z: \ccalP^{\tau}\rightarrow \mathbb{N}$ that takes as an input any state from the sequence $\ccalP^{\tau}$ and returns an index pointing to the corresponding state in 
$\hat{\tau}_H=\tau_H^{\text{pre}}\tau_H^{\text{suf}}$.} 

\textit{Overlap between $\ccalP^{\text{min}}$ and $\ccalP^{\tau}$:} Observe that each pair of consecutive states in  $\ccalP^{\text{min}}$ determines an NBA transition/edge denoted by $(\ccalP^{\text{min}}(m),\ccalP^{\text{min}}(m+1))$. The same holds for $\ccalP^{\tau}$ as well.  
Thus, next we compute the set of NBA edges that  $\ccalP^{\text{min}}$ and $\ccalP^{\tau}$ share. We collect these shared edges in an ordered set $\ccalO$ defined as follows:
\begin{align}
\ccalO = &\left\{ (q_B',q_B'') \mid \exists m, \bar{m}~\text{such that}~\ccalP^{\text{min}}(m)=\ccalP^{\tau}(\bar{m})=q_B'~\right. \nonumber \\ &\text{and}
\left. \ccalP^{\text{min}}(m+1)=\ccalP^{\tau}(\bar{m}+1)=q_B'' \right\}.
\end{align}
\textcolor{black}{In words, the set $\ccalO$ collects all NBA edges $(q_B',q_B'')$ that exist in both $\ccalP^{\text{min}}$ and $\ccalP^{\tau}$  (line \ref{local:CalcO}, Alg. \ref{alg:local}). }
%
%
Hereafter, we informally call the set $\ccalO$ as the overlap between $\ccalP^{\text{min}}$ and $\ccalP^{\tau}$. This set will be used to determine parts of the plan $\tau_H$ that do not require revision.
%
%

\textit{Re-usable Parts of the Plan:}
Let $(q_B',q_B'')$ be any edge in $\ccalO$, where
$\ccalP^{\text{min}}(m)=\ccalP^{\tau}(\bar{m})=q_B'$~\text{and} $\ccalP^{\text{min}}(m+1)=\ccalP^{\tau}(\bar{m}+1)=q_B''$. Consider also the  indices $k_1=z(q_B')$ and $k_2=z(q_B'')$. The part of the plan $\hat{\tau}_H$ starting from the state $\hat{\tau}_H(k_1)$ and ending at the state $\hat{\tau}_H(k_2)$, denoted by $\hat{\tau}_H(k_1:k_2)$, enabled the transition from $q_B'$ to $q_B''$, by construction of $z$ and $\ccalP^{\tau}$, before revising the NBA using Alg. \ref{alg:RP}. However, it may not enable the NBA transition from $q_B'$ to $q_B''$ after revising the NBA. The reason is that the Boolean formulas enabling this NBA transition may have been updated  after running Alg. \ref{alg:RP} and, therefore, it may be associated with different robot-task reassignments. Hereafter, we refer to these sub-plans as non re-usable as they cannot be part of the new optimal plan.

\begin{figure}[t]
  \centering
      \subfigure[Current NBA state sequence $\ccalP_{\text{pre}}^{\tau}$]{
    \label{fig:curNBA}
  \includegraphics[width=0.8\linewidth]{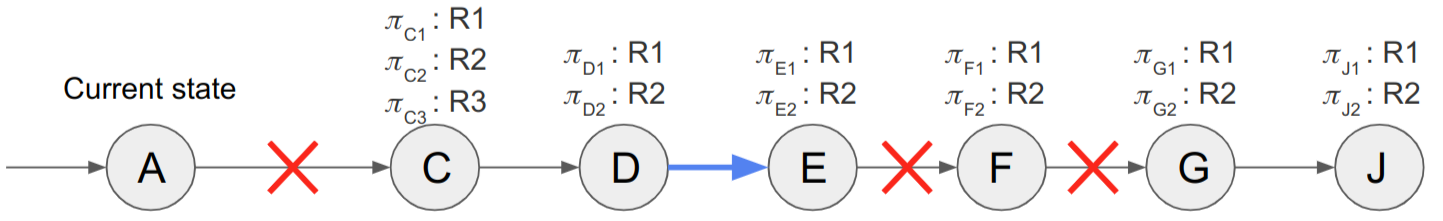}}
  \subfigure[Minimum violating NBA state sequence $\ccalP_{\text{pre}}^{\text{min}}$]{
    \label{fig:revNBA}
  \includegraphics[width=0.8\linewidth]{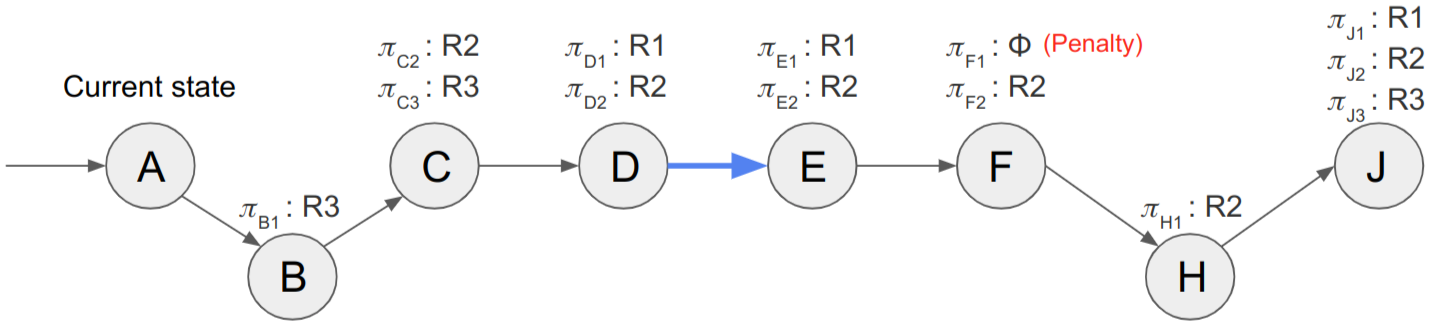}}
  \subfigure[Current Plan]{
    \label{fig:curPath}
  \includegraphics[width=0.8\linewidth]{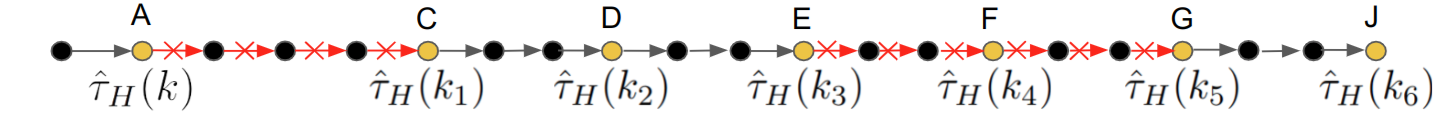}}
  \subfigure[Revised Plan]{
    \label{fig:revPath}
  \includegraphics[width=0.8\linewidth]{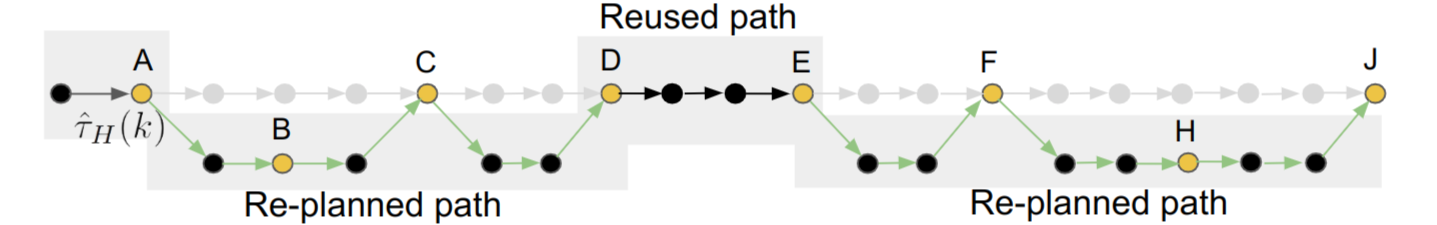}}
  \caption{Hypothetical example of online revision of a part of the prefix plan. NBA states $\text{A,B,\dots,J}\in\ccalQ_{B}$, where $q_B(t)=q_B(k)=\text{A}$. Fig. \ref{fig:curNBA} shows the states in $\ccalP_{\text{pre}}^{\tau}$, along with the predicate-robot assignment that enables the transition to each state (the predicates and robot numbers are for illustrative purposes only). Assume some skills of robot $1$ failed rendering some of the transitions in the NBA infeasible (red cross). Fig. \ref{fig:revNBA} shows the plan $\ccalP_{\text{pre}}^{\text{min}}$ in the NBA that generates the lowest violation cost after fixing the failures. Here the overlap exists from C to F. 
  \textcolor{black}{However, only the transition from D to E will be considered a true overlap. Condition (1) of reusability is satisfied because the most recent predicates done by the robots (R1-$\pi_{\text{D1}}$, R2-$\pi_{\text{D2}}$, and R3-$\pi_{\text{C3}}$) are still the same in both $\ccalP_{\text{pre}}^{\text{min}}$ and $\ccalP_{\text{pre}}^{\tau}$. Condition (2) is satisfied because the boolean $\pi_{\text{E1}}\wedge\pi_{\text{E2}}$ remains unchanged.}
  The disks in Figs. \ref{fig:curPath} and \ref{fig:revPath} capture states in $\hat{\tau}_H$.
  Yellow states $\hat{\tau}_H(k')$ model states for which it holds $q_B(k')\neq q_B(k'-1)$. The part of $\hat{\tau}_H$ connecting  NBA states $q_B'$ and $q_B''$, where $e=(q_B',q_B'')\in\ccalE_{\pi}$ (see Alg. \ref{alg:RP}) is marked with a red color and a red `X' denoting that it requires revision. \textcolor{black}{Note that although the \textcolor{black}{plan} from  $\hat{\tau}_H(k_1)$ (NBA state C) to  $\hat{\tau}_H(k_2)$ (NBA state D) is not marked red, it still needs to be revised as it is not a true overlap.  In this example, the planner re-plans a \textcolor{black}{plan} from current state $\hat{\tau}_H(k)$ (NBA state A) to  $\hat{\tau}_H(k_2)$ (NBA state D) which is the start of a true overlap, and re-plans another \textcolor{black}{plan} from the end of the overlap $\hat{\tau}_H(k_3)$ (NBA state E) to the final state $\hat{\tau}_H(k_6)$ (NBA state J). This is done using lines \ref{re-planner:replanCond}-\ref{re-planner:replanAppend} in Alg. \ref{alg:re-planner}. The planner does not plan for the true overlap section (NBA states D to E) and instead re-uses the original plan $\hat{\tau}_H(k_2:k_3)$ to bridge the state $\hat{\tau}_H(k_2)$ to the state $\hat{\tau}_H(k_3)$. In Alg. \ref{alg:re-planner}, this is done by lines \ref{re-planner:reuseCond}-\ref{re-planner:reuseAppend}. }  
  }
  \label{fig:revise}
\end{figure}

\textit{Conditions for Re-usable Plans:} 
Next, we define two conditions under which sub-plans $\hat{\tau}_H(k_1:k_2)$ associated with edges $(q_B',q_B'')\in\ccalO$ are re-usable; \textcolor{black}{see also Fig. \ref{fig:revise}.} 
Let $\bar{q}_B$ be the NBA state preceding $q_B'$ in $\ccalP^{\text{min}}$, i.e., $\bar{q}_B=\ccalP^{\text{min}}(m-1)$. Notice that the transition $(\bar{q}_B,q_B')$ may not exist in $\ccalO$. 
\textcolor{black}{\textbf{Condition (1)}: The first condition to reuse the plan $\hat{\tau}_H(k_1:k_2)$ is to ensure that the new plan (not designed yet) can reach $\hat{\tau}_H(k_1)$ and can activate the transition from $\ccalP_{\text{min}}(m-1)$ to $\ccalP_{\text{min}}(m)$ as soon as $\hat{\tau}_H(k_1)$ is reached.}
In what follows, we provide the condition that, if satisfied, the above is ensured. To do so, we check sequentially every Boolean formula required to enable all transitions from $\ccalP^{\text{min}}(m')$ to $\ccalP^{\text{min}}(m'+1)$, starting from $m'=1$ until $m'+1=m$ to track the most recent predicate assigned to each robot. We do the same with $\ccalP^{\tau}$ to compute the most recent predicate assigned to each robot while enabling the sequence of NBA transitions in $\ccalP^{\tau}$ until the NBA state $q_B'$ is reached.  Thus if the most recent predicate for a robot, when computed using $\ccalP^{\tau}$ and $\ccalP^{\text{min}}$ are the same, for all robots, it means that we can ensure that the robots can reach the position at the start of that overlap as determined by $\hat{\tau}_H(k_1)$. 
\textbf{Condition (2)}: 
\textcolor{black}{Additionally, we can reuse $\hat{\tau}_H(k_1:k_2)$ only if $\hat{\tau}_H(k_1:k_2)$ enables the transition from $q_B'$ to $q_B''$ in the revised NBA. This means that each state in the plan $\hat{\tau}_H(k_1:k_2-2)$ satisfies $b_{q_B',q_B'}^d$ and $\hat{\tau}_H(k_2-1)$ satisfies  $b_{q_B',q_B''}^d$.}
If there are unassigned predicates in these Boolean formulas, then the corresponding parts of $\hat{\tau}_H(k_1:k_2)$ can satisfy  $b_{q_B',q_B'}^d$ and $b_{q_B',q_B''}^d$ while allowing the unassigned predicates to be considered as true. With slight abuse of notation, we denote by $\hat{b}_{q_B',q_B'}^d$ and $\hat{b}_{q_B',q_B''}^d$ the Boolean formula in which the unassigned predicates have been replace by logic true. 
\textcolor{black}{
Thus, formally, condition (2) is satisfied if $L([\bbp(k),\bbs(k)])\models \hat{b}_{q_B',q_B'}^d; \forall k\in\{k_1,\dots,k_2-2\}$, and $L([\bbp(k_2-1),\bbs(k_2-1)])\models \hat{b}_{q_B',q_B''}^d$, where 
$\hat{\tau}_H(k)=[\bbp(k),\bbs(k),q_B]$.} 

\textit{True Overlap between $\ccalP^{\text{min}}$ and $\ccalP^{\tau}$:} We collect all NBA edges $(q_B',q_B'')\in\ccalO$ associated with re-usable sub-plans $\hat{\tau}_H(k_1:k_2)$ in an \textcolor{black}{ordered} set denoted by $\ccalO^*\subseteq \ccalO$ (line \ref{local:true}, Alg. \ref{alg:local}). Hereafter, we refer to the set $\ccalO^*$ as the ``true" overlap between $\ccalP^{\tau}$ and $\ccalP^{\text{min}}$.
Furthermore, we call a state $q_B'$ as the start of a true overlap if $(\bar{q}_B,q_B')\notin\ccalO^*\text{ and }(q_B',q_B'') \in \ccalO^*$. Similarly, a state $q_B'$ is the end of a true overlap if $(\bar{q}_B,q_B')\in\ccalO^*\text{ and }(q_B',q_B'') \notin \ccalO^*$. \textcolor{black}{Also, for brevity, we refer to the $m$-th element in $\ccalO^*$ as the $m$-th true overlap.}

\begin{algorithm}[t]
\caption{Function $\texttt{Re-planner}$}
\LinesNumbered
\label{alg:re-planner}
\KwIn{ (i) \texttt{Plan\_start},
(ii) Sequence of NBA states $\bar{\ccalP}$,
(iii) Current plan $\hat{\tau}_H$}
\KwOut{New (prefix or suffix) plan $\tau^{\text{new}}_H$}
$\tau^{\text{new}}_H$=[]\;\label{re-planner:initPlan}
$\texttt{Flag}_{\text{end}}$  = False\;\label{re-planner:initFlags}
\For{$i=1\textcolor{black}{,}\dots\textcolor{black}{,} |\bar{\ccalP}|$}{\label{re-planner:iterate}
    \If{$i=|\bar{\ccalP}|$ \& $q_B^i=\bar{\ccalP}(i)$ is NOT end of true overlap}
    {\label{re-planner:flagend}
         $\texttt{Flag}_{\text{end}}$ = True\;\label{re-planner:flagSetTrue} 
    }
    \If{$q_B^i=\bar{\ccalP}(i)$ is start of true overlap \textbf{or} $\texttt{Flag}_{\text{end}}$}{\label{re-planner:replanCond}
        $k_{\textcolor{black}{1}}= z(\bar{\ccalP}(i))$\;\label{re-planner:getStart}
        Replan path from \texttt{Plan\_start} to $\hat{\tau}_H(k_{\textcolor{black}{1}})$\;\label{re-planner:replan}
        Append path to $\tau^{\text{new}}_H$\;\label{re-planner:replanAppend}
        \texttt{Reuse\_start} = $\hat{\tau}_H(k_{\textcolor{black}{1}})$\;\label{re-planner:initReuse}
    }
    \If{$q_B^i=\bar{\ccalP}(i)$ is end of true overlap}{\label{re-planner:reuseCond}
        $k_{\textcolor{black}{2}}= z(\bar{\ccalP}(i))$\;\label{re-planner:getEnd}
        Reuse path from \texttt{Reuse\_start} to $\hat{\tau}_H(k_{\textcolor{black}{2}})$\;\label{re-planner:reuse}
        Append path to $\tau^{\text{new}}_H$\;\label{re-planner:reuseAppend}
        \texttt{Plan\_start} = $\hat{\tau}_H(k_{\textcolor{black}{2}})$\;\label{re-planner:initReplan}
    }
}
Return $\tau^{\text{new}}_H$\;\label{re-planner:returnPlan}
\end{algorithm}

\textit{Overview of Local Re-planning:} 
If $|\ccalO^*|> 0$, we design the new plan by locally revising the current plan $\hat{\tau}_H$ (lines \ref{local:initTau}-\ref{local:returnPlan}, Alg. \ref{alg:local}). \textcolor{black}{First, we $\texttt{Plan\_start}$ is initialized to point to the state in the plan $\hat{\tau}_H$ from where we start the local re-planning process. In the beginning, this points to the current team state at time $t$ i.e., to $\hat{\tau}_H(k)$ where $k= z(\ccalP^{\text{min}}(1))$ (line \ref{local:initStart}, Alg. \ref{alg:local}). }\footnote{\textcolor{black}{Note that $z$ takes as input states from the sequence $\ccalP^{\tau}$. The state $\ccalP^{\text{min}}(1)$ is the same as the state 
$\ccalP^{\tau}(1)$ by construction of these sequences; thus $z(\ccalP^{\text{min}}(1))$ is well defined. Also, $z(\ccalP_{\text{pre}}^{\text{min}}(i))$ and $z(\ccalP_{\text{suf}}^{\text{min}}(i))$ used in lines \ref{re-planner:getStart} and \ref{re-planner:getEnd} in Alg. \ref{alg:re-planner} are well defined since these states exist in $\ccalP^{\tau}$ too by construction of the set $\ccalO^*$.}}
%
We then design the prefix \textcolor{black}{plan} (lines \ref{local:initStart}-\ref{local:prefixPlanner}, Alg. \ref{alg:local}) \textcolor{black}{followed by the construction of the suffix \textcolor{black}{plan} (lines \ref{local:initSufStart}-\ref{local:suffixPlanner}, Alg. \ref{alg:local}).}

The construction of the prefix and suffix \textcolor{black}{plan} is described in Alg. \ref{alg:re-planner}. 
Specifically, to construct the prefix part,
\textcolor{black}{first, we initialize an empty plan $\tau^{\text{new}}_H$, \textcolor{black}{modeling the revised prefix plan that will constructed by composing the re-usable sub-plan with new sub-plans replacing the non-reusable parts of the previous team plan. We also initialize} a flag variable, \textcolor{black}{denoted by $\texttt{Flag}_{\text{end}}$},
to indicate the last iteration of the loop, used in the algorithm (lines \ref{re-planner:initPlan}-\ref{re-planner:initFlags}, Alg. \ref{alg:re-planner}). We then iterate over each state in $\ccalP_{\text{pre}}^{\text{min}}$ (line \ref{re-planner:iterate}, Alg. \ref{alg:re-planner})
till we reach the start of the first true overlap in $\ccalO^*$.} 
Then we design a plan from the current state $\hat{\tau}_H(k)$ (as pointed to by $\texttt{Plan\_start}$) to the state $\hat{\tau}_H(k_1)$, where $k_1=z(q_B')$ points to the start of the first true overlap in $\ccalO^*$ (line \ref{re-planner:replanCond}-\ref{re-planner:replan}, Alg. \ref{alg:re-planner}). \textcolor{black}{This plan is appended to the new plan $\tau^{\text{new}}_H$.
Then, we initialize the variable $\texttt{Reuse\_start}$ to \textcolor{black}{point to} the start of the \textcolor{black}{first} true overlap (line \ref{re-planner:replanAppend}-\ref{re-planner:initReuse}, Alg. \ref{alg:re-planner}).
Next, we continue iterating over $\ccalP_{\text{pre}}^{\text{min}}$ till we reach the end of \textcolor{black}{the first} true overlap (line \ref{re-planner:reuseCond}, Alg. \ref{alg:re-planner}).}
We then reuse the original plan $\hat{\tau}_H(k_1:k_2)$ to bridge the state $\hat{\tau}_H(k_1)$, as pointed to by $\texttt{Reuse\_start}$, to the state $\hat{\tau}_H(k_2)$ where $k_2$ points to the end of the first true overlap (line \ref{re-planner:getEnd}-\ref{re-planner:reuse}, Alg. \ref{alg:re-planner}).
 \textcolor{black}{This reused plan is then appended to the new plan $\tau^{\text{new}}_H$, and we re-initialize $\texttt{Plan\_start}$ to point to the end of the true overlap (line \ref{re-planner:reuseAppend}-\ref{re-planner:initReplan}, Alg. \ref{alg:re-planner}).}
We repeat these steps for each true overlap sequentially. 
\textcolor{black}{
In the last iteration, if the last state in $\ccalP_{\text{pre}}^{\text{min}}$ is the end of a true overlap, then we reuse \textcolor{black}{the part of previous plan connecting the end of the previous true overlap (i.e., \texttt{Reuse\_Start}) up to end of the current true overlap (i.e., $\tau^{\text{new}}_H(k_2)$, where $k_2=\ccalP_{\text{pre}}^{\text{min}}(|\ccalP_{\text{pre}}^{\text{min}}|)$); see lines \ref{re-planner:getEnd}-\ref{re-planner:initReplan}}.
Then Alg. \ref{alg:re-planner} terminates. However, if it is not the end of a true overlap, then we must replan the \textcolor{black}{plan} from the end of the previous true overlap. Thus we set the flag $\texttt{Flag}_{\text{end}}$ to true (lines \ref{re-planner:flagend} and \ref{re-planner:flagSetTrue}, in Alg. \ref{alg:re-planner}), 
allowing us to satisfy the condition in line \ref{re-planner:replanCond}, and replan the last part and append it to $\tau^{\text{new}}_H$.}
\textcolor{black}{Once the prefix \textcolor{black}{plan} is constructed, we repeat the same steps to generate the suffix \textcolor{black}{plan} (line \ref{local:suffixPlanner}, Alg. \ref{alg:local})}. \textcolor{black}{We start building the suffix from where the prefix part ended, as indicated by $\texttt{Plan\_start} = \hat{\tau}_H(\textcolor{black}{\bar{k}})$; where $\textcolor{black}{\bar{k}}= z(\ccalP^{\text{min}}(|\ccalP_{\text{pre}}^{\text{min}}|))$ (line \ref{local:initSufStart}, Alg. \ref{alg:local}). Note that the only difference in Alg. \ref{alg:re-planner} is that instead of giving $\ccalP_{\text{pre}}^{\text{min}}$ as input, we give $\ccalP_{\text{suf}}^{\text{min}}$}. Lastly the prefix and suffix plans are combined and returned as the final plan $\tau_H^*$ (line \ref{local:returnPlan}, Alg. \ref{alg:local}). \textcolor{black}{An example illustrating the re-planning process is provided in Fig. \ref{fig:revise}.} 

\textit{Local Plan Synthesis:} \textcolor{black}{In what follows we discuss how we design the local plans connecting the end of a true overlap  $\hat{\tau}_H(A)$, to the start of the next true overlap $\hat{\tau}_H(B)$. 
To design this plan, we use the sampling-based planner \cite{luo2021abstraction} discussed in Section \ref{sec:samplingAlg}.\footnote{We emphasize again that any other optimal planner can be used. For instance, \cite{gujarathi2022mt} can be used too, after discretizing the environment, which can return the optimal plan in finite time.} The plan begins at $\hat{\tau}_H(A)$ and targets the specific goal state $\hat{\tau}_H(B)$. During the construction of the tree, we impose the following restrictions. 
(i) First, we restrict intermediate transitions to pass only through the NBA states appearing in the optimal prefix $\ccalP^{\text{min}}$, and do so in the order they appear in $\ccalP^{\text{min}}$. This structure is critical for proving the optimality of Algorithm \ref{alg:local} in Section \ref{sec:analysis}.
(ii) Second, we reject all sampled states that violate the hard safety constraints modeled by negations in the LTL formula; see Assumption \ref{as4}. This also ensures that the resulting plan will have a finite violation cost; see Rem. \ref{rem:cost}. 
}
(iii) Third, we treat any unassigned predicates as `true' \textcolor{black}{(which is what may result in a \textcolor{black}{plan} with non-zero violation cost)}.

\textit{Switching to Global Planning:} 
\textcolor{black}{Given $\ccalP^{\text{min}}$, if $|\ccalO^*|=0$ or if there does not exist a plan connecting true overlaps, we repeat the above process for another optimal sequence $\ccalP^{\text{min}}$.}
%
%
If an alternative plan $\ccalP^{\text{min}}$ does not exist, then we trigger global re-planning. In this case, we employ the sampling-based planner \cite{luo2021abstraction}, discussed earlier, so that the initial state is $\tau_H(t)$. 
Global re-planning is subject only to the restrictions (ii) and (iii) mentioned above. Thus, the new globally designed plan is allowed to deviate from $\ccalP^{\text{min}}$. For simplicity of presentation, Alg. \ref{alg:local} does not show the case where alternative sequences $\ccalP^{\text{min}}$ can be considered. 

\vspace{-0.2cm}
\section{Algorithm Analysis}\label{sec:analysis}
\textcolor{black}{In this section, we discuss optimality properties of the proposed algorithm. 
First, in Section \ref{sec:theory1}, we discuss the optimality of task re-allocation process presented in Alg. \ref{alg:RP}-\ref{alg:bfs}. Then, in Section \ref{sec:theory2}, we discuss the completeness and optimality of the proposed re-planning algorithm, summarized in Alg. \ref{alg:local}, given re-assignments of the failed tasks. \textcolor{black}{We emphasize that these results hold for any existing temporal logic planner that can be used for local/global plan synthesis, as long as it is complete and optimal \cite{luo2021abstraction,gujarathi2022mt,kantaros2020stylus}.} Finally, in Section \ref{sec:theory3}, we combine these to results to provide optimality guarantees of the joint task re-allocation and re-planning framework.}

\vspace{-0.1cm}

\subsection{Optimality of Task Re-Allocation Due to Failures}\label{sec:theory1}

The following propositions provide optimality guarantees of the proposed task re-allocation algorithms, described in Alg. \ref{alg:RP}-\ref{alg:bfs}, with respect to the assignment cost \eqref{eq:assignmentCost}. All proofs can be found in Appendix \ref{sec:app1}.

\begin{prop}[Optimality of Alg. \ref{alg:bfs}]\label{prop:opt1}
Consider a failed predicate $\pi\in\mathcal{AP}_F$ and Boolean formula $b_{q_B',q_B''}^d$ that contains $\pi$. 
Alg. \ref{alg:bfs} is optimal in the sense that it will find the re-allocation, determined by a path $p$, \textcolor{black}{with the lowest cost $\mathbb{C}_{q_B',q_B'',d}^{\pi}$, defined in \eqref{eq:assignmentCost}}. 
Among all re-allocations with the lowest penalty, Alg. \ref{alg:bfs} selects the one with the minimum number of re-assignments.
%
\end{prop}

\begin{prop}[Optimality of Alg. \ref{alg:RP}]\label{prop:opt2}
\textcolor{black}{Consider a failed predicate $\pi\in\mathcal{AP}_F$ and a set of NBA edges $\ccalE_{\pi}$. Alg. \ref{alg:RP} will compute  re-allocations (i.e., paths $p$) of all failed predicates for all Boolean formulas $b_{q_B',q_B''}^d$ for a given edge $e\in\ccalE$, and for all $e\in\ccalE$, with the lowest penalty as per \eqref{eq:assignmentCost}. It will also find the solution with the minimum number of re-assignments across all sub-formulas and edges.}
%
\end{prop}

\begin{rem}[Set $\mathcal{AP}_F$]\label{rem:AP_F}
\textcolor{black}{Observe that $\mathcal{AP}_F$ collects only the predicates failed at the current time step $t$ and not any predicates that were pleft unassigned from fixing a previous failure at time $t' < t$. This is because Alg. \ref{alg:RP} sacrificed these predicates at time $t'$ for having the lowest penalty; see Prop. \ref{prop:opt1}. Thus, including them in $\mathcal{AP}_F$ would be redundant, as Alg. \ref{alg:RP} would not find any predicates with a lower penalty to sacrifice, leaving them unassigned once again.}
\end{rem}

\subsection{Optimality of Re-planning Due to Failures}\label{sec:theory2}

\begin{figure}[t]
  \centering
    \includegraphics[width=0.75\linewidth]{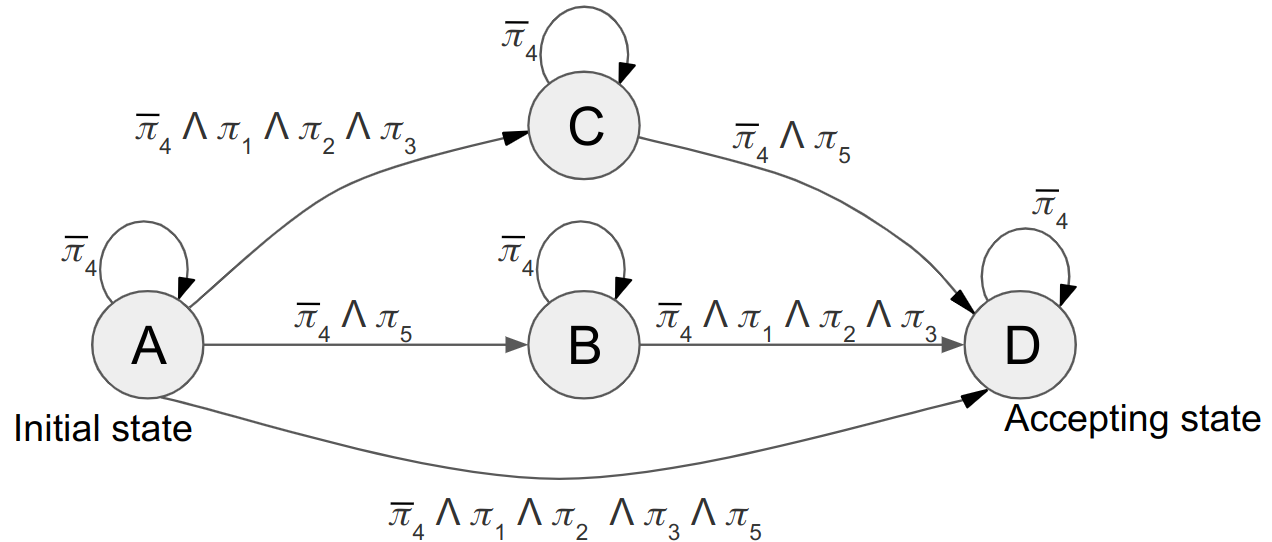}
  \caption{NBA generated by mission in Example \ref{ex:LTL}; $\phi = \Diamond(\pi_1\wedge\pi_2\wedge\pi_3)\wedge\square\Bar{\pi}_4\wedge\Diamond\pi_5$. Assumption \ref{cond:NBA_self_loop} is satisfied, since all self-loops are $b_{q_B',q_B'}=\xi_i$, where $\xi_i$ is a Boolean formula defined over atomic predicates of the form \eqref{eq:negpip}.
  }
  \label{fig:NBAexample}
\end{figure}

\textcolor{black}{Next, in Proposition \ref{prop:optiLVP} \textcolor{black}{and in Corollaries \ref{cor:optCost}-\ref{cor:GlobalOptCost}}, we present the optimality properties of the replaning algorithm presented in Section \ref{sec:onlineReplan}. Specifically, we show that our revised plan $\tau_H^*$ is optimal with respect to \eqref{eq:violation}, given the re-assignments performed by Algorithms \ref{alg:RP}-\ref{alg:bfs}, due to failures, as required in Problem \ref{prob_statement}.} The proofs of these results can be found in Appendix \ref{sec:app2}. 
To state our results, we need first to introduce the following assumption; see also Rem. \ref{rem:limitations} and Fig. \ref{fig:NBAexample}. 

\begin{assumption}\label{cond:NBA_self_loop}
    Assume that the assigned LTL task corresponds to an NBA where self-loops at NBA states $q_B$ either do not exist or the Boolean conditions enabling them are either always true (i.e., $b_{q_B,q_B}=1$) or defined only over predicates of the form \eqref{eq:negpip}  with no negations in front of them (as also required by Assumption \ref{as4}). 
\end{assumption}

%

\begin{prop}[\textcolor{black}{Completeness} \& Optimality of Alg. \ref{alg:local}]\label{prop:optiLVP}
Suppose that failures of robot capabilities occur at time step $t$. Given sub-task re-assignments, generated by Alg. \ref{alg:RP}-\ref{alg:bfs} (or any other approach), the proposed re-planner, presented in Alg. \ref{alg:local}, is guaranteed to find a multi-robot plan $\tau_H^*$, if it exists, that minimizes \eqref{eq:violation}, under Assumption \ref{cond:NBA_self_loop}.\footnote{\textcolor{black}{A multi-robot plan $\tau_H^*$ will not exist if  the mission $\phi$ is infeasible even after replacing unassigned predicates with the logical `true'.}} 
\end{prop}

\textcolor{black}{In the following corollary, we provide conditions, under which the cost of the optimal plan, generated by the local replanner is equal to $\mathbb{C}_{\ccalP^{\text{min}}}$ (see \eqref{eq:NBAviolation}) for any $\ccalP^{\text{min}}$.} 


\begin{cor}[Optimal Cost of Locally Re-planned plans]\label{cor:optCost}
    \textcolor{black}{Assume that there exists 
    a plan $\tau_H^*$ that minimizes \eqref{eq:violation}. Any plan $\tau_H^*$ computed by Alg. \ref{alg:local}, without triggering global replanning, is an optimal plan with respect to \eqref{eq:violation} with cost equal to $\mathbb{C}_{\ccalP^{\text{min}}}$ defined in \eqref{eq:NBAviolation} for any $\ccalP^{\text{min}}$. 
    }
\end{cor}

\begin{cor}[Optimal Cost of Globally Re-planned plans]\label{cor:GlobalOptCost}
    \textcolor{black}{
    If Alg. \ref{alg:local} triggers global re-planning, and the global re-planner finds a plan with violation cost $\mathbb{C}_{\ccalP^{\text{min}}}$ (see  \eqref{eq:NBAviolation}) for any $\ccalP_{\text{min}}$, then that plan is an optimal plan $\tau_H^*$. If there does not exist a plan that can enable the sequence of NBA transitions determined by $\ccalP_{\text{min}}$, then the optimal plan will have a violation cost that is larger than $\mathbb{C}_{\ccalP^{\text{min}}}$. 
    } 
\end{cor}


\begin{rem}[Assumption \ref{cond:NBA_self_loop}]\label{rem:limitations}
\textcolor{black}{
If an NBA satisfies Assumption \ref{cond:NBA_self_loop}, then failed predicates will never appear in the self-loops. 
Assumption \ref{cond:NBA_self_loop} will be violated if, for instance, the LTL task includes requirements of the form $\square \pi_{\ccalT_c}(j,c,\ell_e)$ requiring continuous satisfaction of $\pi_{\ccalT_c}$. The reason is that this would yield Boolean conditions $b_{q_B,q_B}$ of the form $b_{q_B,q_B}=\pi_{\ccalT_c}\wedge\xi$, for some Boolean formula $\xi$, for all $q_B\in\ccalQ_B$. However, safety requirements of the form $\square \bar{\pi}_{\ccalT_{\hat{c}}}(j,c,\ell_e)$ do not violate it. 
In case, Assumption \ref{cond:NBA_self_loop} is violated, Algorithm \ref{sec:onlineReplan} may not necessarily compute an optimal plan since self-loops are omitted when the paths $\ccalP$ are constructed in Section \ref{sec:onlineReplan}. 
A potential approach to design optimal plans even if Assumption \ref{cond:NBA_self_loop} does not hold is to construct the paths $\ccalP$ while accounting for the number of times self-loops need to be activated in order to reach a final state. However, this would significantly increase the number of possible paths $\ccalP$ which would, consequently, increase the computational cost of our re-planning method. 
}
\end{rem}

\vspace{-0.2cm}
\subsection{Joint Optimality of Task Re-Allocation and Re-planning}\label{sec:theory3}

\textcolor{black}{Proposition \ref{prop:optiLVP} demonstrated optimality of the revised plan \textit{given} the re-assignments made by Alg. \ref{alg:RP}-\ref{alg:bfs}. In what follows, we demonstrate the optimality of the joint task reallocation and re-planning method. 
Specifically, Proposition \ref{prop:optiMVP} establishes that, under certain conditions, 
there does not exist alternative re-assignment of predicates which would result in a better revised plan $\tau_H^*$ (with respect to \eqref{eq:violation}).
} 
The proof can be found in Appendix \ref{sec:app3}. 

\begin{prop}[Joint Optimality]\label{prop:optiMVP}
\textcolor{black}{Consider an LTL formula corresponding to a NBA satisfying Assumption \ref{cond:NBA_self_loop}. 
Suppose that failures in robot capabilities occur at time step $t$. Let $p_{\pi}$ denote the path (i.e., the sequence of re-assignments) generated by Alg. \ref{alg:RP}-\ref{alg:bfs} to repair a failed predicate $\pi\in\mathcal{AP}_F$. Let $\hat{p}_{\pi}\neq p_{\pi}$ denote any other path in the tree constructed by Alg. \ref{alg:bfs} assuming the algorithm was allowed to terminate only after exhaustively exploring all possible reassignments. Assume that the following two conditions hold for all paths $\hat{p}_{\pi}$ (including $p_{\pi}$) and all failed predicates $\pi\in\mathcal{AP}_F$.
(i) First, the online re-planner, described in Alg. \ref{alg:local} can compute a plan $\tau_H^*$ without triggering global re-planning. (ii) Second, the sequence $(\ccalP^{\text{min}},\ccalD^{\text{min}})$ used to construct $\tau_H^*$ satisfies
\begin{equation}\label{condII}
    d(m)=\argmin_{\bar{d}\in\{1,\dots,D\}} \mathbb{C}_{q_B',q_B'',\bar{d}},
\end{equation}
where $q_B'=\ccalP^{\text{min}}(m)$ and $q_B''=\ccalP^{\text{min}}(m+1)$, for all $m\in\{1,\dots,|\ccalP^{\text{min}}|-1\}$. 
In \eqref{condII}, $\mathbb{C}_{q_B',q_B'',\bar{d}}$ refers to the total assignment cost associated with the Boolean sub-formula $b_{q_B',q_B''}^{\bar{d}}$ defined in \eqref{eq:proofAssignmentD}. 
Under assumptions (i) and (ii), no alternative reassignment of the predicates, different from the one generated by Alg. \ref{alg:RP}-\ref{alg:bfs}, 
could result in another plan $\hat{\tau}_H^*$, where $\hat{\tau}_H^* \neq \tau_H^*$, produced by 
Alg. \ref{alg:local} (or any other optimal temporal logic planner) such that $\mathbb{C}{\tau_H^*} > \mathbb{C}_{\hat{\tau}_H^*}$.
}
%
%
\end{prop}


\begin{rem}[Prop. \ref{prop:optiMVP} - Assumption  (ii)]
    \textcolor{black}{Assumption (ii) in Prop. \ref{prop:optiMVP} requires the optimal plan to generate symbols enabling NBA transitions from $q_B'$ to $q_B''$ by satisfying the Boolean subformula $b_{q_B',q_B''}^{\bar{d}}$ with the minimum assignment/violation cost $\mathbb{C}_{q_B',q_B'',\bar{d}}$. Informally, this assumption holds if the environment does not prevent the robots from generating such symbols (e.g., all regions are accessible to the robots) and the NBA transitions are independent. By independence, we mean that the symbol $\sigma$ that may be selected by any planner to enable the transition $(q_B', q_B'')$ does not impose any constraints on the symbol $\bar{\sigma}$ that can be selected to enable another NBA transition  $(\bar{q}_B, \bar{q}_B')$ and vice versa. Essentially, the symbol selected to enable any transition $(q_B', q_B'')$  does not depend on the sequence of symbols generated to reach $q_B$ from an initial NBA state. In our simulations, we empirically tested that all the considered case studies satisfy both assumptions made in Prop. \ref{prop:optiMVP}.
    }
\end{rem}

\vspace{-0.2cm}
\section{Experimental Validation} \label{sec:Sim}
\vspace{-0.1cm}
\textcolor{black}{In this section, we present multiple experiments demonstrating the performance of our algorithm in the presence of unexpected failures. \textcolor{black}{Our main evaluation metrics include runtimes for task re-allocation and re-planning as well as the violation cost of the designed plans. First, in Section \ref{sec:setupReplanner} we provide the configuration of the sampling-based planner \cite{luo2021abstraction}. }
In Section \ref{sec:sim3robot}, we report the performance of our algorithm for the scenario discussed in Ex. \ref{ex:LTL} that considers a small team of $4$ robots. \textcolor{black}{In Section \ref{sec:largeTeams} 
we consider a larger team of 24 robots, demonstrating how the algorithm's performance is affected by the number of failures, the timing of those failures and obstacles in the environment. 
We compare the computational efficiency of both the local and global re-planners by examining how the number of failures and obstacles in the environment affect re-planning runtimes. 
Additionally, we compare the performance of the reassignment algorithms (Alg. \ref{alg:RP}, Alg. \ref{alg:bfs}) against a baseline method.}
\textcolor{black}{In Sections \ref{sec:sim3robot}-\ref{sec:largeTeams}, we consider ground robots with simple holonomic dynamics} 
(more details in Section \ref{sec:setupReplanner}). In Section \ref{sec:aerialGazebo}, we consider a Gazebo simulation that involve teams of drones operating in a simulated city. The experiments in Section \ref{sec:sim3robot},\ref{sec:largeTeams} were carried out using Python3 on a computer with Intel Core i7-8565U 1.8GHz and 16Gb RAM while the ones in Section \ref{sec:aerialGazebo} have been conducted on Gazebo (ROS, python3) on a computer with Intel Core i5-8350U 1.7GHz and 16Gb RAM.
%
\textcolor{black}{Finally, in Section \ref{sec:hardware}, we provide hardware experiments demonstrating the real-time performance of our algorithm on a team of drones.} Videos for experiments can be found in \cite{SimResilientViolation}.} 

\vspace{-0.2cm}
\subsection{Setting Up the Re-planning Framework}\label{sec:setupReplanner}

As discussed in Section \ref{sec:onlineReplan}, we use the sampling-based planner presented in \cite{luo2021abstraction} to set up \textcolor{black}{Alg. \ref{alg:local},} \textcolor{black}{since it meets the completeness and optimality requirements discussed in Section \ref{sec:theory2}};  any other complete/optimal temporal logic planner can be used. In what follows, we discuss how we have implemented its steering function, the sampling strategy, and the termination criterion of \cite{luo2021abstraction}.

\textit{Steer Function:} \textcolor{black}{In Sections \ref{sec:sim3robot}-\ref{sec:largeTeams}, we consider the following holonomic robot dynamics:
\begin{equation}\label{eq:nonlinRbt}
\begin{bmatrix}
   p_{j}^1(t+1) \\
  p_{j}^2(t+1)
  \end{bmatrix}=  
  \begin{bmatrix}
   p_{j}^1(t) \\
  p_{j}^2(t)
  \end{bmatrix}+ 
        \begin{bmatrix}
   \tau u \cos(\theta) \\
   \tau u\sin(\theta)
     \end{bmatrix}\\   
\end{equation}
where the robot state $\bbp_j(t)=[p_j^1(t),p_j^2(t)]^T$ captures the position of robot $j$, $\tau$ is the sampling period, and the control inputs $u$ and $\theta$ represent linear velocity, and orientation respectively. }
\textcolor{black}{To implement the `steer' operation of \cite{luo2021abstraction}, we consider a finite set of motion primitives defined as: $u\in\{0,8\}\text{m/s}$ 
and $\theta\in\set{0,\pm 1,\pm 2,\dots, \pm 180}\text{degree}$. 
Thus, to extend the tree towards a new sample, we pick the primitive that drives the system state (determined by the parent tree node of the new sample) closer to the new sample; see \cite{luo2021abstraction} for more details.}

\textit{Sampling Strategy:} We use the biased sampling strategies developed in \cite{luo2021abstraction,Kantaros2022perception}. \textcolor{black}{Specifically, in the local-replanner mode of Alg. \ref{alg:local}, to connect the end of a true overlap to the start of the next true overlap, instead of sampling states uniformly, we bias the sampling process of the planner along the shortest plans that lead to the regions of interest that will enable a sequence of NBA transitions, determined by $\ccalP^{\text{min}}$, leading to the start of the next true overlap. Similarly, in the global-replanner mode of  of Alg. \ref{alg:local}, we bias the sampling strategy towards designing plans that will enable the NBA transitions determined by $\ccalP^{\text{min}}$. A formal presentation of the biased sampling process can be found in \cite{luo2021abstraction,Kantaros2022perception}.}
%
%

\textit{Termination Criterion:} \textcolor{black}{We terminate the sampling-based planner used during the local-replanning mode of Alg. \ref{alg:local}}, when a sub-plan connecting the end of a true overlap to the start of the next true overlap is found. This is because all other plans (if any) that may be computed by the local planner will share the same violation cost; see Cor. \ref{cor:optCost}. \textcolor{black}{Similarly, we terminate the sampling-based planner used during the global-replanning mode as soon as it returns a plan \textcolor{black}{with cost equal to $\mathbb{C}_{\ccalP^{\text{min}}}$,} as this plan is optimal with respect to \ref{eq:violation}; see Cor. \ref{cor:GlobalOptCost}. \textcolor{black}{If the local or global planner does not terminate under the above conditions, then we stop the sampling-based planner after $25,000$ iterations.}\footnote{\textcolor{black}{We note that the employed sampling strategies implicitly also attempt to minimize the total traveled distance as well even though this is not part of our main objective function \eqref{eq:violation} \cite{luo2021abstraction}.}}}

\vspace{-0.4cm}
\subsection{Procuring Samples Task - Single Failure}\label{sec:sim3robot}
\textcolor{black}{We revisit Example \ref{ex:LTL}. The LTL mission corresponds to an NBA with $5$ states.
Due to failure of the retrieval mechanism in robot $2$, $\pi_5$, which was originally assigned to robot $2$, cannot be satisfied. At this point, Alg. \ref{alg:RP} is called to fix all NBA edges associated with $\pi_5$. The total number of affected edges is $5$. In all affected edges, \textcolor{black}{all robots have assigned tasks. Thus, Alg. \ref{alg:bfs}} decides to sacrifice $\pi_6$ since it has the lowest penalty (of 15); see Fig. \ref{fig:bfs}.
The resulting trajectories of the robots can be seen in Fig. \ref{fig:3rb}, where robot $3$ takes over robot $2$'s sampling task at $\ell_2$. 
The time needed for this reassignment is $0.0007$ secs and the new plans are generated in $0.16$ secs. 
The new plans were re-planned globally because there were no reusable plans. 
Informally, this occurred because the robot-task assignments in the Boolean formulas associated these NBA transitions got revised. The violation cost of the plan is $15$.}

\vspace{-0.3cm}
\subsection{Multiple Failures in Large Teams - Performance Analysis}\label{sec:largeTeams}

\textcolor{black}{
We examine a team of $N=24$ ground robots situated in a factory post-disaster, where they must execute sequentially a series of tasks to restore control. 
The robots together possess $|\ccalC|=6$ skills associated with mobility, valve shutdown, fire extinguishing, sample collection, RGB photo capturing, and thermal imaging.
The mission is expressed as: $\phi=\Diamond(\xi_{1}\wedge\Diamond(\xi_{2}\wedge\Diamond\xi_{3}))\wedge\Diamond(\xi_{4}\wedge\Diamond(\xi_{5}\wedge\Diamond\xi_{6}))$,
where each $\xi_i$ is a Boolean formula defined using atomic predicates of the form $\pi_{\ccalT_c}(j,\ell_e)$ for $6$ to $9$ robots. 
For example, 
$\xi_2 =  \pi_{\ccalT_{c_3}}(3, \ell_{24})\wedge\pi_{\ccalT_{c_3}}(4, \ell_{18})\wedge\pi_{\ccalT_{c_5}}(7, \ell_{44})\wedge\pi_{\ccalT_{c_3}}(13, \ell_{9})\wedge\pi_{\ccalT_{c_3}}(16, \ell_{26})\wedge\pi_{\ccalT_{c_2}}(17, \ell_{2})\wedge\pi_{\ccalT_{c_3}}(21, \ell_{14})$.}
\textcolor{black}{Each predicate has an associated penalty ranging from $3$ to $18$. 
This formula corresponds to an NBA featuring $25$ states.}

\subsubsection{Number of failures vs violation cost}
\textcolor{black}{First, we examine the effect of number of robot failures on the violation cost \eqref{eq:violation} of the plan. 
The number of failures in each run is $[1,3,6,12,16,20,22]$ and we have three sets of these runs where the failures occur at different times $t=[2,23,34]$. 
The corresponding violation cost \textcolor{black}{of the plans designed by Alg. \ref{alg:local}} 
for each of the runs is given in Table I. \textcolor{black}{Notice that these are the optimal costs because they are equal to the corresponding cost $\ccalC_{P_\text{min}}$, defined in \eqref{eq:NBAviolation} due to Cor. \ref{cor:optCost}-\ref{cor:GlobalOptCost}.} As expected, fewer failures can be fixed without sacrificing any tasks, leading to a zero violation cost. However, as the number of failures increases, more tasks are sacrificed, resulting in higher violation scores.  
Furthermore, if failures happen at later stages, the violation scores are lower as a majority of the tasks would have been completed.}

\begin{table}[htbp]
\centering
\label{tab:penalties}
\caption{Violation cost of plan}
\begin{tabular}{|c|c|c|c|c|c|c|c|}
\hline
\textbf{Num. of Failures} & \textbf{1} & \textbf{3} & \textbf{6} & \textbf{12} & \textbf{16} & \textbf{20} & \textbf{22} \\
\hline
$t = 3$  & 0   & 0   & 0   & 0   & 41  & 221 & 243 \\
$t = 23$ & 0   & 0   & 0   & 0   & 13  & 172 & 220 \\
$t = 34$ & 0   & 0   & 0   & 0   & 8   & 124 & 165 \\
\hline
\end{tabular}
\end{table}

\subsubsection{Number of failures vs re-planning time}
\begin{figure}[t]
  \centering
\includegraphics[width=0.85\linewidth]{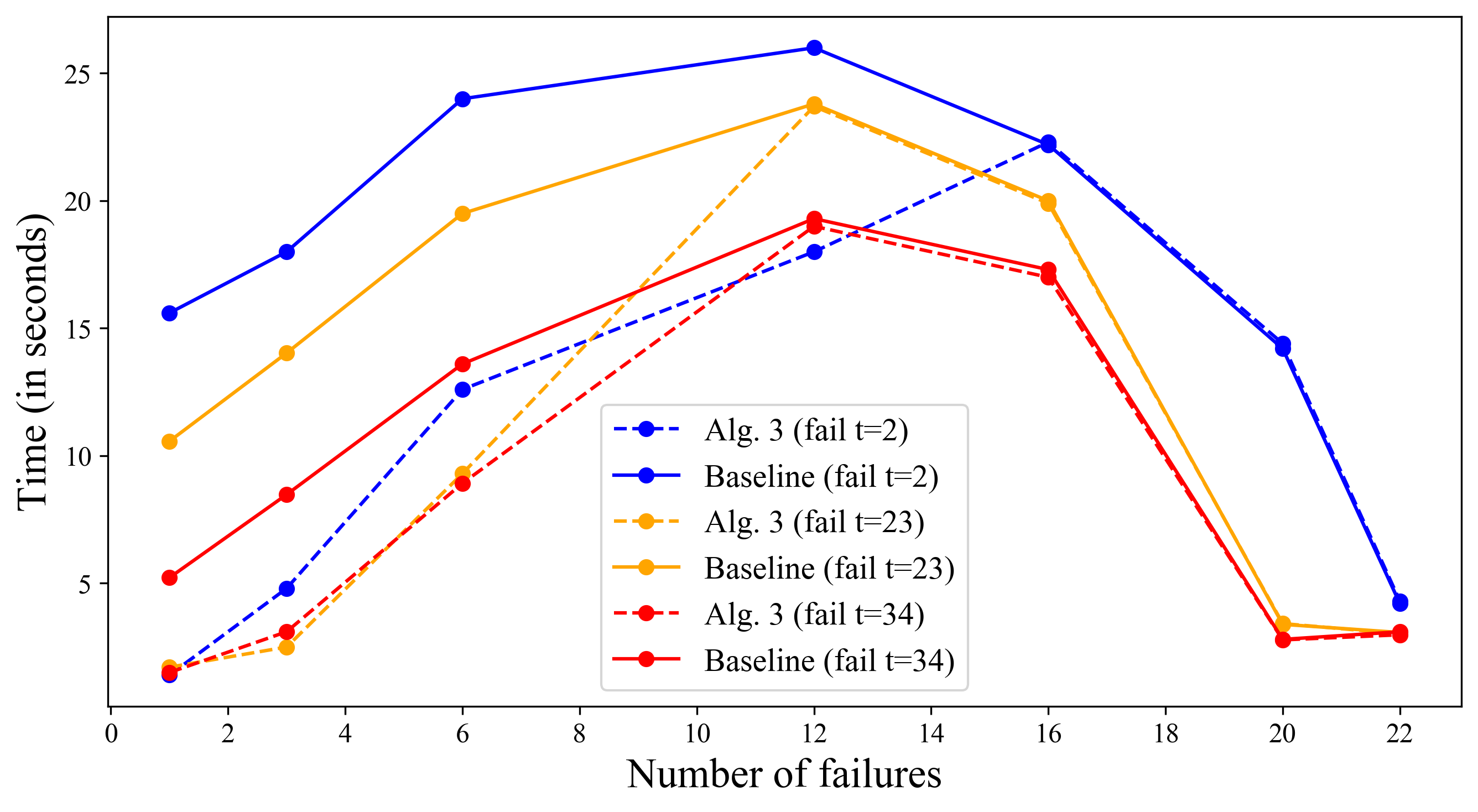}\vspace{-0.4cm}
\caption{\textcolor{black}{Effect of number of failures on re-planning time: Fewer failures generally means fewer plan changes. Thus while a global replanner will need to replan everything, the local replanner can fix only the necessary changes and generate plans faster. However as number of failures increase, the plan needs to be revised at multiple transitions and the local planner would end up planning globally resulting in times similar to the global planner.}}
\vspace{-0.1cm}
\label{fig:failVtime}
\end{figure}

\textcolor{black}{ Second, we examine the effect of number of robot failures on the re-planning times using exactly the same setup as before in terms of the number of failures and the time steps these failures occur.
Specifically, we  compare Alg. \ref{alg:local} against a baseline that always globally replan plans using \cite{luo2021abstraction}. To ensure that our comparisons are fair, we set up the baseline exactly as we set up the global re-planner of Alg. \ref{alg:local}; see Section \ref{sec:setupReplanner}.} We report the post-failure re-planning times for Alg. \ref{alg:local} (dashed lines) and the baseline (solid lines) in Fig. \ref{fig:failVtime}. These times are the average times of $5$ runs for each setting. 
We observe that Alg. \ref{alg:local} is more efficient when the numbers of failures are few. This is because if there are fewer failures, the number of \textcolor{black}{edges that need to be fixed in the NBA could be fewer, resulting in a larger true overlap $\ccalO^*$, and, therefore, allowing the planner to reuse plans (local-replanning mode)}. However, as the number of robot failures increases (around $12$ failures in this setup), the number of true overlaps will become $0$, and Alg. \ref{alg:local} switches to global re-planning. \textcolor{black}{As a result, the runtimes of Alg. \ref{alg:local} and the baseline} may become comparable for large number of failures as also shown in Fig. \ref{fig:failVtime}; see also Rem. \ref{rem:parallel}. 
Also, notice in Fig. \ref{fig:failVtime} that the runtimes of both \textcolor{black}{the baseline and Alg. \ref{alg:local}} first increase and then reduce as the number of failures increases. 
\textcolor{black}{Intuitively, this happens because initially as the number of failures increase, the number of \textcolor{black}{NBA} edges that needs to \textcolor{black}{be repaired increases. This leads to smaller sets $\ccalO^*$, i.e., in fewer reusable plans, which in turn results in longer re-planning runtimes.} However, when the number of failures exceeds a threshold (i.e., $12$ for this specific setup), the re-allocation method starts sacrificing sub-tasks (i.e., not assigning predicates to robots) which combined with fewer robots alive to plan for, results in shorter re-planning times.} 
\textcolor{black}{We also see that, if the failures occur at later time steps, the re-planning times are smaller as the planner needs to re-plan for a shorter horizon.} 

\begin{figure}[t]
  \centering
    \subfigure[Environment with 3 obstacles]{
        \label{fig:3obs}
        \includegraphics[width=0.48\linewidth]{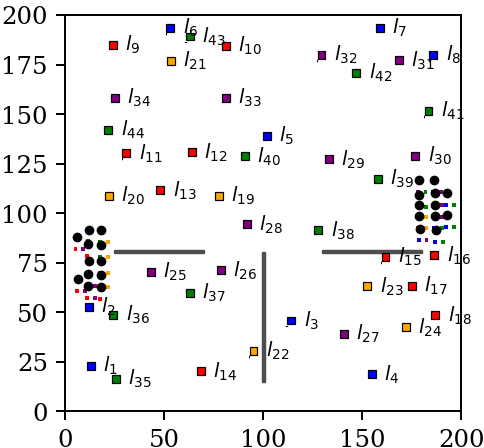}}
    \subfigure[Environment with 12 obstacles]{     
        \label{fig:12obs}
        \includegraphics[width=0.48\linewidth]{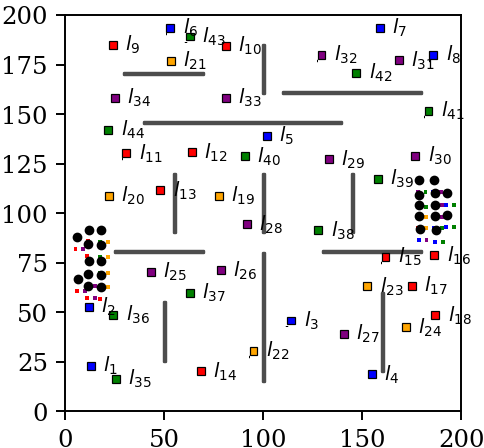}}\vspace{-0.3cm}
\caption{The two figures illustrate the different environments used to assess the effect of number of obstacles (gray walls) on planning times. The starting positions of the 24 robots (black discs) are shown, along with the 44 targets (colored squares; color represents skill needed at that location) that may need to be visited as part of the mission. Intuitively, the environment with 12 obstacles would result in longer plans, requiring more time for planning.}
\label{fig:obstacles}
\end{figure}

\begin{rem}[Local re-planning scenarios]\label{rem:localScenarios}
    Observe from Sec. \ref{sec:sim3robot}- \ref{sec:largeTeams} that Alg. \ref{alg:local} cannot always revise plans locally. 
    \textcolor{black}{In general, the chances of requiring global replanning increase as the number of failed predicates increases, as in this case the number of NBA edges that need to be fixed increase as well which may yield an empty set $\ccalO^*$. This can happen when a large number of robot failures occur or even when a small number of robot failures occur but the corresponding robots were assigned to a large number of predicates in the formula.} 
\end{rem}

\subsubsection{Number of obstacles vs re-planning time}
\begin{figure}[t]
  \centering
\includegraphics[width=0.8\linewidth]{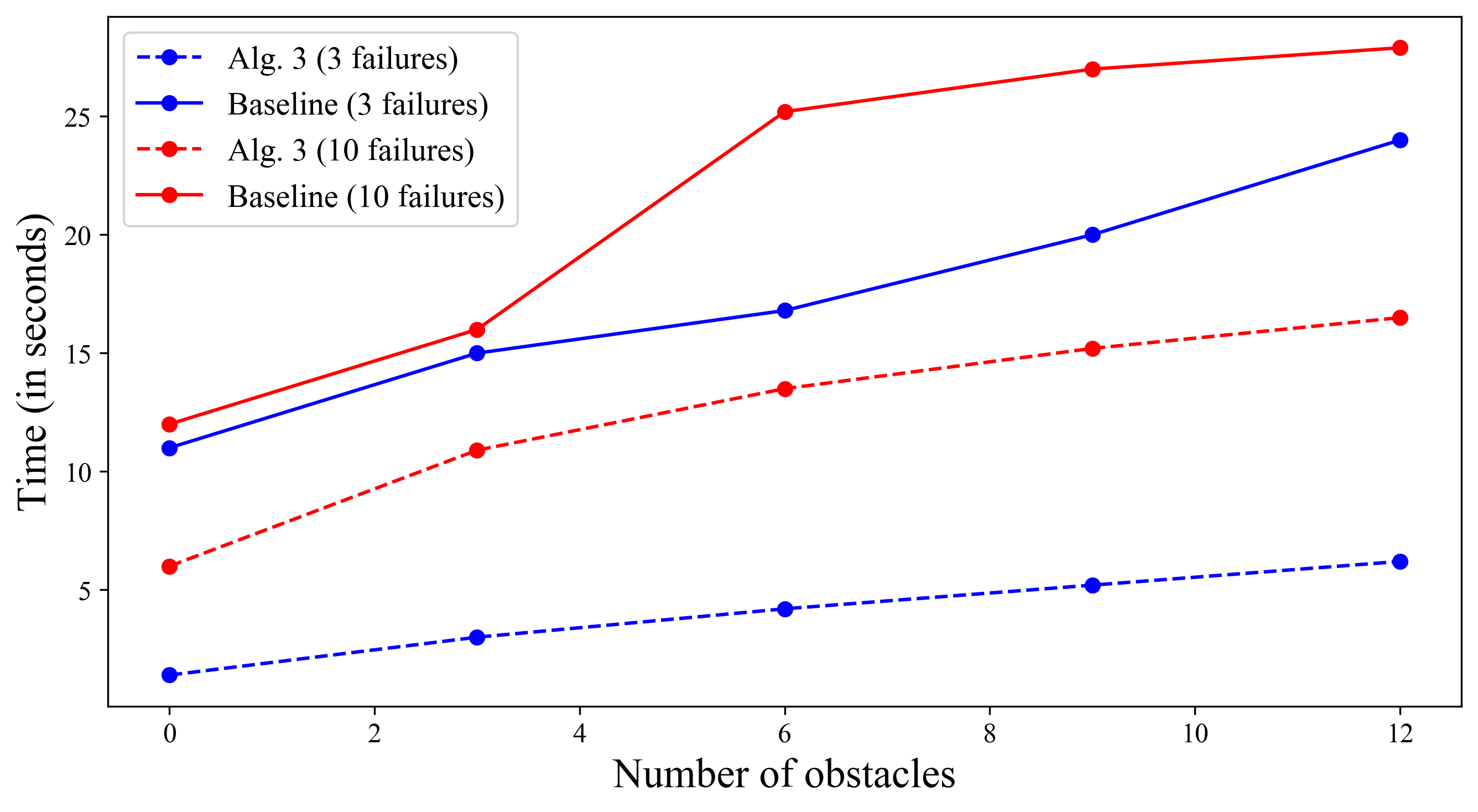}\vspace{-0.4cm}
\caption{Effect of number of obstacles on re-planning time: As the number of obstacles increases the time needed for re-planning increases in all cases. This is because more obstacles results in a cluttered environment forcing the robots to take longer plans to reach their target. The increase in re-planning times with respect to obstacles may not necessarily be linear because depending on the location of the robots taking over a failed task and their destinations, the new obstacles may or may not affect the new plans.}
\label{fig:obsVtime}\vspace{-0.1cm}
\end{figure}

Third, we evaluate the impact of the number of obstacles in the environment on re-planning runtimes. Specifically, we vary the number of obstacles that the robots must navigate around  considering cases with $0, 3, 6, 9$, and $12$ obstacles.
For each scenario, we report the re-planning runtimes of Alg. \ref{alg:local} and the baseline, under two conditions: (1) when 3 robots fail at $t=3$, and (2) when 10 robots fail at $t=3$. The comparative results are reported in Figure \ref{fig:obsVtime}. We observe that with increasing number of obstacles time needed for re-planning for both methods keeps increasing. However, in all cases, the Alg. \ref{alg:local} is more time efficient than the baseline.
\vspace{-0.3cm}

\textcolor{black}{
\subsubsection{Performance of Re-assignment Algorithm vs Baseline}
Fourth, we compare the performance of our reallocation algorithm against a global reallocation baseline that reassigns \textit{all} predicates appearing in each edge $e \in \ccalE_{\pi}$. The baseline iterates over all edges $e \in \ccalE_{\pi}$ that contain at least one failed predicate and uses the Hungarian algorithm to reassign all predicates in the corresponding Boolean formulas $b_{q_B', q_B''}^d$ to the surviving robots, regardless of whether they were affected by failure.
If there are fewer robots available than tasks, the algorithm prioritizes assignment of higher penalty tasks. 
While both the baseline and our approach are optimal, resulting in identical total violation costs for a given transition \eqref{eq:assignmentCost}, the baseline reassigns all sub-tasks, whereas our method minimizes the number of reassignments as shown in Propositions \ref{prop:opt1}-\ref{prop:opt2}.}

\begin{figure}[ht]
    \centering
    \includegraphics[width=0.4\textwidth]{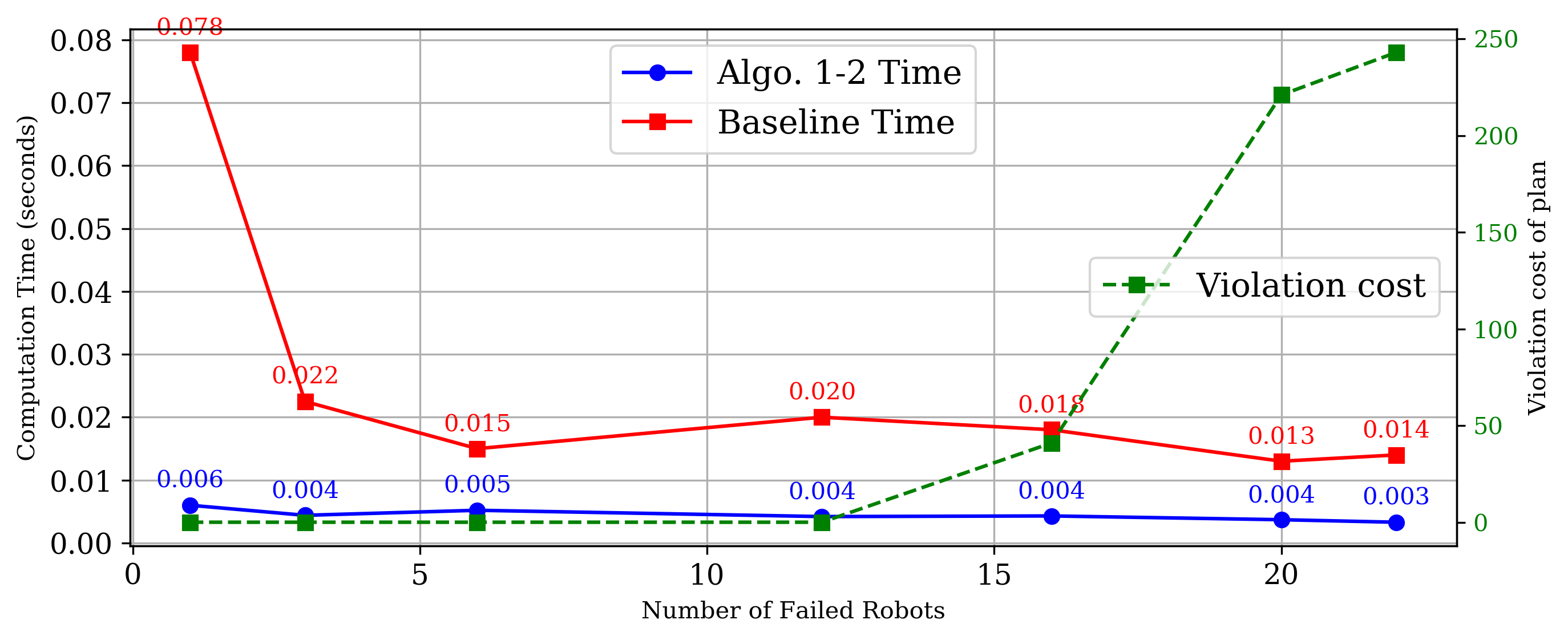}
    \caption{\textcolor{black}{Comparison of \textcolor{black}{Alg. \ref{alg:RP}-\ref{alg:bfs} and the Hungarian-based baseline} in a setting with 24 robots. Failures occur at $t=2$ and timing is reported for reassigning all failed transitions in the NBA. }}
    \label{fig:comparison_figure_24_bot}
\end{figure}

\textcolor{black}{We compare the runtimes of our reallocation algorithm and the baseline as the number of failures increases. The failure counts for each run are $[1,3,6,12,16,20,22]$, occurring at $t=2$. The results, summarized in Fig. \ref{fig:comparison_figure_24_bot}, show the total time required to reassign \textit{all} failed predicates across \textit{all} transitions in the NBA using Alg. \ref{alg:RP}-\ref{alg:bfs} and the baseline. Our method consistently outperforms the baseline in computation time, particularly for smaller failure counts. We note that the baseline time drops with increasing failures till the number of surviving robots is less than number of sub-tasks, after which the times remain consistent. Finally, the performance gap tends to widen significantly as the number of predicates on the NBA edges affected by the failures increases; see Remark \ref{rem:100Bots}.} 

\begin{figure}[ht]
    \centering
    \includegraphics[width=0.4\textwidth]{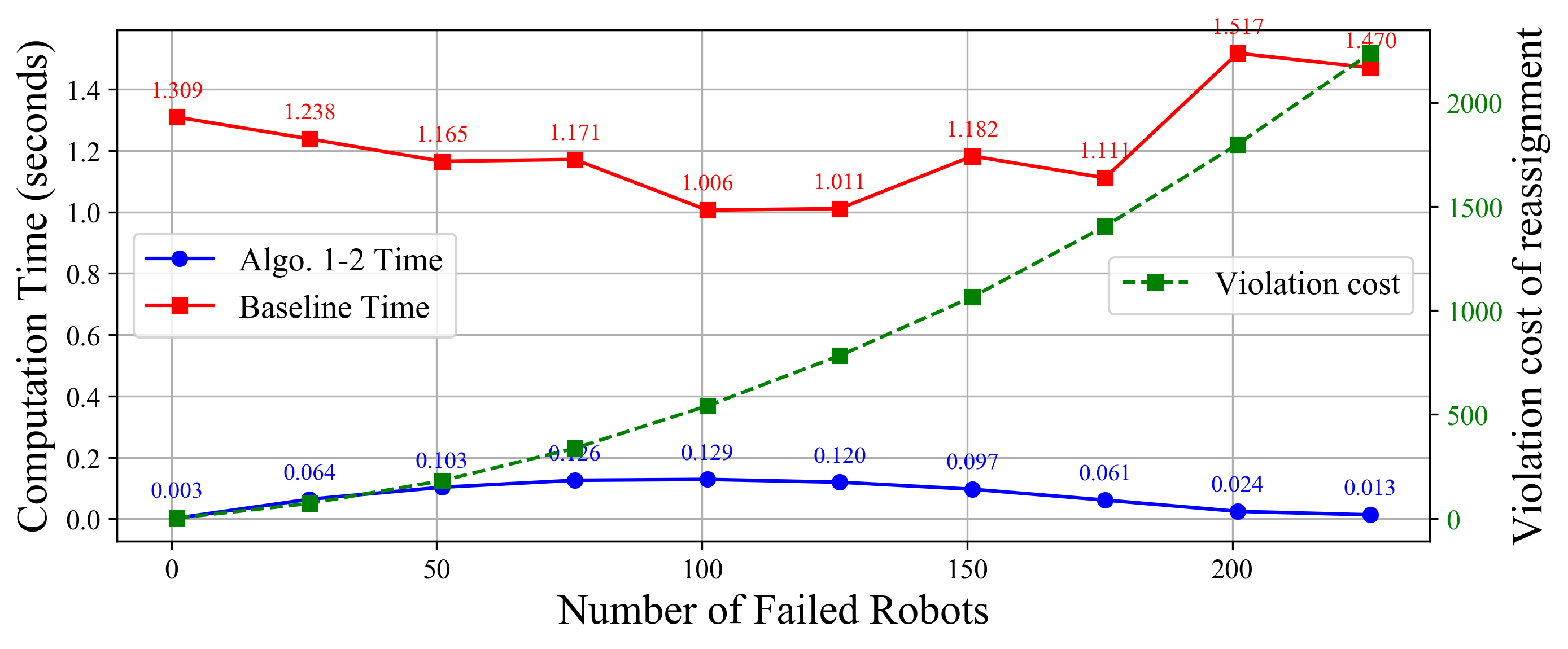}
    \caption{\textcolor{black}{\textcolor{black}{Comparison of Alg. \ref{alg:RP}-\ref{alg:bfs} and the Hungarian baseline} on a single NBA transition  with $250$ predicates. Violation cost is identical in both methods, but Alg. \ref{alg:bfs} reassigns substantially fewer robots and is more time efficient.}}
    \label{fig:comparison_figure2}
\end{figure}

\textcolor{black}{
\begin{rem}[Scalability Test]\label{rem:100Bots}
    We evaluate Algorithm~\ref{alg:bfs}'s scalability and efficiency on a single large NBA transition in a synthetic scenario with $250$ robots and $250$ tasks, each requiring one of three skills, with penalties randomly set between $2$ and $20$. The mission follows the LTL formula $\phi = \Diamond\pi_1 \wedge \Diamond\pi_2 \wedge \dots \wedge \Diamond\pi_{250}$, \textcolor{black}{where $\pi_j$ is a predicate in the form of \eqref{eq:pip} modeling the sub-task assigned to robot $j$. 
    Among all transitions in the resulting NBA, we focus on a specific transition, from the initial to the final state,} that requires all predicates $\pi_j$ to be true at the same time. We compare the runtime of our re-assignment algorithm and the baseline in this \textit{singular} transition under increasing failure counts $[1, 26, 51, \dots, 226]$. 
    Results in Fig.~\ref{fig:comparison_figure2} show both methods give valid reassignments with minimum violation, but the baseline reassigns all surviving robots which increases the computational time. Conversely, Algorithm~\ref{alg:bfs} minimizes reassignments to maintain feasibility, achieving lower runtimes. \textcolor{black}{For instance, when $1$ failure occurs, our method and the baseline require 0.003 and 1.309 seconds.} The gap in computational performance increases as the number of predicates in the NBA edge increases. For instance, when reassigning after a single failure in a transition with 500 predicates, our method takes 0.01 seconds, while the baseline takes an average of 13.2 seconds. 
    This demonstrates that our approach scales efficiently for large reallocation problems while matching optimal outcomes.
\end{rem}
}


\begin{rem}[Parallel Implementation]\label{rem:parallel}
    \textcolor{black}{In our implementation of the local replanner, we sequentially design sub-plans that connect the end of one true overlap to the start of the next. However, these sub-plans can be computed in parallel, as they are independent of one another. This parallel implementation could accelerate the local replanner and potentially result in shorter runtimes compared to the global replanner, even in cases involving a large number of failures. \textcolor{black}{Similarly, we do task reassignment for each edge $e \in \ccalE_{\pi}$ sequentially (line \ref{rp:collect}, Alg. \ref{alg:RP}) but this too can be done in parallel as all sub-tasks are independent (Assumption \ref{as3})}.} 
\end{rem}

\begin{figure}[t]
  \centering
\label{fig:Gazebo2}
\includegraphics[width=0.9\linewidth]{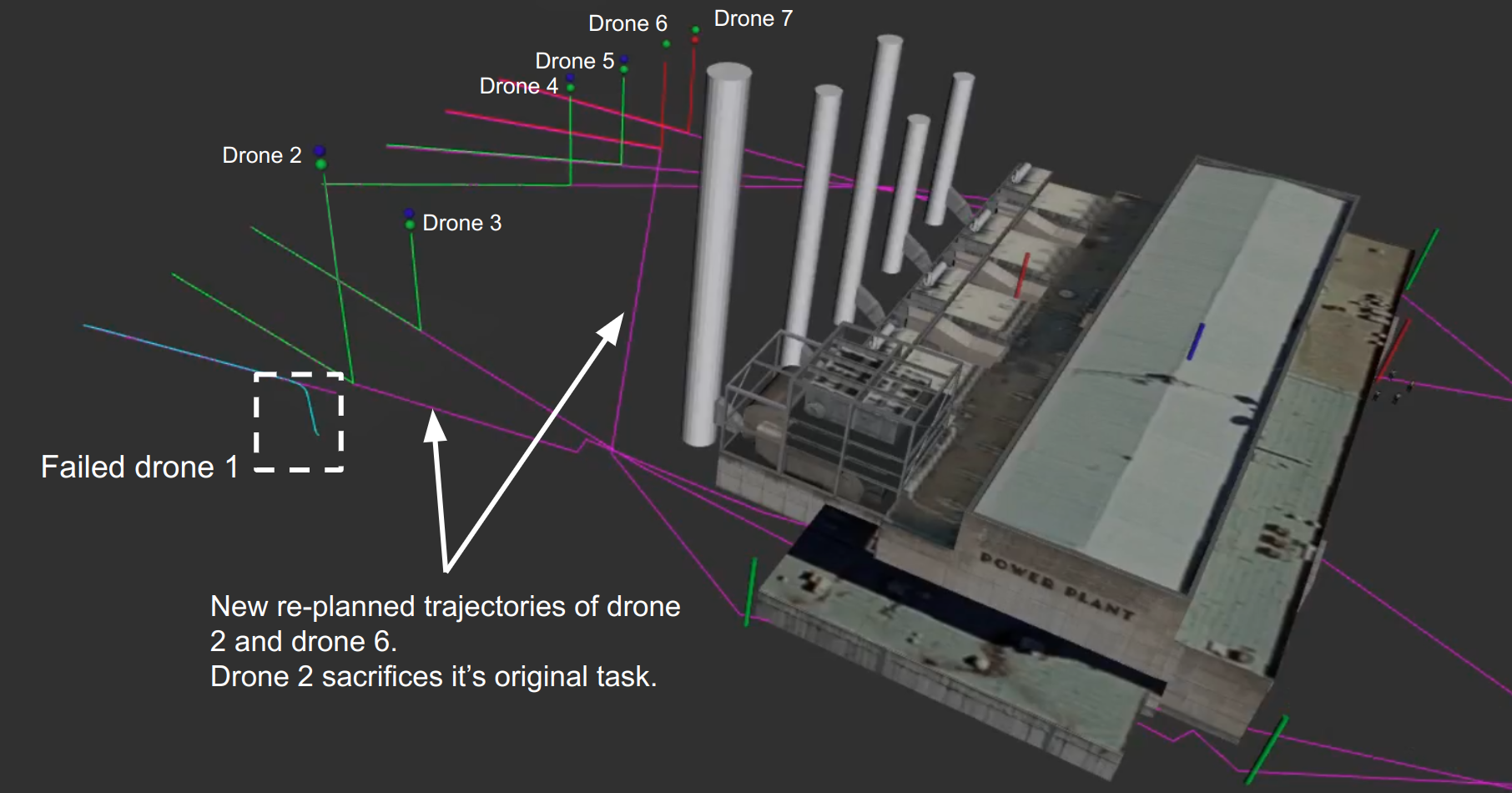}
\vspace{-0.2cm}
\caption{\textcolor{black}{Moment when Drone $1$ fails and new plans (in pink) are re-planned for Drones $2$ and $6$. Failed Drone $1$'s transmission task is assigned to Drone $2$. Drone $6$ takes photos completing the task initially assigned to drone $2$ and is in turn forced to sacrifice its originally assigned task; see video in \cite{SimResilientViolation}. 
}}
\vspace{-0.1cm}
\label{fig:Gazebo_fig}
\end{figure}

\vspace{-0.3cm}
\subsection{Aerial reconnaissance task - Multiple failures}\label{sec:aerialGazebo}
\textcolor{black}{In this section, we present a simulation conducted in Robotic Operating System (ROS) using AsTech Firefly Unmanned Aerial Vehicles (UAVs) that operate over a city with dimensions $150m\times 150m$. 
The AsTech Firefly UAV operates under first-order dynamics, where the state includes its position, velocity, orientation, and biases in measured angular velocities and acceleration; more information is available in \cite{Furrer2016}. Generally, more complex robot dynamics require longer time for sampling-based methods to generate feasible plans, as they must explore a larger state and control space. To address this, we first generate plans using the holonomic drive dynamics described in \eqref{eq:nonlinRbt}.
Then, given the plan waypoints, we compute minimum snap trajectories that smoothly transition through all waypoints every $T=1$ seconds, for all drones; \textcolor{black}{this also ensures synchronous motion of the drones} \cite{mellinger2011minimum}.
The UAVs are controlled to follow these trajectories using the ROS package developed in \cite{Furrer2016}.}

\textcolor{black}{We consider a surveillance mission involving $N=7$ drones. 
The abilities of the drones are defined as $c_1$, $c_2$, $c_3$, and $c_4$ pertaining to mobility, data transmission capability, RGB photos, and infrared imaging, respectively.
Drone $1$ has abilities $c_1$ and $c_2$. Drones $2$, $3$, $4$ and $5$ have abilities $c_1$, $c_2$, and $c_3$, and drones $6$ and $7$ have abilities $c_1$, $c_3$, $c_4$. 
\textcolor{black}{This mission is captured by the following formula: $ \phi = \Diamond(\pi_1\wedge\xi_1\wedge\xi_2)\wedge\Bar{\pi}_2\bigcup\pi_1\wedge\Diamond(\pi_3\wedge(\pi_4\vee\pi_5))\wedge\Diamond((\pi_3\wedge(\pi_6\vee\pi_7))\wedge\square\Diamond(\pi_3\wedge\pi_8)\wedge\square\Diamond(\pi_3\wedge\pi_9)$, where $\xi_1$ and $\xi_2$ are Boolean formulas defined as $\xi_1 =  \pi_{\ccalT_{c_3}}(2, \ell_2)\wedge\pi_{\ccalT_{c_3}}(3, \ell_3)\wedge\pi_{\ccalT_{c_3}}(4, \ell_4)\wedge\pi_{\ccalT_{c_3}}(5, \ell_5)$, $\xi_2=\pi_{\ccalT_{c_4}}(6, \ell_6)\wedge\pi_{\ccalT_{c_4}}(7, \ell_7)$, and 
$\pi_1=\pi_{\ccalT_{c_2}}(1, \ell_1)$,
$\Bar{\pi}_2=\Bar{\pi}_{\ccalT_{c_2}}(\varnothing, c_2, \ell_{13})$,
$\pi_3=\pi_{\ccalT_{c_2}}(1, \ell_{13})$,
$\pi_4=\pi_{\ccalT_{c_4}}(6, \ell_{9})$,
$\pi_5=\pi_{\ccalT_{c_3}}(6, \ell_{9})$,
$\pi_6=\pi_{\ccalT_{c_3}}(4, \ell_{10})$,
$\pi_7=\pi_{\ccalT_{c_3}}(3, \ell_{11})$,
$\pi_8=\pi_{\ccalT_{c_3}}(7, \ell_{8})$,
$\pi_9=\pi_{\ccalT_{c_4}}(7, \ell_{12})$.}
The associated penalties are, $F(\pi)=25, \forall\pi\in\xi_1$, $F(\pi)=10, \forall\pi\in\xi_2$, $F(\pi_1)=F(\pi_3)=50$, $F(\pi_4)=F(\pi_5)=F(\pi_9)=10$, and $F(\pi_6)=F(\pi_7)=F(\pi_{8})=20$.
Essentially, the drones need to accomplish a surveillance mission on an enemy location. \textcolor{black}{Due to terrain-induced communication limitations, drone 1 ($\ccalT_{c_2}$) flies at a higher altitude and turns on the communication relay to transmit data to the base station only when required (when other drones collect data) to maintain stealth.} The first task requires simultaneous collection of photos and thermal images ($\xi_1$, $\xi_2$) at multiple sites and concurrent transmission of the collected data ($\pi_1$) to obtain situational snapshot (e.g., of guard positions). \textcolor{black}{Drone $1$ must relay this data before doing the next part of the mission (captured by $\Bar{\pi}_2\bigcup\pi_1$)}. This is followed by optional photo/thermal imaging tasks such as ($\pi_4$ or $\pi_5$) and ($\pi_6$ or $\pi_7$), with concurrent transmission ($\pi_3$). Finally, surveillance and transmission at $(\pi_3 \wedge \pi_8)$ and $(\pi_3 \wedge \pi_9)$ must be carried out by drones $1$ and $7$ infinitely often, modeling persistent monitoring at key checkpoints.
This mission corresponds to an NBA with $13$ states.}

\textcolor{black}{We simulate the complete failure of drone $1$ at $t=5$ (when $q_B(5)=q_B^0$). 
Due to this, $|\ccalE|=13$ NBA edges need to be repaired. 
Algorithm \ref{alg:RP} reassigns drone $1$'s transmission task to drone $2$ and drone $2$'s task of taking a photo to and $6$. In this scenario, drone $6$ sacrificed its task of taking infrared image as $\xi_2$ had a smaller penalty.
This reassignment process took $0.0012$ secs and planning the new plans took $2.533$ secs. 
A screenshot of the simulation is shown in Fig. \ref{fig:Gazebo_fig}. 
At $t=25$, robots $3$, $4$, $5$, and $7$ fail completely as well. However in this case, robots $2$ and $6$ can take over all the failed tasks (reassignment in $0.001$ secs and re-planning in $0.5883$ secs) and complete the mission with no additional penalty.}

\vspace{-0.2cm}
\subsection{Hardware Validation}\label{sec:hardware}
\vspace{-0.1cm}
\textcolor{black}{We conducted a hardware experiment with 4 Crazyflie 2.1 drones to test the real-time performance of the minimum violation planner. LEDs are placed on the drones to indicate different skills being applied. The drones can fly ($c_1$), and implement blue skill ($c_2$ representative of taking photos) and implement red skill ($c_3$ representative of taking infrared images). Drone $1$ can perform all skills, drones $2$ and $3$ can fly and take photos, while drone $4$ can fly and take infrared images. We use the OptiTrack system to localize our drones in the environment. We run a program mimicking a health monitoring system that can communicate with the main planner the health of the drones and can induce a failure at any specified time.}

\textcolor{black}{The mission of the drones is captured by the following formula: $ \phi = \Diamond(\xi_1\wedge\Diamond(\pi_2\wedge\pi_3))\Diamond(\pi_4\wedge\Diamond\pi_5)\wedge\Bar{\pi}_6\bigcup\xi_1$,
where
$\xi_1=\pi_{\ccalT_{c_2}}(1, c_2, \ell_1)\wedge \pi_{\ccalT_{c_2}}(2, c_2, \ell_2)\wedge \pi_{\ccalT_{c_2}}(3, c_2, \ell_3)\wedge \pi_{\ccalT_{c_3}}(4, c_3, \ell_6)$,
$\pi_2=\pi_{\ccalT_{c_2}}(1, c_2, \ell_4)$, $\pi_3=\pi_{\ccalT_{c_2}}(3, c_2, \ell_5)$,
$\pi_4=\pi_{\ccalT_{c_3}}(4, c_3, \ell_7)$,
$\pi_5=\pi_{\ccalT_{c_3}}(4, c_3, \ell_8)$, and
$\Bar{\pi}_6=\pi_{\ccalT_{c_3}}(4, c_3, \ell_7)$. The penalties for not completing the tasks are  
$F(\pi)=30$ for all predicates $\pi$ in $\xi_1$, $F(\pi_2)=20$, $F(\pi_3)=50$, and $F(\pi_4)=F(\pi_5)=20$. 
This mission corresponds to an NBA with $12$ states. One timestep after $\xi_1$ is completed, we cause complete failure in drones $2$, $3$, and $4$. $|\ccalE|=15$ failed NBA edges are fixed in $0.0011$ seconds. After revision, drone $1$ is assigned to complete tasks $\pi_4$ and $\pi_5$. However, since $\pi_2$ and $\pi_3$ need to be done simultaneously, drone 1 sacrifices $\pi_2$ which it was originally assigned to, since $\pi_2$ has a lower penalty of $20$ and, instead, satisfies $\pi_3$ which has a higher penalty of $50$. The re-planning takes $0.06$ seconds. An additional case study is included in \cite{SimResilientViolation} where only drones $2$ and $4$ fail, and the planner is able to complete the mission without any penalty.}

\begin{figure}[t]
  \centering
\includegraphics[width=0.9\linewidth]{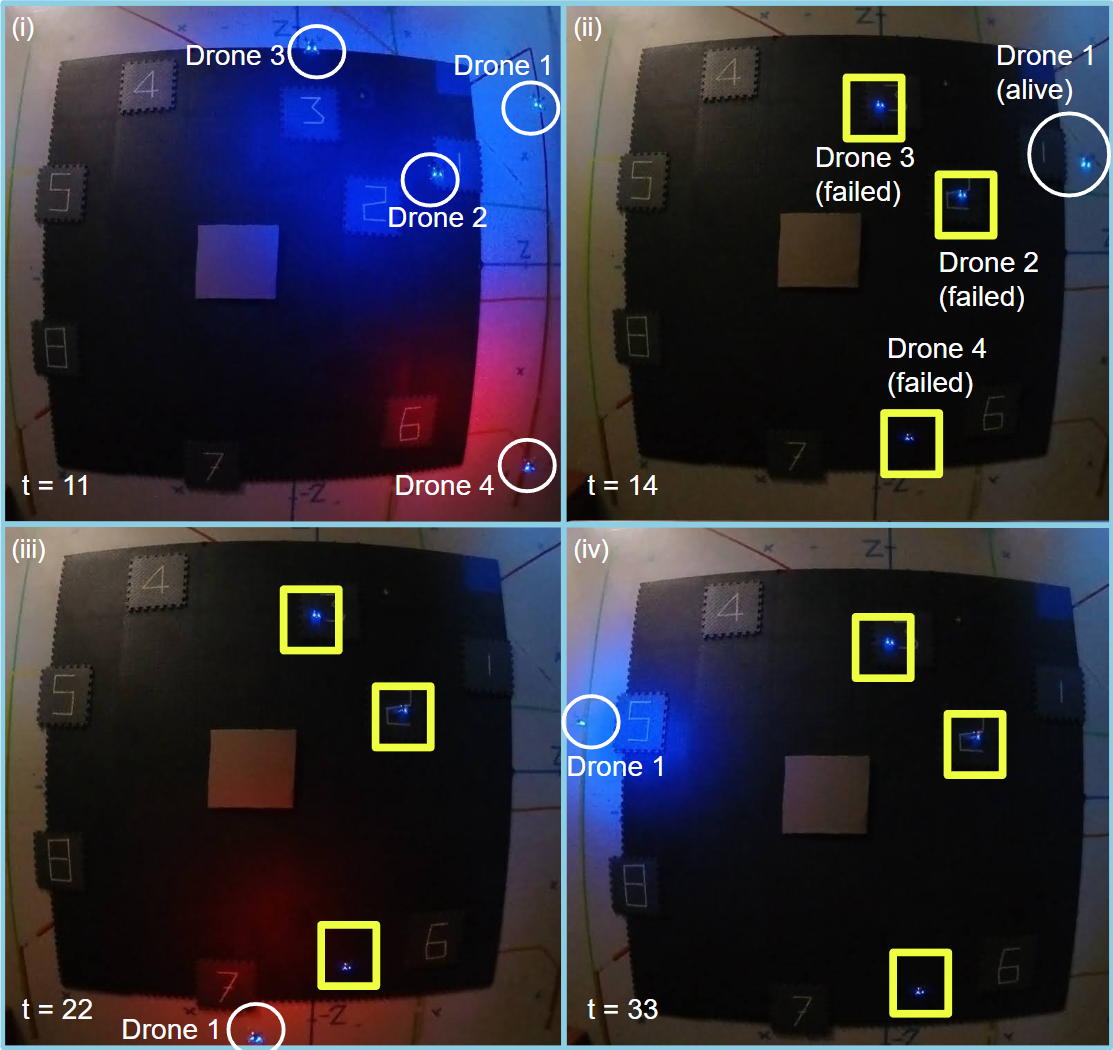}\vspace{-0.3cm}
\caption{Snapshots shows experiment in Sec. \ref{sec:hardware}. 
(i) Drones $1$, $2$, and $3$ take a photo, while drone $4$ takes infrared image. (ii) Drones $2$, $3$, and $4$ fail and fall down, and drone $1$ is reassigned to do tasks $\pi_3$, $\pi_4$, and $\pi_5$ [$t=14$]. (iii) Drone $1$ performs the task ($\pi_4$) of drone $4$ at $\ell_7$. (iv) Drone $1$ sacrifices its task ($\pi_2$) at $\ell_4$, and instead does the task ($\pi_3$) of drone $3$ at $\ell_5$.} 
\label{fig:hardware}
\end{figure}

 \vspace{-0.2cm}
\section{Conclusion} \label{sec:Concl}
\vspace{-0.1cm}
This paper proposed multi-robot temporal logic planning problem that can adapt to unexpected online robot capability failures. The main novelty of the proposed algorithm lies in its ability to design minimum-violation plans when mission completion becomes impossible due to limited number of functioning robots.  The proposed algorithm was supported both theoretically and experimentally. Our future work will focus on relaxing the assumption of independent sub-tasks in the LTL-encoded missions as well as on extending to language-based missions.
 \vspace{-0.2cm}
\appendices
\vspace{-0.2cm}
\section{Proofs of Propositions \ref{prop:opt1}-\ref{prop:opt2}}\label{sec:app1}
\subsection{Proof of Proposition \ref{prop:opt1}}
   \textcolor{black}{ This result holds by the construction of Alg. \ref{alg:bfs}. First consider the case where there exists a path $p$ towards a node $a^*$ with the lowest possible penalty, i.e., $\mathbb{C}_{q_B',q_B'',d}^{\pi}=0$. 
 Then Alg. \ref{alg:bfs} will find it by the completeness of the BFS algorithm. Also, since, by construction, search over $\ccalG$ occurs in a breadth-first manner, Alg. \ref{alg:bfs} will compute the path with the minimum number of hops from the root $a_\text{root}$. This equivalently results in the minimum number of re-assignments. }
    
  \textcolor{black}{Second, consider the case where there does not exist a path $p$ with $\mathbb{C}_{q_B',q_B'',d}^{\pi}=0$. In other words, there does not exist a node $a^*$ satisfying  $F(g_{q_B',q_B''}^d(a^*))=0$. Then Alg. \ref{alg:bfs} will exhaustively search over the entire graph and it will return the path from node $a^*$ to $a_\text{root}$ with the smallest cost $F(g_{q_B',q_B''}^d(a^*))$. If there exists more than one nodes achieving the same optimal cost, Alg. \ref{alg:bfs} returns the one that was computed first. Due to the breadth-first nature of the search process, this path has the smallest number of hops among all other paths reaching nodes with the same cost. This results in the minimum number of possible re-assignments.}

\vspace{-0.4cm}
\subsection{Proof of Proposition \ref{prop:opt2}}
\vspace{-0.1cm}

\textcolor{black}{Due to Assumption \ref{as3}, a failed predicate $\pi$ is repaired independently across all Boolean formulas $b_{q_B',q_B''}^d$, associated with an edge $e\in\ccalE_{\pi}$, and all edges $e\in\ccalE_{\pi}$. 
Thus, the result holds directly due to Proposition \ref{prop:opt1}.}

\vspace{-0.2cm}
\section{Proof of Proposition \ref{prop:optiLVP} and Corollaries \ref{cor:optCost}- \ref{cor:GlobalOptCost}}\label{sec:app2}
\vspace{-0.2cm}
\vspace{-0.7cm}
\textcolor{black}{
\subsection{Proof of Proposition \ref{prop:optiLVP}}
To show this result, we consider the following two complementary and mutually exclusive cases. Case I: Algorithm \ref{alg:local} computes a plan without triggering global re-planning. Case II: Algorithm \ref{alg:local} triggers global re-planning \textcolor{black}{(i.e., no true overlap was found)}.
We will show that in both cases, Algorithm \ref{alg:local} computes the plan, if it exists, with the lowest violation score, given the revised NBA, i.e., the re-assignments performed by Algorithms \ref{alg:RP}-\ref{alg:bfs}.}

\textcolor{black}{\textit{Case I:} In Case I, Alg. \ref{alg:local} will generate a plan, \textcolor{black}{if it exists,} denoted by $\tau_H^*$, that goes through all the NBA states that appear in $\ccalP^{\text{min}}$ and only through them. \textcolor{black}{Note that if such a plan exists, the local re-planner  will find it as long as any existing complete temporal logic planner is used for local plan synthesis, such as \cite{luo2021abstraction,gujarathi2022mt,kantaros2020stylus}.
} Formally, the NBA run generated by $\tau_H^*$ can be expressed as $\rho=\rho^{\text{pre}}[\rho^{\text{suf}}]^\omega$. The prefix run is defined as 
\begin{align}\label{prerun}
\rho^{\text{pre}}=&\underbrace{\ccalP^\text{min}_{\text{pre}}(1),\dots,\ccalP^\text{min}_{\text{pre}}(1)}_{K_1~ \text{times}},\underbrace{\ccalP^\text{min}_{\text{pre}}(2),\dots,\ccalP^\text{min}_{\text{pre}}(2)}_{K_2 ~\text{times}},\dots,\\&\underbrace{\ccalP^\text{min}_{\text{pre}}(m),\dots,\ccalP^\text{min}_{\text{pre}}(m)}_{K_m~ \text{times}},\dots,\ccalP^\text{min}_{\text{pre}}(K^{\text{min}}_{\text{pre}})\nonumber
\end{align}
where $K^{\text{min}}_{\text{pre}}$ is the length of the sequence $\ccalP^\text{min}_{\text{pre}}$, i.e., $K^{\text{min}}_{\text{pre}}=|\ccalP^\text{min}_{\text{pre}}|$. Observe in \eqref{prerun} that $\rho^{\text{pre}}$ goes through all NBA states of $\ccalP^\text{min}_{\text{pre}}$ in the same order that they appear in $\ccalP^\text{min}_{\text{pre}}$. However, the robot may stay at an NBA state $\ccalP^\text{min}_{\text{pre}}(m)$ for $K_m\geq 1$ time steps before moving to the next state $\ccalP^\text{min}_{\text{pre}}(m+1)$, for all $m\in\{1,\dots,K^{\text{min}}_{\text{pre}}-1\}$. 
The reason is that there may not exist multi-robot plans that can move from $\ccalP^\text{min}_{\text{pre}}(m)$ to $\ccalP^\text{min}_{\text{pre}}(m+1)$ within 1 time step. The suffix run $\rho^{\text{suf}}$ can be defined accordingly using $\ccalP^\text{min}_{\text{suf}}$ instead of $\ccalP^\text{min}_{\text{pre}}$.}

\textcolor{black}{If the NBA satisfies Assumption \ref{cond:NBA_self_loop}, then this means that the violation score $\mathbb{C}_{\sigma}$ (see \eqref{eq:edgeviolation}) of all symbols $\sigma$ generated by $\tau_H^*$ associated with self loops in $\rho$ is $0$.\footnote{By self loops in $\rho$, we refer to transitions where the next NBA state in $\rho$ is the same as the previous one.} The reason is that these self loops are associated with Boolean formulas that are either always true or defined over predicates of the form \eqref{eq:negpip}. In the former case, the violation score $\mathbb{C}_{\sigma}$ is trivially $0$ while in the latter case $\mathbb{C}_{\sigma}$ is also $0$ since the planner is not allowed to violate predicates of the form \eqref{eq:negpip}; see restriction (ii) in Section \ref{sec:onlineReplan} (\textit{Local Plan Synthesis}).
Thus, a non-zero, and finite, violation score of the new plan $\tau_H^*$ may occur only when transitions to a new NBA is made, i.e., when $q_B''=\ccalP^\text{min}_{\text{pre}}(m+1)$ is reached from $q_B'=\ccalP^\text{min}_{\text{pre}}(m)$, for some/all $m\in\{1,\dots,K^{\text{min}}_{\text{pre}}-1\}$; recall also that $\tau_H^*$ is designed so that the transition from $q_B'$ to $q_B''$ is enabled by satisfying the Boolean sub-formula $b_{q_B',q_B''}^d$, where $d=\ccalD^{\text{min}}(m)$. In what follows, we show that this implies 
\begin{equation}\label{eq:CteqCp}
\mathbb{C}_{\tau_H^*}(t)=\mathbb{C}_{\ccalP^{\text{min}}},
\end{equation}
where $\mathbb{C}_{\ccalP^{\text{min}}}$ and $\mathbb{C}_{\tau_H^*}$ are defined in \eqref{eq:NBAviolation} and \eqref{eq:violation}, respectively. \textcolor{black}{If \eqref{eq:CteqCp} holds, then this implies} that if there exists a plan $\bar{\tau}_{H}^*$ so that  $\mathbb{C}_{\bar{\tau}_{H}^*}< \mathbb{C}_{\tau_H^*}$ then there must exist a path $\ccalP$, constructed as discussed in Section \ref{sec:onlineReplan}, so that $\mathbb{C}_{\ccalP}<\mathbb{C}_{\ccalP^{\text{min}}}$. However, this cannot occur by construction of $\ccalP^{\text{min}}$. Thus, we conclude that in Case I, Algorithm \ref{alg:local} will generate a plan, if it exists, that achieves the lowest finite violation cost \eqref{eq:violation}.}

To show \eqref{eq:CteqCp}, let $t'$ denote the time step when the  transition from $q_B'=\ccalP^\text{min}(m)$ to $q_B''=\ccalP^\text{min}(m+1)$ is enabled by the plan $\tau_H^*$, where
$b_{q_B',q_B''}=\bigvee_{d=1}^D b_{q_B',q_B''}^d$. Let $\sigma^d(t')$ denote the symbol generated by $\tau_H^*$ at $t'$, i.e., $\sigma^d(t')=L(\bbp(t'),\bbs(t'))$, where $\tau_H^*(t')=[\bbp(t'),\bbs(t'),q_B']$, to satisfy $b_{q_B',q_B''}^d$, $d=\ccalD(m)$. 
Also, let $\mathcal{AP}_b^d$ collect all predicates that appear in $b_{q_B',q_B''}^d$ and let $\Sigma_b^d=2^{\mathcal{AP}_b^d}$. 
Then, we define the set $\Sigma_b^{*,d}=\{\sigma\in\Sigma_b^d~|~\sigma^d(t')\sigma\models b_{q_B',q_B''}^d \}$. In words, this set collects all symbols $\sigma$ that if they had been generated by $\tau_H^*(t')$, then the transition from $q_B'$ to $q_B''$ would have been enabled without incurring any penalty. 
%
\textcolor{black}{Since the Boolean formula $b_{q_B',q_B''}^d$ does not contain any disjunctions by construction, we have that that $\Sigma_b^{*,d}$ is either empty or singleton. Thus, using \eqref{eq:edgeviolation}, we get that the violation score of the symbol $\sigma^d(t')$ is: 
\begin{equation}\label{eq:proofViolationD}
\mathbb{C}_{\sigma^d(t')} =  \sum_{\pi\in\sigma} (F(\pi)),
\end{equation}
where $\sigma$ is the single element in $\Sigma_b^{*,d}$ that may consist of multiple predicates $\pi$. By definition of the violation cost of a plan in \eqref{eq:violation}, $\mathbb{C}_{\sigma^d(t')}$ will be the violation cost incurred to transition from $\tau_H^*(t')=[\bbp(t'),\bbs(t'),q_B]$ to $\tau_H^*(t'+1)=[\bbp(t'+1),\bbs(t'+1),q_B']$, i.e., to enable the NBA transition from $q_B'=\ccalP^{\text{min}}(m)$ to $q_B''=\ccalP^{\text{min}}(m+1)$, for all $m\in\{1,\dots,|\ccalP^{\text{min}}|-1\}$.} Let $\ccalL$ be a set collecting the time steps $t'\in\{t,\dots,T+K\}$ where a transition from $q_B'=\ccalP^{\text{min}}(m)$ to $q_B''=\ccalP^{\text{min}}(m+1)$ is enabled, for all $m\in\{1,\dots,|\ccalP^{\text{min}}|-1\}$.  Since, as discussed earlier, any violation cost in $\tau_H^*$ is incurred only at time steps $t'\in\ccalL$, \textcolor{black}{by applying \eqref{eq:violation},} we get that the cost of $\tau_H^*$ is:
\begin{equation}\label{eq:newCostPlan}
    \mathbb{C}_{\tau_H^*}(t)=\sum_{t'\in\ccalL}\mathbb{C}_{\sigma^d(t')}
\end{equation}

Now let us look at the assignment cost of the same transition $b_{q_B',q_B''}^d$, where $d=\ccalD^{\text{min}}(m)$. Let $\mathcal{AP}^U_{q_B',q_B'',d}$ collect any unassigned/sacrificed predicates in $b_{q_B',q_B''}^d$ after the reassignment at time $t$. Note that this set will also consist of any predicates that were sacrificed when fixing failures that occurred at past time steps $t''<t$.
Thus the total assignment cost for this transition, given by (\ref{eq:NBAEdgeviolation}), is 
\begin{equation}\label{eq:proofAssignmentD}
\mathbb{C}_{q_B',q_B'',d} = \sum_{\pi\in\mathcal{AP}^U_{q_B',q_B'',d}}\mathbb{C}^{\pi}_{q_B',q_B'',d}  = \sum_{\pi\in\mathcal{AP}^U_{q_B',q_B'',d}}F(\pi), 
\end{equation}
\textcolor{black}{where the last equality is due to \eqref{eq:assignmentCost}.}

\textcolor{black}{Now let revisit how our local re-planner designs plans. Here we consider two cases for $q_B'=\ccalP^\text{min}_{\text{pre}}(m)$ to $q_B''=\ccalP^\text{min}_{\text{pre}}(m+1)$: (i) our local re-planner needs to design a new plan (connecting the end of a true overlap to the start of the next true overlap) that goes through the NBA states $q_B'$ to $q_B''$ appearing in $\ccalP_{\text{min}}$; (ii) our local re-planner reuses a previously designed plan that goes through the NBA states $q_B'$ to $q_B''$ appearing in $\ccalP_{\text{min}}$.}

\textcolor{black}{First, we focus on case (i).} 
\textcolor{black}{Since the local re-planner sacrifices unassigned tasks by considering them to be true, the predicates present in the symbol $\sigma$, where $\Sigma_b^{*,d}=\{\sigma\}$ (if $\Sigma_b^{*,d}$ is non-empty), would be only the sacrificed predicates that do not have any assigned robots after running Alg. \ref{alg:bfs}.} These are same predicates that are collected in $\mathcal{AP}^U_{q_B',q_B'',d}$ by construction of this set, i.e., 
\textcolor{black}{$\mathcal{AP}^U_{q_B',q_B'',d}=\Sigma_b^{*,d}$}. Thus, the costs $\mathbb{C}_{\sigma^d(t')}$ and $\mathbb{C}_{q_B',q_B'',d}$ computed in \eqref{eq:proofViolationD} and \eqref{eq:proofAssignmentD}, respectively, are the same as both sum up the penalties (denoted by $F(\pi)$) incurred over the same set of predicates. This gives us: 
\vspace{-0.2cm}
\begin{align}\label{eq:violationEdgeEquality}
   \mathbb{C}_{\sigma^d(t')} = \mathbb{C}_{q_B',q_B'',d},
\end{align}
for all edges $(\ccalP^{\text{min}}(m),\ccalP^{\text{min}}(m+1))$ and $d=\ccalD^{\text{min}}(m)$ \textcolor{black}{associated with case (i)}.

\textcolor{black}{Second, we focus on case (ii) and our goal is to show that \eqref{eq:violationEdgeEquality} still holds. Here the re-used plan has either a non-zero violation score because it sacrificed satisfaction of any unassigned predicate (due to a previous failure) to enable the transition from $q_B'$ to $q_B''$ or a zero violation cost (i.e., no predicates were sacrificed).}
In both situations \eqref{eq:violationEdgeEquality} still holds by using exactly the same analysis as in case (i). 
\textcolor{black}{Specifically, since the plan is re-usable, it must satisfy the condition (2) of the reusability criteria ensuring that given the robot task assignments for the predicates, including the unassigned predicates in $b_{q_B',q_B''}^d$, the reused path will satisfy it (see \textit{Conditions for Re-usable Plans} in Section \ref{sec:onlineReplan}). Thus, the sacrificed predicates (if any) in the symbol $\sigma$ will be same as the unassigned predicates in $\mathcal{AP}^U_{q_B',q_B'',d}$ and, therefore, \eqref{eq:violationEdgeEquality} holds.} 

\textcolor{black}{Thus, we conclude that \eqref{eq:violationEdgeEquality} holds for all edges $(\ccalP^{\text{min}}(m),\ccalP^{\text{min}}(m+1))$ and $d=\ccalD^{\text{min}}(m)$, whether they fall in case (i) or (ii). Therefore,} we can re-write \eqref{eq:newCostPlan} as:
\begin{equation}\label{eq:endProof1}
    \mathbb{C}_{\tau_H^*}(t)=\sum_{m=1}^{|\ccalP^{\text{min}}|-1}\mathbb{C}_{q_B',q_B'',d},
\end{equation}
where $q_B'=\ccalP^{\text{min}}(m)$, $q_B''=\ccalP^{\text{min}}(m+1)$, and $d=\ccalD^{\text{min}}(m)$. Comparing the costs in \eqref{eq:endProof1} and \eqref{eq:NBAviolation}, we get $\eqref{eq:CteqCp}$ completing the Case I part of the proof.


\textcolor{black}{\textit{Case II:} \textcolor{black}{In Case II, global re-planning is triggered when there does not exist $\ccalP^{\text{min}}$ that results in a non-empty set $\ccalO^*$.}
In this case, global replanning starts from the state $\tau_H(t)$ where the robots are when the failures occur so that a new prefix-suffix plan is constructed. Alg. \ref{alg:local} performs global re-planning by using any existing complete and optimal temporal logic planner, such as \cite{luo2021abstraction,gujarathi2022mt,kantaros2020stylus}. Thus, Algorithm \ref{alg:local} will return a plan, if it exists, that minimizes \eqref{eq:violation}, completing the proof.
}

\vspace{-0.3cm}
\subsection{Proof of Corollary \ref{cor:optCost}}
\vspace{-0.1cm}

    \textcolor{black}{This result holds by construction of the local re-planner. First, consider the case where the local replanner computes a finite sub-plan connecting the end of a true overlap, denoted by $\hat{\tau}_H(A)$, to the start of the next true overlap corresponding to a state $\hat{\tau}_H(B)$. Due to the restrictions (i)-(ii) (in Section \ref{sec:onlineReplan} \textit{Local Plan Synthesis}), \textcolor{black}{any} sub-plans, computed by the local re-planner, connecting $\hat{\tau}_H(A)$ to $\hat{\tau}_H(B)$ will have the same violation cost \eqref{eq:violation}. Thus, this means that all revised plans $\tau_H^*$ (constructed by stitching together re-used and newly designed sub-plans) computed by the local re-planner share the same violation cost.\footnote{In practice, among them, we pick one randomly or based on any user-specified criterion such as traveled distance).}  %
    Second, in Proposition \ref{prop:optiLVP}, we show that the cost of the optimal plan, in terms of \eqref{eq:violation} is equal to $\mathbb{C}_{\ccalP_{\text{min}}}$ (see \eqref{eq:CteqCp}). By combining these two observations, we conclude that any plans returned by the local replanner are optimal and their violation cost is $\mathbb{C}_{\ccalP^{\text{min}}}$. Also, if such plans exist, the local replanner will find them as long as a complete planner is used for local plan synthesis (e.g., \cite{luo2021abstraction}), thus completing the proof.} 

\vspace{-0.3cm}
\subsection{Proof of Corollary \ref{cor:GlobalOptCost}}
\vspace{-0.1cm}

\textcolor{black}{Following a similar analysis as the one in Case I of Proposition \ref{prop:optiLVP}, we can show that if \textcolor{black}{there exists a plan} with violation cost equal to $\mathbb{C}_{\ccalP^{\text{min}}}$, \textcolor{black}{then that plan is the optimal one.} 
The global re-planner will compute the optimal plan $\tau_H^*$, as long as global replanning is performed by an optimal planner (e.g., \cite{luo2021abstraction}). If \textcolor{black}{a plan that enabling the sequence of NBA transitions in $\ccalP^{\text{min}}$ does not exist, then the global re-planner will compute the optimal plan but its cost will be greater than $\mathbb{C}_{\ccalP^{\text{min}}}$ since $\mathbb{C}_{\ccalP^{\text{min}}}\leq\mathbb{C}_{\ccalP}, \forall \ccalP$.}
} 



\vspace{-0.2cm}
\section{Proof of Proposition \ref{prop:optiMVP}}\label{sec:app3}
\vspace{-0.1cm}

\textcolor{black}{We will show this result by contradiction. Specifically, our proof consists of two main steps. First, we will show that if there exists an assignment of predicates different from the one generated by Alg. \ref{alg:RP}-\ref{alg:bfs}, that would have allowed Alg. \ref{alg:local} to compute a plan $\hat{\tau}_H^*$ with lower cost, then that contradicts Proposition \ref{prop:opt2}. 
The second step will show that no other optimal temporal logic planner can compute a ‘better’ plan under any other reassignments, which directly follows from Proposition \ref{prop:optiLVP}. Specifically,
given the optimality properties of Alg. \ref{alg:local} (due to Proposition \ref{prop:optiLVP}) and that it cannot compute a plan better than $\tau_H^*$ under other possible reassignments (due to the first step), we conclude that no other optimal temporal logic planner could generate a plan with lower cost which will conclude the proof. In what follows we provide the detailed steps of the first step.}

\textcolor{black}{Assume that there exists a path $\hat{p}_{\pi}$, $\hat{p}_{\pi}\neq p_{\pi}$, that results in a plan $\hat{\tau}_H^*\neq\tau_H^*$ generated by any optimal temporal logic planner, such that:
\begin{equation}\label{eq:tocontradict}
\mathbb{C}_{\hat{\tau}_H^*}<\mathbb{C}_{\tau_H^*}
\end{equation}}

\vspace{-0.4cm}
\textcolor{black}{Recall from the proof of Proposition \ref{prop:optiLVP} that, under assumption (i),
the cost of the revised plan $\tau_H^*$, constructed using $(\ccalP^{\text{min}},\ccalD^{\text{min}})$, is $ \mathbb{C}_{\tau_H^*}(t)=\sum_{m=1}^{|\ccalP^{\text{min}}|-1}\mathbb{C}_{q_B',q_B'',d}$, where $\mathbb{C}_{q_B',q_B'',d}$ is the total assignment cost (see \eqref{eq:NBAEdgeviolation}) associated with repairing failed predicates for the NBA transition from $q_B'=\ccalP^{\text{min}}(m)$, to $q_B''=\ccalP^{\text{min}}(m+1)$, and $d=\ccalD^{\text{min}}(m)$ (see \eqref{eq:endProof1}). Due to assumption (ii), we have that $d$ is determined as per \eqref{condII} for all $m\in\{1,\dots,|\ccalP^{\text{min}}|-1\}$. We can similarly define the cost of the plan $\hat{\tau}_H^*$, designed by Alg. \ref{alg:local} using a sequence $(\hat{\ccalP}^{\text{min}},\hat{\ccalD}^{\text{min}})$ and reassignments determined by paths $\hat{p}_\pi$, as
$ \mathbb{C}_{\hat{\tau}_H^*}(t)=\sum_{m=1}^{|\hat{\ccalP}^{\text{min}}|-1}\hat{\mathbb{C}}_{\hat{q}_B',\hat{q}_B'',\hat{d}}$. In the definition of $\mathbb{C}_{\hat{\tau}_H^*}$, the `hat' on top of the variables is used, with slight abuse of notation, only to differentiate them from the corresponding ones used in $\mathbb{C}_{{\tau}_H^*}$. For instance,  $\hat{\mathbb{C}}_{\bar{q}_B,\bar{q}_B',d}$ denotes the total assignment cost when the failed predicates $\pi$ in $b_{\bar{q}_B,\bar{q}_B',d}$ are fixed using reassignments determined by $\hat{p}_{\pi}$. 
}

\textcolor{black}{Thus, if \eqref{eq:tocontradict} holds, there must exist at least one edge $e=(\bar{q}_B,\bar{q}_B')$ that 
$\hat{\tau}_H^*$ goes through,
associated with a Boolean formula $b_{\bar{q}_B,\bar{q}_B'}=\bigvee_{d=1}^D b_{\bar{q}_B,\bar{q}_B',d}$, that contains the currently failed predicate(s) and satisfies the following:\footnote{Note that the total assignment cost of NBA edges, as per \eqref{eq:proofAssignmentD}, that do not contain any failed predicates cannot be affected by Alg. \ref{alg:RP}. Also, the plan $\tau_H^*$ may not necessarily go through the edge  $e=(\bar{q}_B,\bar{q}_B')$.}
\begin{equation}\label{eq:ineq1}
    \min_{d\in\{1,\dots,D\}}\hat{\mathbb{C}}_{\bar{q}_B,\bar{q}_B',d}<    \min_{d\in\{1,\dots,D\}}\mathbb{C}_{\bar{q}_B,\bar{q}_B',d}.
\end{equation}
The result in \eqref{eq:ineq1} equivalently means that there exists at least one $d\in\{1,\dots,D\}$, denoted by $\bar{d}$, that satisfies:}
\begin{equation}\label{eq:ineq2}
\hat{\mathbb{C}}_{\bar{q}_B,\bar{q}_B',\bar{d}}<    \mathbb{C}_{\bar{q}_B,\bar{q}_B',\bar{d}}.
\end{equation}
\textcolor{black}{Using \eqref{eq:NBAEdgeviolation}, we can re-write \eqref{eq:ineq2} as:
\begin{equation}\label{eq:ineq3}
\sum_{\pi\in\hat{\mathcal{AP}}_{\bar{q}_B,\bar{q}_B',\bar{d}}^U}\hat{\mathbb{C}}_{\bar{q}_B,\bar{q}_B',\bar{d}}^{\pi}<\sum_{\pi\in\mathcal{AP}_{\bar{q}_B,\bar{q}_B',\bar{d}}^U} \mathbb{C}_{\bar{q}_B,\bar{q}_B',\bar{d}}^{\pi},
\end{equation}
where recall that $\mathcal{AP}_{\bar{q}_B,\bar{q}_B',\bar{d}}^U$ collects all predicates $\pi$ that remain unassigned after fixing the failed predicates in $b_{\bar{q}_B,\bar{q}_B',\bar{d}}$ using paths $p_{\pi}$ while $\mathcal{AP}_{\bar{q}_B,\bar{q}_B',\bar{d}}^U$ captures the same but using re-assignments determined by $\hat{p}_{\pi}$. Recall also that these sets collect predicates that remain unassigned due to failures occurred at past time steps $t' < t$. These `past' unassigned predicates exist in both $\mathcal{AP}_{\bar{q}_B,\bar{q}_B',\bar{d}}^U$ and $\hat{\mathcal{AP}}_{\bar{q}_B,\bar{q}_B',\bar{d}}^U$. The reason is that once a predicate becomes unassigned during the execution of Alg. \ref{alg:RP}-\ref{alg:local}, it will always remain unassigned. Thus, the sets $\mathcal{AP}_{\bar{q}_B,\bar{q}_B',\bar{d}}^U$ and $\hat{\mathcal{AP}}_{\bar{q}_B,\bar{q}_B',\bar{d}}^U$ differ only on the predicates that become unassigned due to repairing the failures occurred at time $t$ (i.e., the predicates in $\mathcal{AP}_F$). Thus, \eqref{eq:ineq3} implies that there exists at at least one failed predicate $\mathcal{AP}_F$, denoted by $\bar{\pi}$ appearing in $b_{\bar{q}_B,\bar{q}_B',\bar{d}}$ for which it holds that the cost of repairing it using $\hat{p}_{\pi}$ is smaller than when it is repaired using $p_{\pi}$, i.e., $\hat{\mathbb{C}}_{\bar{q}_B,\bar{q}_B',\bar{d}}^{\bar{\pi}}<\mathbb{C}_{\bar{q}_B,\bar{q}_B',\bar{d}}^{\bar{\pi}}$.
This in turn means that the re-assignment, as determined by $p_{\bar{\pi}}$, generated by Alg. \ref{alg:bfs}, is not optimal with respect to the assignment cost in \eqref{eq:assignmentCost}. However, this contradicts Proposition \ref{prop:opt2} completing the proof.}

\vspace{-0.1cm}

\bibliographystyle{IEEEtran}
\bibliography{SK_bib.bib}

\vspace{-1.5cm}
\begin{IEEEbiography}[{\includegraphics[width=1in,height=1.25 in,clip,keepaspectratio]{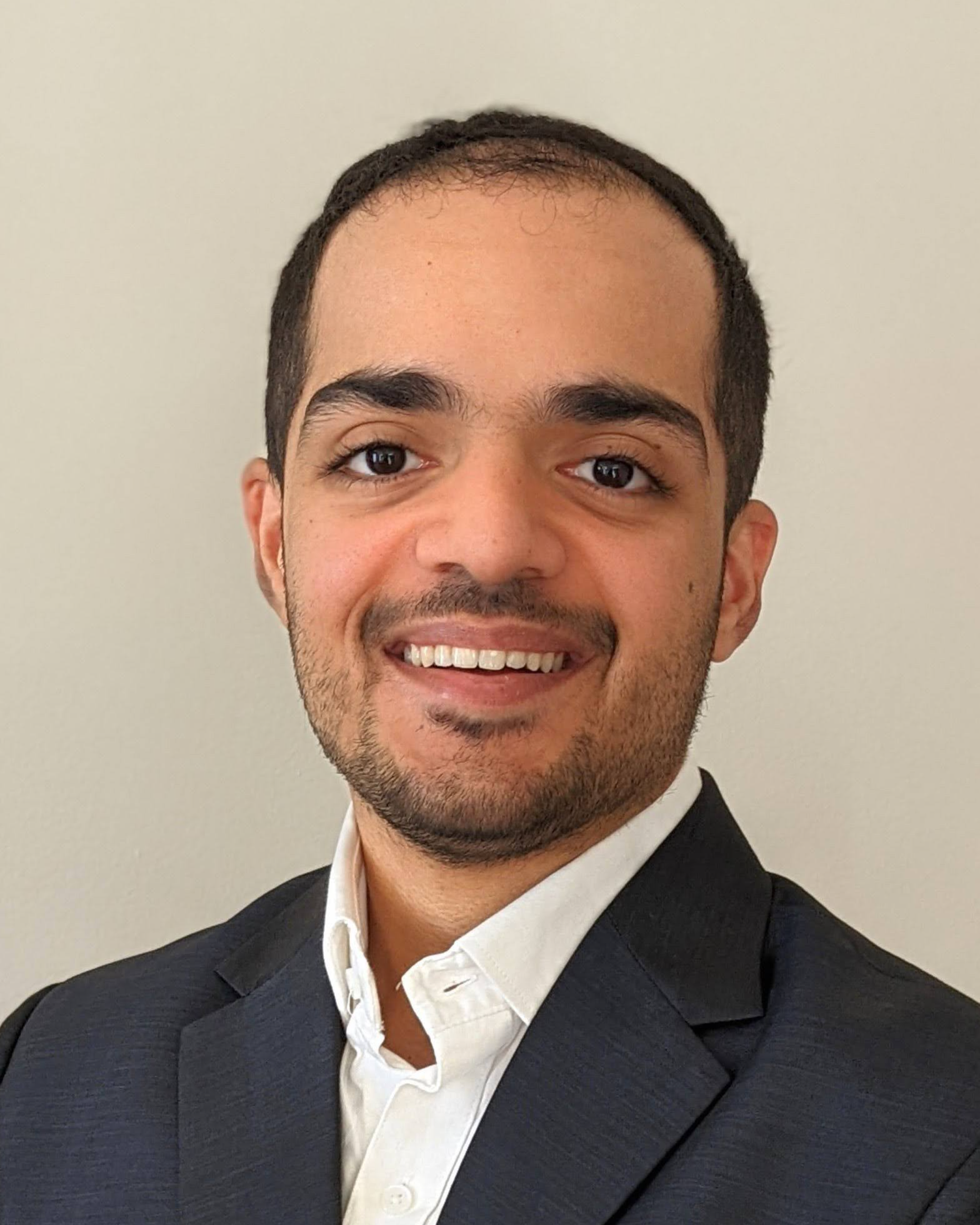}}]
{Samarth Kalluraya}  received the B.E. degree in mechanical engineering in 2018 from University of Pune, Pune, India and the M.S.E degree in mechanical engineering and applied mechanics from the University of Pennsylvania, Philadelphia, PA in 2020. He is currently pursuing a Ph.D. in electrical and systems engineering at Washington University in St. Louis, St. Louis, MO, USA. His research interests include motion planning, machine learning, and robot mission and task execution.
\end{IEEEbiography}

\vspace{-1.5cm}
\begin{IEEEbiography}[{\includegraphics[width=1in,height=1.25 in,clip,keepaspectratio]{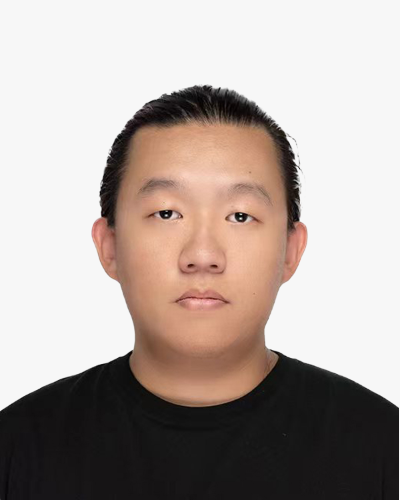}}]
{Beichen Zhou} received the B.E. degree in electronic engineering in 2023 from the University of Electronic Science and Technology of China, Chengdu, China, and is currently pursuing the M.S. degree in electrical and systems engineering at Washington University in St. Louis, MO, USA. His research interests include motion control, path planning, and human trajectory prediction, with applications in robotics and autonomous systems.
\end{IEEEbiography}

\vspace{-1.3cm} 

\begin{IEEEbiography}[{\includegraphics[width=1in,height=1.25in,clip,keepaspectratio]{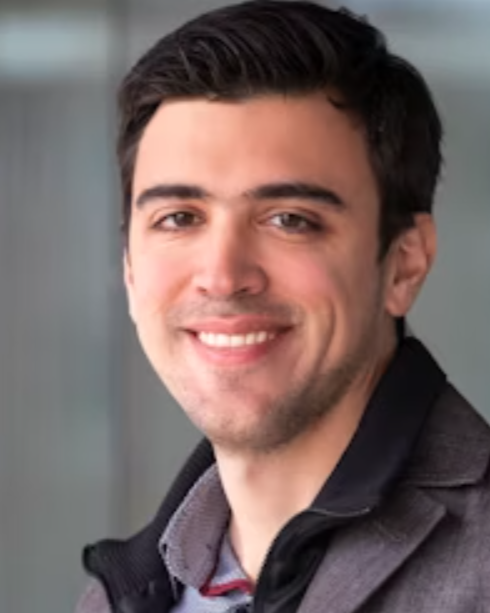}}]
{Yiannis Kantaros}(S'14-M'18) is an Assistant Professor in the Department of Electrical and Systems Engineering, Washington University in St. Louis (WashU), St. Louis, MO, USA. He received the Diploma in Electrical and Computer Engineering in 2012 from the University of Patras, Patras, Greece. He also received the M.Sc. and the Ph.D.  degrees in mechanical engineering from Duke University, Durham, NC, in 2017 and 2018, respectively. Prior to joining WashU, he was a postdoctoral associate in the Department of Computer and Information Science, University of Pennsylvania, Philadelphia, PA. His current research interests include machine learning, distributed control and optimization, and formal methods with applications in robotics. 
He received the Best Student Paper Award at the 2nd IEEE Global Conference on Signal and Information Processing (GlobalSIP) in 2014 and the Best Multi-Robot Systems Paper Award, Finalist, at the IEEE International Conference on Robotics and Automation (ICRA) in 2024. Additionally, he received the 2017-18 Outstanding Dissertation Research Award from the Department of Mechanical Engineering and Materials Science at Duke University and a 2024 NSF CAREER Award.
\end{IEEEbiography}

\end{document}